\documentclass[sn-mathphys-num]{sn-jnl}

\usepackage{graphicx}%
\usepackage{multirow}%
\usepackage{amsmath,amssymb,amsfonts}%
\usepackage{amsthm}%
\usepackage{mathrsfs}%
\usepackage[title]{appendix}%
\usepackage{xcolor}%
\usepackage{textcomp}%
\usepackage{manyfoot}%
\usepackage{booktabs}%
\usepackage{algorithm}%
\usepackage{algorithmicx}%
\usepackage{algpseudocode}%
\usepackage{listings}%

\usepackage{bbm}
\usepackage{ulem}
\usepackage{array}
\DeclareMathOperator*{\argmax}{arg\,max}
\DeclareMathOperator*{\argmin}{arg\,min}

\theoremstyle{plain}%
\newtheorem{theorem}{Theorem}
\newtheorem{proposition}[theorem]{Proposition}%
\newtheorem{assumption}{Assumption}

\theoremstyle{remark}%

\theoremstyle{definition}%

\begin{document}

\title[Article Title]{Applications of 0-1 Neural Networks in Prescription and Prediction}


\author[1]{\fnm{Vrishabh} \sur{Patil}}\email{vmpatil@andrew.cmu.edu}
\author[2]{\fnm{Kara} \sur{Hoppe}}\email{khoppe2@wisc.edu}

\author*[3]{\fnm{Yonatan} \sur{Mintz}}\email{ymintz@wisc.edu}


\affil[1]{\orgdiv{Tepper School of Business}, \orgname{Carnegie Mellon University}, \orgaddress{\street{5000 Forbes Avenue}, \city{Pittsburgh}, \postcode{15213}, \state{PA}, \country{USA}}}

\affil[2]{\orgdiv{Maternal-Fetal Medicine} \orgname{University of Wisconsin -- Madison}, \orgaddress{\street{1010 Mound Street}, \city{Madison}, \postcode{53715}, \state{WI}, \country{USA}}}

\affil*[3]{\orgdiv{Industrial and Systems Engineering}, \orgname{University of Wisconsin -- Madison}, \orgaddress{\street{1513 University Avenue}, \city{Madison}, \postcode{53706}, \state{WI}, \country{USA}}}



\abstract{A key challenge in medical decision making is learning treatment policies for patients with limited observational data. This challenge is particularly evident in personalized healthcare decision-making, where models need to take into account the intricate relationships between patient characteristics, treatment options, and health outcomes. To address this, we introduce prescriptive networks (PNNs), shallow 0-1 neural networks trained with mixed integer programming that can be used with counterfactual estimation to optimize policies in medium data settings. These models offer greater interpretability than deep neural networks and can encode more complex policies than common models such as decision trees. We show that PNNs can outperform existing methods in both synthetic data experiments and in a case study of assigning treatments for postpartum hypertension. In particular, PNNs are shown to produce policies that could reduce peak blood pressure by 5.47 mm Hg (p=0.02) over existing clinical practice, and by 2 mm Hg (p=0.01) over the next best prescriptive modeling technique. Moreover PNNs were more likely than all other models to correctly identify clinically significant features while existing models relied on potentially dangerous features such as patient insurance information and race that could lead to bias in treatment. 
}




\maketitle

\section{Introduction}\label{intro}

Advances in machine learning and data science theory with the advent of Artificial Neural Networks (ANNs) have brought about substantial progress in learning complex tasks in the context of classification, prediction, and prescription. ANNs have become ubiquitous tools to address numerous classification tasks in domain areas such as natural language processing and image recognition. The burgeoning availability and accessibility of data in recent years have only improved the utility of ANNs. Despite the benefits of acquiring such data, challenges such as the high cost and time required to collect data, the presence of biased data, or incomplete data persist \citep{wiseman1972methodological, rimando2015data}. These challenges are particularly prevalent in areas of interest to the operations and management science community; for example, epidemiological research on the prescription of drugs or the affects of smoking necessitate obtaining limited data through observational studies as it is often unethical to obtain a large amount of patient data through randomized controlled trials \citep{jadad1998randomised, polubriaginof2017challenges}. In addition to limitations due to ethical concerns, studies involving a small subset of the population, such as treating postpartum birthing persons for hypertension, are a naturally data-scarce settings. Moreover, using data that may be biased or incomplete can lead to inaccurate or unreliable models \citep{gilbert2019epistemic}. The consequences of enacting policies derived from such models can be dire. In the context of medicine and healthcare, these could include lawsuits, increased costs of insurance, and adverse health effects to individuals targeted by the policy. This can be particularly problematic in public health, where bias in data for a specific demographic leads to discriminatory policies.

In data-rich settings, deep neural networks, i.e, networks with a large number of hidden layers, have taken center stage in learning modern complex tasks. However, current state-of-the-art optimization tools used to train ANNs, such as stochastic gradient descent (SGD), fall short when data availability is limitted \citep{forman2004learning, aminian2005application, shaikhina2017handling}. Furthermore, criticisms against the interpretability of deep neural networks are increasingly prevalent in domains such as medicine, public health, and social intervention where fairness and accountability are critical \citep{chakraborty2017interpretability, fan2021interpretability, zhang2021survey}. Recent techniques emerging from operations research literature have explored the use mixed-integer programming to address some of the limitations of ANNs trained with SGD \citep{toro2019training, kurtz2021efficient, thorbjarnarson2023optimal}. These works have established the potential of integer programming in data-limited setting. However, there is currently no consensus on modelling an appropriate loss function for the mixed-integer programming (MIP) formulations.

In this paper, we explore the problem of training shallow neural networks with limited data using MIP with a focus on personalized prescription problems. While prediction remains a rich field for learning tasks, decision makers in several fields are motivated to make personalized decisions tailored to individuals. These problems arise in domains such as personalized online advertising \citep{li2010contextual}, direct mail marketing campaigns \citep{hitsch2018heterogeneous}, and personalized medical treatment assignments \citep{flume2007cystic}. We provide empirical insight into designing personalized prescription policies in the medical domain where conducting randomized control trials are expensive, unethical, or otherwise infeasible. Specifically, we consider the tasks of deciding treatments for hypertension in postpartum birthing persons. Our methods are well-suited to solve such problems since the availability of data is limited, and the need for understanding which patient characteristics influence policy design is high. In addition to these experiments, in the appendix we provide supplementary experiments showing the use of our prescriptive networks for designing optimal policies to prescribing warfarin for patients susceptible to blood clots. Additionally, in the appendix, we produce empirical results demonstrating the use of our models for prediction using the MNIST digit recognition problem for the 2-digit case.


\subsection{Problem Description}

Machine learning is commonly used to solve prediction problems, which involve building models that can use data to predict future events or outcomes. These models learn from labeled data to identify patterns and relationships, which can then be applied to new, unseen data. Successfully solving prediction problems can provide valuable insights for decision-making in a variety of disciplines.

Alternately, prescription problems are a growing array of tasks in personalized decision-making. These problems arise when managers and decision-makers need to make general policy decisions that account for the heterogeneous responses to the policy among different individuals in the population. In such cases, prediction alone is not sufficient as individuals may respond to the same treatments differently. While adaptive or reinforcement learning can be used to address prescription problems \citep{mintz2020nonstationary,mintz2023behavioral,he2023model}, there are cases where neither are entirely appropriate – adaptive learning is a promising tool in online learning settings, and reinforcement learning is particularly useful in data-rich settings. However, given limited observational data, we aim to optimize over expected outcomes directly. In this paper, we consider the problem of prescribing the best treatment to an individual as a function of their characteristics such that the outcome of such a prescription is optimized. 

These problems are particularly challenging because they require us to determine not only what the best treatments are, but also why they are the best treatments. In other words, we need to understand the causal relationships between treatments and outcomes. Indeed, we use techniques from causal inference to define an objective that maximizes (or minimizes) the outcome by estimating the counterfactual information from historical observations. By doing so, we can gain a better understanding of the causal relationships between actions and outcomes, allowing us to make better policy decisions. For instance, when prescribing blood-pressure medication to postpartum patients, we may be interested in ensuring that the model identifies those features that attribute to a positive response to receiving treatment as well as those that attribute to a negative response.

\subsection{Contributions}

The overarching goal of this paper is to provide a structured framework to training ANNs using MIP techniques for prescription problems.

We make several methodological contributions to these problem domains.  In the realm of prescription problems, we design an optimization framework that uses MIP-based ANNs, which we refer to as Prescriptive Neural Networks (PNN). PNNs designs interpretable policies with the goal of optimizing population average outcomes given limited observational data. This problem is non-trivial because we need to account for the outcomes of unobserved treatments, as well as incorporate causal effects into our framework. To address this challenge, the PNN optimizes over-estimates of the conditional average treatment effects (CATE) to account for unobserved counterfactual outcomes.
To implement the framework into commercial solvers, we propose a novel mixed-integer linear objective function for the prescription problem that incorporates outcomes estimated by the Inverse Propensity Weighting (IPW) method, Direct Method (DM) or the Doubly Robust (DR) estimator. We also provide statistical guarantees for the proposed formulations that address both prediction and prescription problems. In particular, our analysis shows that these frameworks produce statistically consistent estimators in both cases, to the best of our knowledge this is one of the first derived statistical guarantees for ANNs trained with MIP.  

We validated our framework through a series of experiments on both simulated and real-world data. To validate PNNs, we benchmark our methods against state-of-the-art prescriptive methods and heterogeneous treatment effect estimation techniques with simulated experiments common in the prescription literature \citep{athey2016recursive} along with two case studies in personalized healthcare. The first case study entails designing prescriptive policies to treat postpartum patients against hypertension. 
The second case study, which we present in the appendix, involves incorporating prescription in personalized warfarin dosing. 
Our contributions also extend to assessing the interpretability of the PNN, and its ability to identify key features that impact in the models' decisions. By doing so, we derive a sense of our model's strengths and identify where it is most apt to use prescriptive neural networks over others. In particular, we show that PNNs are more likely to identify relevant clinical features for treatment decisions, unlike existing methods that may use biased surrogate features such as patient insurance information. 

While not in the main paper but in the appendix, we also present contribution to the use of MIP ANNs in the realm of prediction. We develop a novel MIP formulation of the negative log-likelihood loss (NLL) that incorporates a softmax activation function applied over the output layer of an ANN formulated by mixed-integer constraints. 
Existing literature that explores using MIP to train ANNs has considered a Support Vector Machine loss, minimizing empircal classification loss, and other linear and non-linear loss functions \cite{toro2019training, kurtz2021efficient, thorbjarnarson2023optimal}. The softmax function is an appealing activation function since it considers the output of the neural network as a probabilistic distribution over predicted output classes \citep{bridle1990probabilistic}. By incorporating this well-studied loss function from the literature into MIP-based ANNs, our work helps bridge the divide between MIP-based training methods and classical training methods. To evaluate the performance of our MIP-based ANN with the NLL objective, we conduct numerical experiments using the MNIST digit recognition dataset. We benchmark our performance against state-of-the-art stochastic gradient descent (SGD) 
based algorithms. These experiments are conducted over a variety of ANN architectures and hidden-layer activation functions.



\section{Literature Review}\label{litreview}

Our work contributes to several important streams of literature in machine learning and operations research.
The MIP formulations we develop in this paper builds upon the existing literature on using integer programming and constraint programming to train ANNs. Our work on applying this framework for personalized prescription problems parallel previous papers that consider the intersection of optimization and statistical inference. The first stream of literature pertains to training ANNs using MIPs. In the second stream, we review works that estimate heterogeneous causal effects in the context of learning treatments effects across subsets of the population with causal trees and forests. Along these lines, we also discuss non-tree methods used in statistical learning. Additionally, we present related work on learning prescriptive trees that optimize using MIP formulations.

Using integer programming to train neural networks has gained some popularity in recent years. \cite{toro2019training} use Constraint Programming, MIP, and a hybrid of the two to train Binarized Neural Networks (BNNs), a class of neural networks with the weights and activations restricted to the set \{-1, +1\}. 
To relax the binarized restriction on weights, \cite{thorbjarnarson2023optimal} propose a formulation to train Integer Neural Networks with different loss functions. 
\cite{kurtz2021efficient} developed a formulation for integer weights, and offer an iterative data splitting algorithm that uses the k-means classification method to train on subsets of data with similar activation patterns. For problems with continuous activation functions, \cite{serra2018bounding} introduce a mixed-integer linear programming (MILP) formulation for deep neural networks to count the number of linear regions represented by it. While the authors do not use the formulation to train neural networks, they utilize a systematic one-tree approach \citep{danna2007generating} to count integer solutions of the MILP. MIPs have also been used for already trained networks to provide adversarial samples that can improve network stability \citep{fischetti2017deep,anderson2020strong, tjeng2017evaluating}. Additionally, there has been research directed towards using ANNs to solve MIPs \citep{nair2020solving}.

 In the context of statistical learning based approaches for causal inference, 
 \cite{athey2016recursive} proposed causal tree estimators that use recursive partitions to estimate conditional average treatment effects. The causal tree structure uses the Inverse Propensity Weighting \citep{horvitz1952generalization} to estimate counterfactual outcomes. \cite{wager2018estimation} build on the concept of causal trees by extending estimation using random forests with a causal structure, and provide asymptotic confidence intervals for their work. \cite{powers2018some} further build on causal forests and introduce Pollinated Transformed Outcome forests for the treatment effect estimation problem in high dimensions. They additionally apply modifications to boosting and MARS (Multivariate Adaptive Regression Splines) to use in the context of structured causal problems. Along the lines of estimating counterfactual estimation, \cite{dudik2011doubly} provides a doubly robust framework that combines inverse propensity weighting methods with ``direct methods'', i.e, directly estimating counterfactual outcomes using learning models with a given sample. The same paper proposes methodologies for devising optimal treatment policies using the doubly robust framework. A study on using deep neural networks for semiparametric inference was conducted by \cite{farrell2021deep}, which apply theoretical results on treatment effect and expected welfare estimation.

 We now discuss tree-based approaches for personalized prescription problems. The works reviewed in this avenue of research build on the classification trees \citep{breiman1984cart}. A key motivation presented in literature for personalized trees is to provide a sense of interpretability to prescription problems. 
\cite{kallus2017recursive} proposes personalization trees with a mixed-integer optimization framework to learn to choose the optimal treatment from a discrete set of treatments. \cite{bertsimas2019optimal} modify this approach with by incorporating regularization for variance-reduction in the objective function of the problem. The authors use coordinate descent with multiple starts to solve the optimization problem numerically. 
A prescriptive tree approach presented by \cite{jo2021learning} builds on a MIP-based classification tree \citep{aghaei2021strong} formulated as a flow network. Both \cite{kallus2017recursive} and \cite{bertsimas2019optimal} provide structure for learning optimal trees without explicitly estimating counterfactual outcomes, while the MIP formulation detailed by \cite{jo2021learning} incorporates inverse propensity weighting methods, direct methods, or doubly robust methods in their optimization problem. \cite{amram2022optimal} extend the work of \cite{bertsimas2019optimal} to the problem of learning prescription policies using
estimated counterfactual information, and to problems with continuous-valued treatments. Finally, a new type of prescription trees is presented by \cite{bertsimas2022data}, which uses threshold-based methods to devise decision-rules that can be used in-tandem with the optimal policy trees \citep{amram2022optimal} under budget constraints.

\section{Mixed-Integer Programming Formulation of ANNs}\label{MIP-ANNs}

In this section we consider MIP formulations for ANNs in the prescriptive setting. However, in the appendix, we discuss how the formulation can be adapted to the predictive setting.  Consider a personalized decision-making problem where the goal is to assign the best treatment $t \in \mathcal{T}$ for an individual given their covariates $X \in \mathcal{X} \subseteq \mathbb{R}^\mathcal{F}$ such that the individual's potential outcome $Y(t) \in \mathcal{Y} \subseteq \mathbb{R}$ is either minimized or maximized over the treatment group. We assume both $\mathcal{X},\mathcal{Y}$ are compact sets and that $\mathcal{T}$ is a finite set. In this problem, decision-makers use observational data to design a personalized treatment policy $s : \mathcal{X} \mapsto \mathcal{T}$ that should then be able to prescribe an optimal treatment $s(X)$ for an individual based on their covariates $X$. In the case where all of the potential outcomes $\{Y_i(t)\}_{t\in \mathcal{T}}$ for an individual $i$ are known, the problem of learning an optimal policy $\hat{s}$ is a classification problem where the model classifies the patient to the treatment group that maximizes or minimizes individual's outcome. Recall that we are interested in the observational case when prescribing a patient every possible treatment to procure all potential outcomes can be impractical due to ethical concerns or resource constraints. In this problem, we observe an outcome $\{Y_i(t_i)\}$ for individual $i \in \{1,\ldots,n\}$ having received treatment $t_i$, but not their counterfactual outcomes $\{Y_i(t)\}_{t \in \mathcal{T} \setminus t_i}$. Our goal is to learn a policy $\hat{s}$ from $n$ independent and identically distributed observations of participants $\{(X_1, t_1, Y_1(t_1)), \ldots, (X_n, t_n, Y_n(t_n))\}$. 
%
%

We use the notation $[n]_m = \{m,m+1,...,n\}$ for any two integers $n>m$. 
For our problem, we will think of policies $s$ being parameterized by $\theta \in \Theta$, where $\Theta$ is a compact set, and our goal will be to find $\theta^*$ such that for any $X\in \mathcal{X}$, $s(X;\theta^*)$ is able to closely predict the optimal treatment. Our goal will be to estimate $\theta^*$ by finding a $\hat{\theta}$ that minimizes an empirical loss $\mathcal{L}_n: \mathcal{T}\times \mathcal{Y} \mapsto \mathbbm{R}$ that is $\hat{\theta} = \mathrm{argmin}_{\theta \in \Theta} \frac{1}{n}\sum_{i = 1}^n \mathcal{L}_n(s(X_i;\theta),Y_i)$. We assume $s$ is an ANN that can be expressed as a functional composition of $L$ vector valued functions that is $s(X;\theta) = h_L\circ h_{L-1} \circ ... \circ h_1 \circ h_0(x)$. Here each function $h_\ell(\cdot,\theta_\ell)$ is referred to as a layer of the neural network, we assume that $h_0:\mathcal{X} \mapsto \mathbbm{R}^K$, $h_L:\mathbbm{R}^K \mapsto \mathcal{T}$, and $h_\ell: \mathbbm{R}^K \mapsto \mathbbm{R}^K$, we denote each component of $h_\ell$ as $h_{k,\ell}$ for $\ell \in [L-1]_0$ and $k \in [K]_1$, and each component of $h_L$ as $h_{t,L}$ for $t \in \mathcal{T}$. We refer to these components as units or neurons. We refer to the index $\ell$ as the layer number of the ANN, and the index $k$ as the unit number. The layer corresponding to the index $L$ is referred to as the output layer, and all other indices are referred to as hidden layers. Each unit has the functional form: $h_{k,0} = \sigma(\alpha_{k,0}^\top X + \beta_{k,0}) $, $h_{k,\ell} = \sigma(\alpha_{k,\ell}^\top h_{\ell-1} + \beta_{k,\ell})$ for $\ell \in [L-1]_1$, $h_{t,L} = \varphi(\alpha_{t,L}^\top h_{L-1} + \beta_{t,L})$, where $\sigma:\mathbbm{R}\mapsto \mathbbm{R}$ is a non-linear function applied over the hidden layers $\ell = 0,...,L-1$ called the activation function, $\varphi:\mathbbm{R}\mapsto \mathbbm{R}$ is a potentially different non-linear function applied over the output layer $L$, and $\alpha$ and $\beta$ are the weights and biases matrices respectively where $(\alpha,\beta) = \theta$. In our work, we consider $\sigma$ to be the binary activation and we assume for any input $z \in \mathbbm{R}$, $\sigma(z) = \mathbbm{1}[z \geq 0]$, where $\mathbbm{1}$ is the indicator function. The notation $\alpha_{a,b,\ell}$ indicates the weight being applied to unit index $a$ in layer $\ell -1$ for the evaluation of unit index $b$ in layer $\ell$. Likewise, $\beta_{k,\ell}$ indicates the bias associated with unit index $k$ in layer $\ell$ of the network.

We show how the above training problem and model can be formulated as a MIP. The key to our reformulation is the introduction of decision variables $h_{i,k,\ell}$ which correspond to the unit output of unit $k$ in layer $\ell$ when the ANN is evaluated at data point with index $i$. Having these decision variables be data point dependant, and ensuring that $\alpha_{k^\prime,k,\ell},\beta_{k,\ell}$ are the same across all data points forms the backbone of our formulation. Thus, if $X_{i,d}$ denotes the $d^{th}$ feature value of individual $i$, the general form of the optimization problem is:

\begin{subequations}
\begin{align}
\hat{\theta}_n = \argmin_{\theta \in \Theta} & \frac{1}{n}\sum_{i=1}^n\sum_{t \in T} \mathcal{L}_n(h_{i,j,t},y_{i,j})  & \label{loss} \\
\textrm{subject to } \notag \\
 & h_{i,k,0} = \sigma(\sum_{d \in \mathcal{F}}\alpha_{d,k,0}X_{i,d} + \beta_{k,0}), \notag \\ & \forall \ k,i \in [K]_1 \times [n]_1, \label{l_0 output_const} \\
 & h_{i,k,\ell} = \sigma(\sum_{k'=1}^K\alpha_{k^\prime,k,\ell}h_{i,k^\prime,\ell-1} + \beta_{k,\ell}), \notag \\ & \forall \ \ell,k,i \in [L-1]_1 \times [K]_1 \times [n]_1, \label{l output_const} \\ 
 & h_{i,t,L} = \varphi(\sum_{k'=1}^K\alpha_{k^\prime,j,L}h_{i,k^\prime,L-1} + \beta_{t,L}), \notag \\ & \forall j,i \in \mathcal{T} \times [n]_1. \label{L output_const} 
\end{align}
\label{eq:general_form}
\end{subequations}
In this section we will discuss how the constraints of \eqref{eq:general_form} can be represented using a MIP formulation. We will also discuss how different forms of common causal inference loss functions $\mathcal{L}_n$ can be formulated using MIP constraints. We will conclude by proving that the resulting formulation provides consistent estimates of the optimal policy. That is, that as $n \rightarrow \infty$, $\hat{\theta}_n \overset{p}{\rightarrow} \theta^*$. This property is crucial, since it indicates that as decision makers gather more data our technique will generate more effective policies. Moreover, it is not clear if many ANN architectures trained using stochastic gradient based approaches exhibit this property in general.

\subsection{Binary Activations \label{Binary MIP}}

In this section, we present a MIP reformulation of the constraints of \eqref{eq:general_form} for binary activations in the hidden layers that can be solved using commercial solvers. Our formulation can be seen as an extension of prior MIP formulations for ANN training by using continuous instead of integer valued model weights \citep{toro2019training}. Specifically, we relax the bounds such that $\alpha_{d,k,0}, \alpha_{k^\prime,k,l}, \alpha_{k^\prime,j,L} \in [\alpha^{L}, \alpha^{U}]$. We rewrite Constraints \eqref{l_0 output_const},\eqref{l output_const},\eqref{L output_const} for a single unit as $h_{i,k,0} = \mathbbm{1}[ \sum_{d \in \mathcal{F}} \alpha_{d,k,0}X_{i,d} + \beta_{k,X0} \geq 0]$, $h_{i,k,\ell} = \mathbbm{1}[ \sum_{k^\prime =1}^K \alpha_{k,k^\prime,\ell}h_{i,k^\prime,\ell-1} + \beta_{k,\ell} \geq 0]$, $h_{i,j,L} = \sum_{k^\prime =1}^K \alpha_{k^\prime,j,L}h_{i,k^\prime,L-1} + \beta_{j,\ell} $ respectively. Note that for all layers that are not the input layer, the above constraints contain bi-linear products which make this formulation challenging to solve. We propose the following reformulation:

\begin{proposition}
 In the binary activation case, the Constraints \eqref{l_0 output_const},\eqref{l output_const},\eqref{L output_const} can be reformulated as a set of MIP constraints. Specifically for all $k\in [K]_1, i\in[n]_1$ Constraint \eqref{l_0 output_const} can be reformulated as:
 \begin{align}
 & \sum_{d \in \mathcal{F}} (\alpha_{d,k,0}X_{i,d}) + \beta_{k,0} \le Mh_{i,k,0}, \label{h0_define C1} \\
 & \sum_{d \in \mathcal{F}} (\alpha_{d,k,0}X_{i,d}) + \beta_{k,0} \ge \epsilon + (-M-\epsilon)(1-h_{i,k,0}), \label{h0_define C2} 
 \end{align}
 For all, $\ell \in [L-1]_1,k\in[K]_1,i\in[n]_1$ Constraints \eqref{l output_const} can be reformulated as:
\begin{align}
 & \sum_{k^\prime=1}^{K} (z_{i,k^\prime,k,\ell}) + \beta_{k,\ell} \le Mh_{i,k,\ell}, \label{hl_define C1} \\
 & \sum_{k^\prime=1}^{K} (z_{i,k^\prime,k,\ell}) + \beta_{k,\ell} \ge \epsilon + (-M-\epsilon)(1-h_{i,k,\ell}), \label{hl_define C2} \\
 & z_{i,k^\prime,k,\ell} \le \alpha_{k^\prime,k,\ell} + M(1-h_{i,k^\prime,\ell-1}), \ \forall \ k^\prime \in [K]_1, \label{zl_define C1}\\
 & z_{i,k^\prime,k,\ell} \ge \alpha_{k^\prime,k,\ell} -M(1-h_{i,k^\prime,\ell-1}), \ \forall \ k^\prime \in [K]_1, \label{zl_define C2}\\ 
 & -Mh_{i,k^\prime,\ell-1} \le z_{i,k^\prime,k,\ell} \le Mh_{i,k^\prime,\ell-1}, \ \forall \ k^\prime \in [K]_1, \label{zl_define C3}
\end{align}
And for all $t\in \mathcal{T},i \in [n]_1$, Constraints \eqref{L output_const} can be reformulated as:
\begin{align}
 & \sum_{k^\prime=1}^{K} (z_{i,k^\prime,k,L}) + \beta_{k,L} \le Mh_{i,k,L}, \label{hL_define C1} \\
 & \sum_{k^\prime=1}^{K} (z_{i,k^\prime,k,L}) + \beta_{k,L} \ge \epsilon + (-M-\epsilon)(1-h_{i,k,L}) \label{hL_define C2}, \\
 & z_{i,k^\prime,t,L} \le \alpha_{k^\prime,t,L} + M(1-h_{i,k^\prime,L-1}), \ \forall \ k^\prime \in [K]_1, \label{zL_define C1} \\
 & z_{i,k^\prime,t,L} \ge \alpha_{k^\prime,t,L} - M(1-h_{i,k^\prime,L-1}), \ \forall \ k^\prime \in [K]_1, \label{zL_define C2} \\
 & -Mh_{i,k^\prime,L-1} \le z_{i,k^\prime,t,L} \le Mh_{i,k^\prime,L-1}, \ \forall \ k^\prime \in [K]_1. \label{zL_define C3} 
\end{align}
\label{prop:reform_bin}
 \end{proposition}

Constraints \eqref{h0_define C1} and \eqref{h0_define C2} are big-M constraints that, when combined with the integrality of $h_{i,k,0}$, impose the output of the first hidden layer to be 1 if the linear combination of the units of the input vector summed with the bias term is greater than or equal to some small constant $\epsilon > 0$, and 0 otherwise. To reformulate Constraints \eqref{l output_const}, we introduce an auxiliary variable $z_{i,k^\prime,k,\ell} = \alpha_{k^\prime,k,\ell}h_{i,k^\prime,\ell-1}$. Constraints \eqref{hl_define C1} and \eqref{hl_define C2} are then similar to Constraints \eqref{h0_define C1} and \eqref{h0_define C2}, and define the output of the remaining hidden layers $\ell \in [L-1]_{1}$. We leverage the fact that the bi-linear term is a product of a continuous and binary variable for Constraints \eqref{zl_define C1}, \eqref{zl_define C2}, and \eqref{zl_define C3};  $z_{i,k^\prime,k,\ell}=0$ when $h_{i,k^\prime,\ell-1}=0$ or $z_{i,k^\prime,k,\ell}=\alpha_{k^\prime,k,\ell}$ when $h_{i,k^\prime,\ell-1}=1$. Similarly for the definition of Constraints \eqref{hL_define C1}-\eqref{zL_define C3}, which define the binary activated output of the output later, and ensures $z_{i,k^\prime,t,L} = \alpha_{k^\prime,t,L}h_{i,k^\prime,L-1}$, the bi-linear terms associated with the output layer.

A full proof of this proposition can be found in the Appendix \ref{proofsappendix}. However, we will present a sketch here. The main techniques for this proof rely on first using a big-M formulation for disjunctive constraints \citep[see, for example][]{wolsey1999integer, conforti2014integer} to model the binary activation. Then bi-linear terms are reformulated using the techniques applied to products of binary and continuous variables.

\subsection{Loss Functions for Prescription \label{PNN}}

In this section we design MIP losses for ANNs in the context of prescription problems. 


The key challenge of using observational data is the lack of counterfactual information. Without the counterfactual, it would be impossible to identify the best treatment in expectation for a given individual. Thus, we consider this problem  as a conditionally randomized experiment, by making the following key assumptions \citep{hernan2010causal}. 

\begin{assumption} \label{ass:conditional exchangeability}
    For every treatment $t \in \mathcal{T}$, the outcome $Y(t)$ is independent of observed treatment, $T$, given the covariates $X$. That is, conditional exchangeability, $Y(t) \perp\!\!\!\!\perp T|X$, holds.
\end{assumption}

Exchangeability in this context refers to the case where for any two treatments $t,t^\prime \in \mathcal{T}$, the group that receive a treatment $t$ would observe the same outcome distribution as the group that receive a different treatment $t^\prime \neq t$ had their treatments been \textit{exchanged}. Here, we specifically assume that \textit{conditional exchangeability} $Y(t) \perp\!\!\!\!\perp T|X$ holds as, a sufficient assumption to conceptualize observational studies as \textit{conditionally} randomized experiments. The key for this assumption to hold is that the treatment groups are conditionally exchangeable on $X$ since we consider only information from $X$ to prescribe treatments. Indeed this is a sufficiency condition. That is, the counterfactual outcome and the observed treatments are independent conditioned on the covariates since only the patient information is used to assign the treatment.

\begin{assumption}\label{ass:positivity}
For every $t \in \mathcal{T}$, the probability that an individual $i \in [n]_1$ with covariates $X_i$ is assigned $t$ is strictly positive. That is positivity, $\mathbb{P}(t_i=t|X_i) > 0$, holds.
\end{assumption}

Both assumptions are standard in literature and are known to hold in practice. For example, in the domain of personalized medicine, Assumption \ref{ass:conditional exchangeability} holds considering that only the patient's covariates are available in an observational study. Particularly, patients features are the only information used for most medical guidelines used prescribes a treatment for their patient. Assumption \ref{ass:positivity} holds considering that practitioners often use intuition developed over clinical or medical training to often prescribe treatments that deviate from treatment protocol. That is, patients need not be prescribed the same treatment despite having the same characteristics. Moreover, the assumption is tenable given that practitioners may disagree on treatment plans.

We now discuss methods from the literature that estimate the average causal effect of observational studies that satisfy the above conditions. The first is the Inverse Propensity Weighting (IPW) method, which was introduced to estimate the expected value of counterfactual treatment outcomes and average treatment effect \citep{horvitz1952generalization}. The method creates a \textit{pseudo-population} where every individual in the data is given all of the possible treatments, simulating a randomized controlled trial. The IPW method first estimates the propensity scores for each individual $i$, $\mathbb{\hat{P}}[T=t_i|X=X_i]$, and then re-weights their outcome by taking its inverse. Given a policy $s$, we leverage the conditional exchangeability assumption to estimate the performance of $s$ as a measure of the average outcome:
\begin{align} 
   \hat{\pi}^{IPW}(s) := \frac{1}{n}\sum_{i=1}^{n}\frac{\mathbbm{1}(s(X_i)=t_i)}{\mathbb{\hat{P}}[T=t_i|X=X_i]}Y_i 
\end{align}
If the estimated propensity score converges almost surely to the true propensity score, the IPW estimator is unbiased but suffers from high variance for cases where $\mathbb{\hat{P}}[T=t_i|X=X_i]$ is small \citep{hernan2010causal}.

The second method we are interested in is the so-called Direct Method (DM) to directly estimate the average causal effect of observational studies. DM is an adaption of the Regress-and-Compare (RC) approach for estimating counterfactual outcomes for optimal treatment assignment. By the RC framework, we first partition the dataset by treatment, then for $X \in \mathcal{X}$ learn a model $\hat\mu_{t}(X)$ for $\mathbb{E}[Y|X,T=t]$ using the subpopulation that was assigned treatment $t$, and finally estimate the counterfactual outcomes $\hat{Y}(t)$ as $\hat\mu_{t}(X)$. In the case where we are maximizing over the outcomes, the RC approach would then take $\argmax_{t \in T}\hat\mu_{t}(X)$ as the prescription for an individual \citep{qian2011performance}. The DM approach modifies this approach to estimate average outcome of a $s$:
\begin{align}
    \hat{\pi}^{DM}(s) := \frac{1}{n}\sum_{i=1}^{n}\hat\mu_{s(X_i)}(X_i)
\end{align}
However, we note that the performance of RC relies heavily on the quality of the estimate $\hat\mu_t(X), t \in T$ \citep{beygelzimer2009offset}. As is the case with IPW, if the estimated outcome $\hat\mu_t(X)$ is unbiased, then the DM estimator is unbiased as well.

Lastly, we discuss the Doubly Robust (DR) estimation method as proposed by \cite{dudik2011doubly} and adapted by \cite{athey2016recursive}. To counteract the reliance on one estimator being unbiased, this method combines the IPW and DM approaches to reduce the errors individually brought in by the two estimators; the DR estimator of the outcome appropriately combines the estimates $\hat\mu_t(X)$ and $\mathbb{\hat{P}}[T=t_i|X=X_i]$. It can then estimate an individual's outcome using the doubly robust framework as:
\begin{align}
    \hat\psi_{t}(z_i) = \hat\mu_{s(X_i)} (X_i) + \frac{\mathbbm{1}(s_0(X_i)=t_i)}{\mathbb{\hat{P}}[T=t_i|X=X_i]}(Y_i - \hat\mu_{t_i} (X_i))
\end{align} 
where $z_i = (x_i,t_i,Y_i)$ is the tuple of the individual's covariates, treatment, and outcome from the data, and we let $\mathcal{Z} := \mathcal{X} \times \mathcal{T} \times \mathcal{Y}$. Under Assumptions \ref{ass:conditional exchangeability} and \ref{ass:positivity}, the bias of the doubly robust estimator is small if $\hat\mu(X)$ is close to $\mu(X)$ or $\hat{\mathbb{P}}[T=t_i|X=X_i]$ is close $\mathbb{P}[T=t_i|X=X_i]$. Moreover, the bias $\mathbb{E}[\mathbb{E}[\hat{Y}_{DR}] - \mathbb{E}[Y]]$ of the estimated outcomes $\mathbb{E}[\hat{Y}]$ is asymptotically zero when either the IPW model or the DM model is consistent. This doubly robust estimator benefits from second-order bias, so estimating $\hat\mu(X)$ and $\hat{\mathbb{P}}[T=t_i|X=X_i]$ with machine learning estimators can lead to a smaller bias than standard parametric models \citep{hernan2010causal}.

Given the doubly robust estimator $\hat\psi_{t}$, \cite{jo2021learning} learn the optimal policy by solving the optimization problem $\argmax_{t \in T}\hat\psi_{t}(z)$ in a maximization problem. Alternately, one may be interested in optimizing over the treatment effects rather than the estimated outcomes directly. \cite{farrell2021deep} present a one such approach for evaluating a policy $s^\prime$ against a baseline policy $s_0$. In the dichotomous treatment case, the function $\hat\pi(s^\prime,s_0)$ can be used to evaluate $s^\prime$ against the baseline $s_0$:
\begin{align}
\hat\pi(s^\prime,s_0) & = \mathbb{E}_{n}\big[[s^\prime(X_i)-s_0(X_i)]\hat\psi_{1}(z_i) - [s^\prime(X_i)-s_0(X_i)]\hat\psi_{0}(z_i)\big] \\
  & = \mathbb{E}_{n}[(s^\prime(X_i)-s_0(X_i))(\hat\psi_{1}(z_i) - \hat\psi_{0}(z_i)] \\
  & = \mathbb{E}_{n}[(s^\prime(X_i)-s_0(X_i))\hat\tau(X_i)]
\end{align}
where $\hat\tau(X_i) = \mathbb{E}_{n}[\hat\psi_{1}(z_i) - \hat\psi_{0}(z_i)]$ is the estimator for the average treatment effect. For cases where we consider more than two treatments, we can then define the following:

\begin{align}
  \hat\pi(s) & = \mathbb{E}_{n}[\sum_{t \in T}\mathbbm{1}\{s(X_i)=t\}\hat\psi_{t}(z_i)] \label{pi} \\
  \hat\pi(s^\prime,s_0) & = \mathbb{E}_{n}[\sum_{t \in T}\mathbbm{1}\{s^\prime(X_i)=t\}\hat\psi_{t}(z_i) -\sum_{t \in T}\mathbbm{1}\{s_0(X_i)=t\}\hat\psi_{t}(z_i)] \label{compare pi}
\end{align}

By using $\hat{\pi}$ as our desired estimator, we explicitly consider the average treatment while evaluating the performance of a policy. Such a formulation also allows us to directly optimize against a known baseline policy. We now propose a linear objective function derived from Equations \ref{pi} and \ref{compare pi} through Proposition \ref{prop:pi obj}:

\begin{proposition}
Under Assumption \ref{ass:conditional exchangeability} and \ref{ass:positivity}, and in the case where larger outcomes are preferred, the objective of the optimal prescription problem that optimizes over the CATE with DR estimators can be formulated as
\begin{align}
 \max \frac{1}{n}\sum_{i=1}^{n}\sum_{t \in T}h_{i,t,L}\hat\psi(z_i)_{t} \label{cate obj general}
\end{align}

when we do not have a baseline policy, and

\begin{align}
 \max \frac{1}{n}\sum_{i=1}^{n}\sum_{t \in T}(h_{i,t,L}-s_0(x_{i})_t)\hat\psi(z_i)_{t} \label{cate compare obj}
\end{align}

when we have a baseline policy $s_0$.
\label{prop:pi obj}
\end{proposition}

The proof of Proposition \ref{prop:pi obj} follow directly from the derivations detailed in \cite{farrell2021deep}. The proof can be found in the Appendix \ref{proofsappendix} for the sake of completeness. 

Clearly, Proposition \ref{prop:pi obj} can be adapted to formulate the objective with the IPW and DM estimators as well. From our findings in Section \ref{Binary MIP} along with the above reformulated objectives, our main result extends to the formulation of Prescriptive Neural Networks. Matching the notation of this proposition to that presented in \eqref{eq:general_form} we can think of the loss function in this case as $\mathcal{L}_n(s(X_i),Y_i) = - \sum_{t \in \mathcal{T}} h_{i,t,L}\hat{\psi}_n(z_i)_t  $

We note the need to include one additional constraint to the MIP formulation that necessitates only one treatment to be assigned for each individual: 
\begin{align}
    \sum_{t \in \mathcal{T}} h_{i,t,L} = 1, \ \forall \ i \in [n]_1 \label{strictly one treatment}
\end{align}

\subsection{Consistency of Prescriptive Networks}
An advantage of policies calculated using our MIP formulation is that they can be shown to produce consistent prescriptive policies. In general, to show consistency of estimates derived from solving an optimization problem one must show the convergence of the minimizers of a sequence of stochastic optimization problems to the minimizer of a target optimization problem. This is not trivial since it requires showing that the value functions of these optimization problems properly converge to the value function of the target problem. For instance point-wise convergence of these functions is insufficient to guarantee this property \citep{rockafellar2009variational}. Thus our focus will be to show that in the case of MIP trained prescriptive networks, we obtain a necessary form of convergence that yields consistency.  To obtain this property we will have to make one simplifying assumption on the parameters of the input layer of the network.

\begin{assumption}
\label{ass:input_assump}
    Parameters $\alpha_{d,k,0}$ come from a finite set $\mathcal{A}$.
\end{assumption}

Note that this assumption is only required on the parameters of the input layer and not any of the subsequent layers. Moreover, this assumption is not too restrictive since we are already assuming all parameters come from a compact set and thus we can think of $\mathcal{A}$ as an arbitrarily fine grid of small width. In our computational experiments, we did not place this integrality restriction on our parameters and found no reduction in performance. In addition to this assumption we need to make another standard assumption when considering our causal inference based loss.

\begin{assumption}
\label{ass:consistancy_assump}
    Estimators $\hat{\mu}_n,\mathbbm{\hat{P}}_n,\hat{\psi}_n$ are consistent. That is, as $n \rightarrow \infty$, $\hat{\mu}_n  \overset{p}{\rightarrow} \mu,\mathbbm{\hat{P}}  \overset{p}{\rightarrow} \mathbbm{P},\hat{\psi}_n  \overset{p}{\rightarrow} \psi$, where   $\mu,P,\psi$ are the true values of the direct, indirect, and doubly unbiased estimators as evaluated with the population parameters.
\end{assumption}

This assumption is standard when analyzing prescriptive analytics \citep{jo2021learning,kallus2017recursive}. Essentially we are assuming that the methods employed for counterfactual estimation are statistically consistent, since if this is not the case then it would be impossible to learn a consistent policy.  

With these assumptions in mind we will define our target optimization problem as $\Theta^* = \argmin_{\theta \in \Theta} \mathbb{E}[ \sum_{t\in \mathcal{T}} \mathbbm{1}\{s(X,\theta) = t\}\psi(Z)]$, where $Z \in \mathcal{Z}$, and the expectation is taken with respect to the true probability law of the population, which we will call $\mathbb{P}_0$. Note that, $\Theta^* \subset \Theta$ is a set and not a single parameter since there could be multiple paramaeterizations for the optimal network structure. Without loss of generality, we are considering the case where counterfactual estimation is given by the doubly robust estimate, though in principle our arguments will also hold for the direct and indirect estimates. Our main result is showing the following theorem:
\begin{theorem}
\label{thm:main_consistant}
    Given Assumptions \ref{ass:conditional exchangeability}-\ref{ass:consistancy_assump} and all parameter and modeling assumptions. For all $\epsilon>0$, $\hat{\theta}_n$ as computed by \eqref{eq:general_form} satisfies:
    \begin{equation}
        \lim_{n \rightarrow \infty} \mathbb{P}_0(\text{dist}(\hat{\theta}_n, \Theta^*) > \epsilon) = 0
    \end{equation}
Where for a point $x$ and set $S$ $\text{dist}(x,S):= \min_{y \in S} \|x- y\|$.
\end{theorem}

To show this main result we will need to prove several intermediate results on the structure of the value functions of the target optimization problem and the training problem. For notational brevity let $\varphi(\theta,Z) = \sum_{t\in \mathcal{T}} \mathbbm{1}\{s(X,\theta) = t\}\psi(Z)$ and $\varphi_n(\theta,Z) = \sum_{t\in \mathcal{T}} \mathbbm{1}\{s(X,\theta) = t\}\psi_n(Z)$. Our first result will establish the continuity of these functions.

\begin{proposition}
\label{prop:lsc}
    Under Assumption \ref{ass:input_assump}, the functions $\varphi(\theta,Z),\varphi_n(\theta,Z)$ are lower semicontinuous for all pairs $(\theta,Z) \in \Theta \times \mathcal{Z}$.
\end{proposition}

The full proof of this result can be found in the appendix. However, this proposition will follow from the construction of $\varphi,\varphi_n$ as value functions of parametric MIP problems with the parameters appearing as affine terms in the constraints \citep{ralphs2014value}. Lower semi continuity is a crucial property since it ensures the closure of the epigraph of the value function, and is one of the conditions necessary to establish proper convergence of minima. Next we establish a point-wise condition for the convergence of the empirical loss function to the target loss function.

\begin{proposition}
\label{prop:pointwise_lln}
    Given Assumptions \ref{ass:conditional exchangeability}-\ref{ass:consistancy_assump}. For any $\epsilon >0$ define $U_\epsilon(\theta):= \{\theta' \in \Theta: \|\theta' -\theta\| < \epsilon\}$. Then consider the following quantities:
    \begin{align}
        H_n^\epsilon(\theta,Z) = \frac{1}{n}\sum_{i=1}^n \inf_{\theta' \in U_\epsilon(\theta)} \varphi_n(\theta',z_i) \\
        H^\epsilon(\theta) = \mathbb{E}[ \inf_{\theta' \in U_\epsilon(\theta)} \varphi(\theta',Z) ]
    \end{align}
    Then for all $\theta \in \Theta, \epsilon >0$, $H^\epsilon(\theta) > -\infty$ and  for any $n \geq 1$ both $H_n^\epsilon(\theta,Z), H^\epsilon(\theta)$ exist and are well defined. Moreover, $H_n^\epsilon(\theta,\cdot) \overset{l-prob}{\rightarrow} H^\epsilon(\theta)$, where the relation $ H_n \overset{l-prob}{\rightarrow} H$ implies the sequence $H_n$ is a lower semicontinuous approximation in probability to the target $H$ as per Definition 4.1 in \citep{vogel2003continuous}.
\end{proposition}

This proposition establishes a point-wise functional law of large numbers for our MIP ANN models. Note that this result is stronger than a standard law of large numbers since we are required to control the convergence behavior of our value functions within an open neighborhood of the parameters. While a full proof of the proposition will be in the appendix here we will provide a brief sketch. Since we are assuming the data $Z$ is i.i.d, the two key arguments needed to establish this relation are showing that the quantities in the proposition satisfy Kolmogorov's Law of Large Numbers, and  then that a strong enough convergence is achieved to apply Slutsky's Theorem \citep{keener2010theoretical}.  Note that this is still a point-wise result since it holds for each individual $\theta$, and it is not by itself sufficient to prove our main result. Thus we require one additional intermediate result before proceeding to prove our main theorem.

\begin{proposition}
\label{prop:val_func_lsc}
Given Assumptions \ref{ass:conditional exchangeability}-\ref{ass:consistancy_assump}. Let $\Phi,\Phi_n:\theta \mapsto \mathbb{R}$ be defined as $\Phi(\bar{\theta}) := \min_{\theta \in \Theta} \big\{ \mathbb{E}[ \varphi(\theta,Z) ] : \theta = \bar{\theta}  \big\}$ and $\Phi_n(\bar{\theta}) := \min_{\theta \in \Theta} \big\{ \frac{1}{n}\sum_{i = 1}^n \varphi_n(\theta,z_i) : \theta = \bar{\theta}  \big\}$  respectively. Then $\Phi_n \underset{\Theta }{\overset{l-prob}{\rightarrow} } \Phi$, where the operator indicates that $\Phi_n$ are a lower semi-continuous approximation in probability to $\Phi$ across the entire parameter set $\Theta$.
\end{proposition}
Note that this is a stronger global result across the entire parameter set $\Theta$ than the point-wise result in Proposition \ref{prop:pointwise_lln}. Moreover, note that the objective $\frac{1}{n}\sum_{i = 1}^n \varphi_n(\theta,z_i)$ is equivalent to the objective function in \eqref{loss} meaning $\Phi(\theta)$ can  be thought of as the value function of the training problem and $\Phi$ as the value function of the target problem. This proposition implies that as we collect additional data, we have the proper form of epi-convergence between our empirical and traget problems. The full proof of this proposition is in the appendix, and relies on using the results of Proposition \ref{prop:pointwise_lln} in conjunction with existing results from the convergence of stochastic optimization problems \citep{vogel2003continuous}. Combining this final result with properties of convergence of minima of stochastic optimization problems yields the main result of Theorem \ref{thm:main_consistant}.

\section{Numerical Experiments on Prescriptive Problems}\label{prescriptive experiments}

In this section, we evaluate the performance of the PNN in numerical experiments. We use experimental data from three settings:  a synthetic dataset commonly used in literature introduced by \cite{athey2016recursive}, real-world data from a study on postpartum hypertension \citep{mukhtarova2021evaluation}, and a dataset for personalized warfarin dosing collected by the International Warfarin Pharmacogentic Consortium \citep{international2009estimation}. We benchmark our methods against four state of the art models. These include prescriptive tree models from \cite{kallus2017recursive}, \cite{bertsimas2019optimal}, and \cite{jo2021learning} along with the causal forest approach from \cite{wager2018estimation}. For brevity we refer to these methods as B-PT, K-PT, J-PT, and CF respectively.

Prescriptive trees are mixed-integer optimization frameworks that learn policies for prescription problems that can be encoded as a binary decision trees. The notation used to describe each model follow from \cite{jo2021learning}. A limitation shared by all prescriptive tree models is the need for binary valued features as input parameters in the optimization framework. We provide a description of the prescriptive tree methods below.

\textbf{K-PT:} This model introduced by \cite{kallus2017recursive} is one of the first MIP models used to learn prescriptive trees.
The formulation relies on producing trees by recursively partitioning the covariates such that no feature falls under two partitions. Formally, given that $\mathcal{L}$ represents the set of leaf nodes and $\mathcal{X}_l$ denote the subset of covariate values for the data points collected in leaf $l \in \mathcal{L}$, the model creates partitions $\mathcal{X} = \cup_{l \in \mathcal{L}} \mathcal{X}_l$ such that $\mathcal{X}_l \cap \mathcal{X}_{l^\prime} = \emptyset, \ l \neq l^\prime$. A key assumption made to support the theoretical guarantees of the K-PT formulation is that depth of the trees are sufficiently deep such that the treatment assignment in the data and the covariates are independent at each leaf node. With respect to the experiments detailed in this section, our analysis is conducted on depth-1 and depth-2 trees, considering that the optimality gap of the K-PT model scales poorly with size of the tree (see Appendix \ref{completedata}).

\textbf{B-PT:} \cite{bertsimas2019optimal} address some limitations of the K-PT model and provide a modified MIP objective that regularizes against high variance in the outcomes of the data points conditioned on treatment assigned in each leaf. In particular, the authors include the term $ \mu \sum_{i=1}^{n}{\big(Y_i - \hat{Y}_{\mathcal{X}_{l(i)}:l(i)\in\mathcal{L}}(t_i)\big)}^{2} $ to the K-PT objective where $\mu \geq 0$ is an appropriately tuned hyperparameter. In our experiments, we found that the hyperparamter $\mu = 0.5$ yielded strong results for the B-PT formulation, a consistent result with experiments conducted in \cite{bertsimas2019optimal}. Again, our model is benchmarked against depth-1 and depth-2 B-PT models since these models scale poorly with deeper trres. 

\textbf{J-PT:} To address the need for the assumptions on the depth of the perscriptive tree, \cite{jo2021learning} directly optimize the outcomes estimated with the IPW, DM or DR methods. The formulation proposed in their work follows as an adaptation of the weighted classification tree formulation introduced by \cite{aghaei2021strong} for the prescriptive setting. The J-PT is modelled as a flow graph with edges between a source node and the root of a binary tree, and the leaves of the tree and a sink node. We analyze experimental results for depth 1 and depth 2 trees to maintain consistency with the other prescriptive trees. 

We use the CF as another model to benchmark our work in our numerical experiments. A description of the CF is given below.

\textbf{CF:} We benchmark against an adapted causal forest model proposed by \cite{wager2018estimation}. The original work estimates heterogeneous treatment effects using nonparametric random forests.
CF estimates the CATE by recursively splitting the feature space such that training examples are close with respect to their treatments and outcomes, where the measure of closeness is identified by data points falling in the same leaf of the tree. Since CF does not explicitly prescribe treatments, we adapt the original method to use the sign of the estimated treatment effect for each individual to assign treatments. In experiments with multiple treatment groups, we construct a pipeline to estimate the pairwise individual treatment effect and assign those treatments which yield the largest outcome for each individual. Experiments on the CF model use the EconML package \citep{econml} with their default arguments, which builds 100 trees in the forest.

All experiments were run on a high throughput computing cluster systems allocated with 4 CPU Cores, 32GB of RAM, and 4GB of Disk Space, with Python version 3.9.5 \citep{10.5555/1593511} and Gurobi Optimizer version 9.1.2 \citep{gurobi}. 

\subsection{Experimental Setup}\label{setup}

We discuss the nature of the datasets used in our experiments in this subsection. Since J-PT and PNN optimize over doubly robust objectives, we also provide insight into estimating the propensity scores and counterfactual outcomes for each experiment.

\subsubsection{Simulated Data}\label{setup:simulated}

The experiments with the simulated data were adapted from \cite{athey2016recursive} for data simulation, and \cite{jo2021learning} for the experiment design. Given that we simulate the counterfactual outcomes, a synthetic dataset allows us to evaluate learned policies against the ground truth. We considered three models of varying feature sizes $\mathcal{F}$. As given by Equation \ref{synthetic outcome}, the potential outcome for each individual in the simulation $Y_i:\mathbbm{R}^{\mathcal{F}}\times \{0,1\} \mapsto \mathbbm{R}$ is a function of their covariates $X_{i} \sim \mathcal{N}(\boldsymbol{0},\boldsymbol{1}) \in \mathbbm{R}^{\mathcal{F}}$, where $\boldsymbol{0}$ and $\boldsymbol{1}$ are $\mathcal{F}$-dimensional vectors, and their assigned treatment $t_i \in \{0,1\}$:
\begin{align}
    Y_{i}(t) = \eta(X_i) + \frac{1}{2}\cdot(2t-1)\cdot\kappa(X_i)+\epsilon_i \label{synthetic outcome}
\end{align}
We define the mean effect by $\eta: \mathbbm{R}^{\mathcal{F}_1 \subseteq \mathcal{F}} \mapsto \mathbbm{R}$ and the treatment effect by $\kappa: \mathbbm{R}^{\mathcal{F}_2 \subseteq \mathcal{F}} \mapsto \mathbbm{R}$, and introduce a noise term $\epsilon_i \sim \mathcal{N}(0,0.01)$ independent of $X_i$. The three model designs we consider are as follows:
\begin{enumerate}
  \item $\displaystyle \mathcal{F} = 2; \; \eta(x) = \frac{1}{2}x_1 + x_2; \; \kappa(x) = \frac{1}{2}x_1$ 
  \item $\displaystyle \mathcal{F} = 10; \; \eta(x) = \frac{1}{2}\sum_{f=1}^{2}x_f + \sum_{f=3}^{6} x_f; \; \kappa(x) = \sum_{f=1}^{2}\mathbbm{1}\{x_{f} > 0\}\cdot x_{f}$ 
  \item $\displaystyle \mathcal{F} = 20; \; \eta(x) = \frac{1}{2}\sum_{f=1}^{4}x_f + \sum_{f=5}^{8} x_f; \; \kappa(x) = \sum_{f=1}^{4}\mathbbm{1}\{x_{f} > 0\}\cdot x_{f}$ 
\end{enumerate}
Note that $\eta$ and $\kappa$ are functions mapped from subsets of features. Furthermore, in designs 2 and 3, some features are not present in either function. Such a setup should help us assess the models' ability to train over only those features that are relevant to the patient outcome. In designs 2 and 3, the treatment effect $\kappa(x)$ is always non-negative, and so, not accounting for the noise, treatment $t_i = 1$ always yields an outcome that is equal to or better than the outcome if treatment $t_i = 0$. The purpose of such a design is to test the regularization capabilities of complex learning against skewed data. Note that simpler models, such as prescriptive trees, are incentivized to design policies where every individual receives treatment $t_i = 1$.   

To simulate conditionally randomized treatments, we use varying parameter $p \in \{0.1,0.25,0.5,0.75,0.9\}$ for the probability of correct treatment assignment. Marginally randomized treatments are simulated at $p=0.5$. We generate 5 random datasets with a train:test split of 100:10,000 datapoints, and 5 with a train:test split of 500:10,000 datapoints for each $p$. We evaluate each model by taking the average Out-of-Sample Performance (OOSP) – the percentage of correct treatment assignments – over the 5 datasets. The prescriptive tree models used as benchmarks necessitate binary feature for a feature-based branching rule. Therefore, we transform the simulated real and continuous covariates into binary-valued features by discretizing them into deciles from a normal distribution and obtaining their one-hot encoded vectors. The one-hot encoded vectors were further transformed by converting $x_{i(j+1)}$ to $1$'s if $x_{ij} = 1$ for all $j \in \{1,..,10\}$ and for all $x_i \in \mathcal{X}$. We refer to these transformed covariates as \textit{adapted binarized covariates}. The adapated binarized covariates were used for all three experimental designs for the prescriptive trees. To address the value that adapted binarization adds to the prescriptive tree formulations, we run one additional set of experiments on experimental design 1 with one-hot binarized covariates. 

The PNN and J-PT models necessitate counterfactual outcome estimation. We utilize a doubly robust estimation model, where we consider linear regression and a lasso regression model to fit the counterfactual outcomes directly, and a decision tree and a logistic regression model to estimate the propensity scores. The linear regression and decision tree models were found to be the best pair of models via a model selection regime that uses 10-fold cross validation for all simulated data and compares classification accuracies for each pair. The parameters for the linear regression and decision tree models were passed as arguments to the \cite{wager2018estimation} causal forest model, which estimates the propensity score and counterfactual data internally. We also used 10-fold cross validation to tune $\ell_0$ and $\ell_1$ regularization hyperparameters for the PNN models.

\subsubsection{Personalized Postpartum Hypertension Treatments}\label{setup:hypertension}

We use prescriptive models to design treatment policies that minimize the maximum postpartum systolic blood pressure (SBP) at a patient-level. Hypertensive disorders are one of the leading causes of morbidity and mortality worldwide \citep{garovic2022hypertension}. Approximately one third of morbidity and mortality occurs during the first six weeks postpartum. Partially, this is because pregnancies complicated by hypertensive disorders will often have exacerbation of hypertension in the postpartum period that is likely secondary to a combination of mobilization of extravascular fluid as well as a persistence of preeclampsia \citep{garovic2022hypertension}. This can lead to elevations in blood pressure that are often described in the first 3 to 7 days postpartum \citep{hirshberg2016clinical}. Thus evaluating how different treatment policies effect SBP during this time period is a problem of significant importance to the obstetrics community. The data comes from a retrospective cohort study from a single United States Midwestern academic center of all women who delivered from 2009-2018. The original data set includes 32,645 deliveries, where 170 were readmitted in the postpartum period due to a hypertension-related diagnosis \citep{hoppe2020telehealth}. This data set is an observational study that includes 67 features collected from individuals during specific milestones – baseline/pre-pregnancy through the prenatal period (first, second, and third trimesters, pre-admission period and the day of admission for delivery) and up to 42 days postpartum. Of the 67 features, a subset of 18 were chosen as covariates (12 binary, 3 categorical, and 3 continuous) after eliminating correlated features by consulting with domain experts. We considered two treatment conditions – whether or not the patient was prescribed blood pressure medication, and which day the patient was discharged postpartum. We take a patient's maximum SBP postpartum to be a function of their covariates conditioned upon the treatment that they received. We designed three experiments with this dataset, all with the objective to estimate policies that minimize the maximum SBP and differ by the treatment conditions: (1) Blood pressure medication as a binary treatment condition, (2) Discharge day postpartum as an ordinal treatment condition, (3) Both blood pressure medication and discharge day postpartum.


\begin{figure}[ht]
    \centering
    \includegraphics[width=0.20\textwidth]{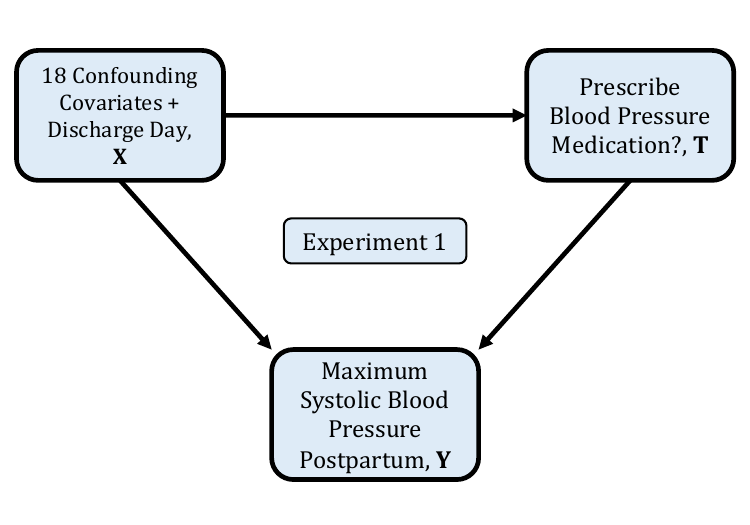}
    \includegraphics[width=0.20\textwidth]{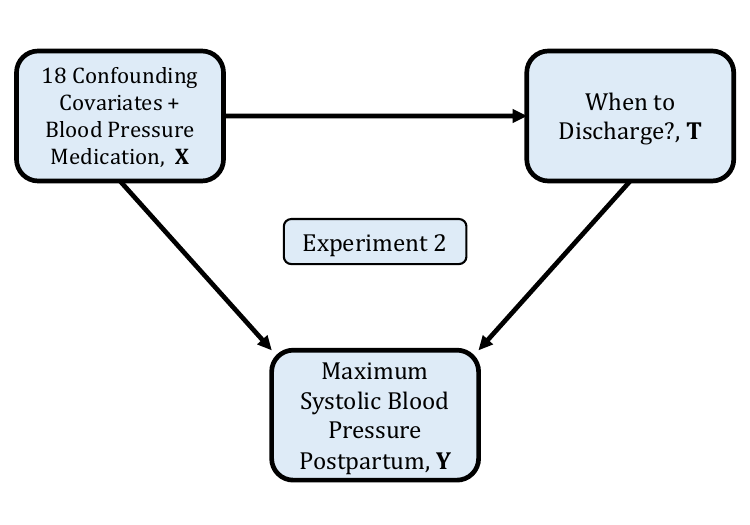}
    \includegraphics[width=0.20\textwidth]{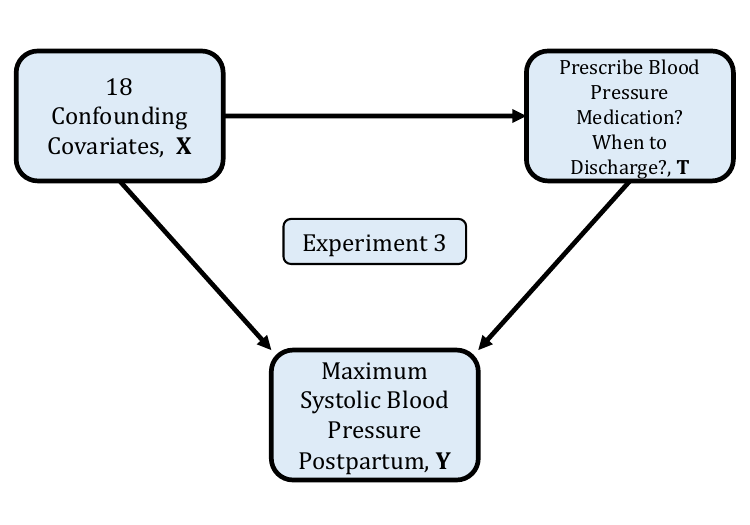}
    \caption{Causal Graphs Detailing the Covariates, Treatments, and Outcomes Along with their Causal Structures for the Three Hypertension Experiments.}
    \label{causal graphs}
\end{figure}

Figure \ref{causal graphs} depicts the characteristics of the three experiments in the form of causal graphs. For experiments 1 and 2, we considered the alternate treatment condition as a covariate (discharge day postpartum and blood pressure medications, respectively). We conducted experiments on a subset of the complete dataset to account for missing and incomplete data. Additionally, we discarded a single outlier – a patient who was not prescribed any blood pressure medication but was discharged on the 5th day postpartum – to address the case where experiment 3 sees only one patient with this treatment regime. The final dataset included 291 individuals from whom we had at least 38 blood pressure measurements. The discharge day postpartum for every individual was transformed into 4 ordinal categories: \{day 1 or day 2, day 3, day 4, day 5+\}. We performed 10-fold cross validation to tune the $\ell_0$ regularization hyperparameter for the PNN models. Finally, for the decision tree models, the continuous covariates in the dataset were discretized and transformed into adapted binarized covariates, and the categorical covariates were split into one-hot encoded binary vectors.

We considered a logistic regression model with elastic net regression, a random forest regressor, and the multilayer perceptron regressor to estimate the counterfactual outcomes for these experiments and evaluated their performance using their root-mean-squared errors. To estimate the propensity scores, we fit a logistic regression model, decision tree classifier, and random forest classifier and evaluated them using their classification accuracies. We used 10-fold cross validation to conclude that the logistic regression models produced the best estimators for experiment types 1 and 2, with the random forest regressor yielding better estimators was better for some cross-treatments in experiment 3.  We evaluate each model using the average performance of the learned policy $\pi(s)$ over each experiment run with 10-fold cross-validation. 

\subsection{Results}\label{results}

In this section, we analyze the results of the experiments conducted on simulated and real-world data. We build a PNN of depth 2 (one hidden layer, and one output layer) for every experiment. PNNs of widths 3 and 10 were benchmarked against the prescriptive tree formulations of \cite{bertsimas2019optimal} (B-PT), 
\cite{kallus2017recursive} (K-PT), and \cite{jo2021learning} (J-PT), all with depths 1 and 2. Experiments on the Causal Forest (CF) model use the EconML package \citep{econml} with their default arguments, which builds 100 trees in the forest. All approaches were allotted a run time of 1 hour.

\subsubsection{Simulated Data}\label{results:simulated}

A summary of the results of the three experimental designs for the simulated dataset can be found in Figures \ref{simfig_design1} and \ref{simfig_design2&3}. The complete table of results can be found in Appendix \ref{completedata}. Results for the hyperparameter tuning can be found in Appendix \ref{pnn_hyperparameter_tuning}. It is evident from these figures that the PNN models match or outperform the OOSP of its benchmarks in every experiment. Additionally, their performance is unperturbed by the probability of correct treatment assignment in the data. Furthermore, the PNN achieves a strong OOSP given a 100-sample training set as well as a 500-sample training set in the allocated time. 

\begin{figure*}[ht]
    \centering
    \includegraphics[width=1\textwidth]{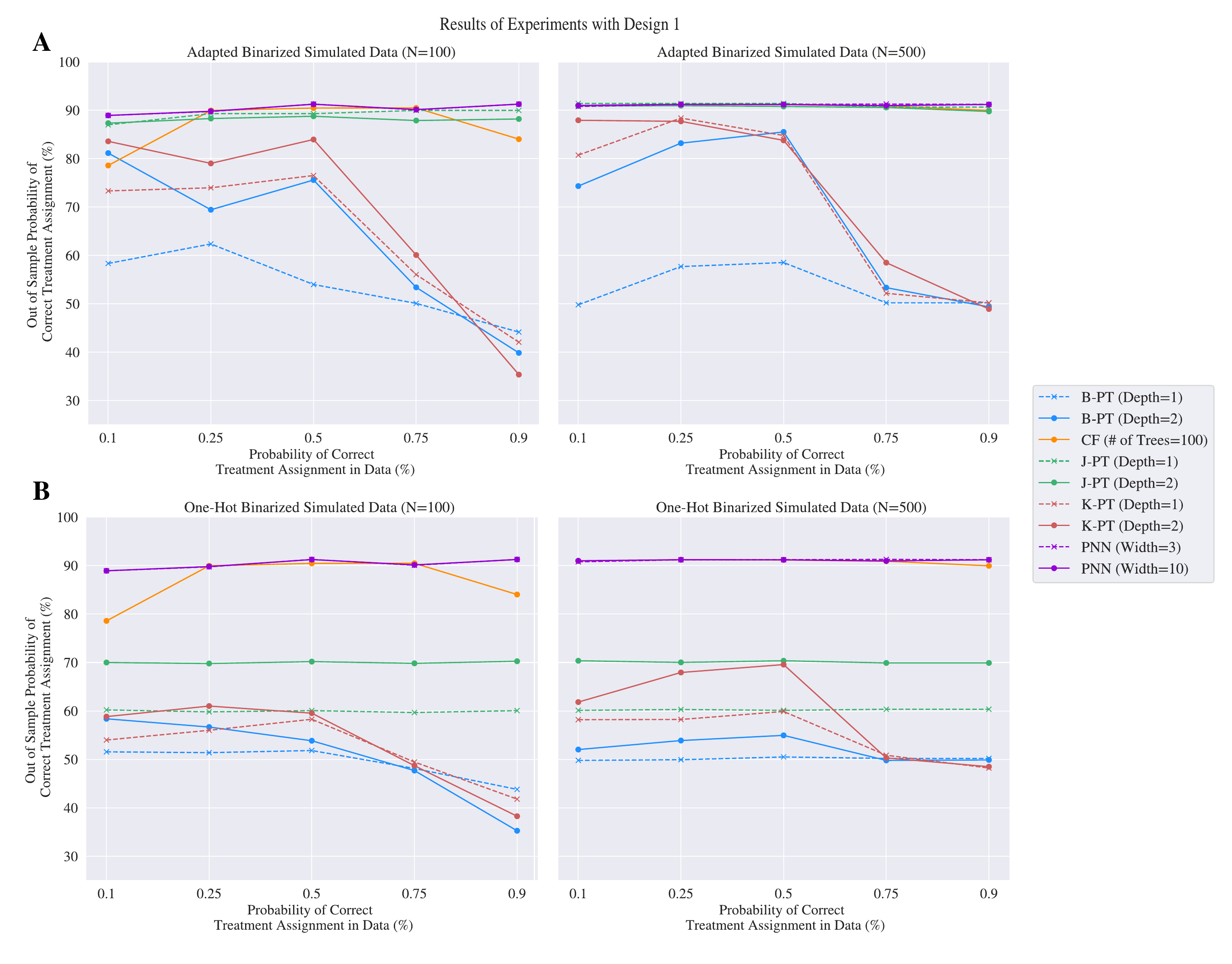}
    \caption{Results of the simulated dataset with experimental design 1. Panel \textbf{A} visualizes the results when the data is binarized using adapted encoding. Panel \textbf{B} visualizes the results when the data is binarized using one-hot encoded binarization. The left plots are policies learned from 100 datapoints and the right plots are policies learned from 500 datapoints. 
    }
    \label{simfig_design1}
\end{figure*}

Panel A of Figure \ref{simfig_design1} shows the experimental results of design 1 when the covariates have the adapted binarization for the prescriptive trees. Panel B of Figure \ref{simfig_design1} depicts the results for the same design, but with one-hot encoded binarization. These results show that the performance of prescriptive trees are sensitive to different featurization of the data. While the best prescriptive tree model achieves an OOSP around 90\% across all treatment assignments when the covariates are binarized to support feature splitting with the adapated binarization, the best prescriptive tree only yields a policy with a 70\% OOSP when discretized into deciles without the featurization. These experiments also show that the performance of the causal forest model is dependant on to the probability of correct treatment assignment in the data, as the OOSP for this model tapers off at the 0.1 and 0.9 probabilities. This means the causal forest designs policies that are stronger when the treatments in the data are marginally randomized, but are weaker with more conditionally randomized treatments. Additionally, the B-PT and K-PT formulations drop below a 50\% OOSP when the probability of correct treatment assignment approaches 0.9. That is, when the informed policy assigns the correct treatment to most patients, these trees design a policy that are akin to a random treatment assignment. Given that these models achieve a 0\% optimality gap for these datasets, this drop is performance is likely attributed to a failure in branching on key features at the given depths.

\begin{figure*}[ht]
    \centering
    \includegraphics[width=1\textwidth]{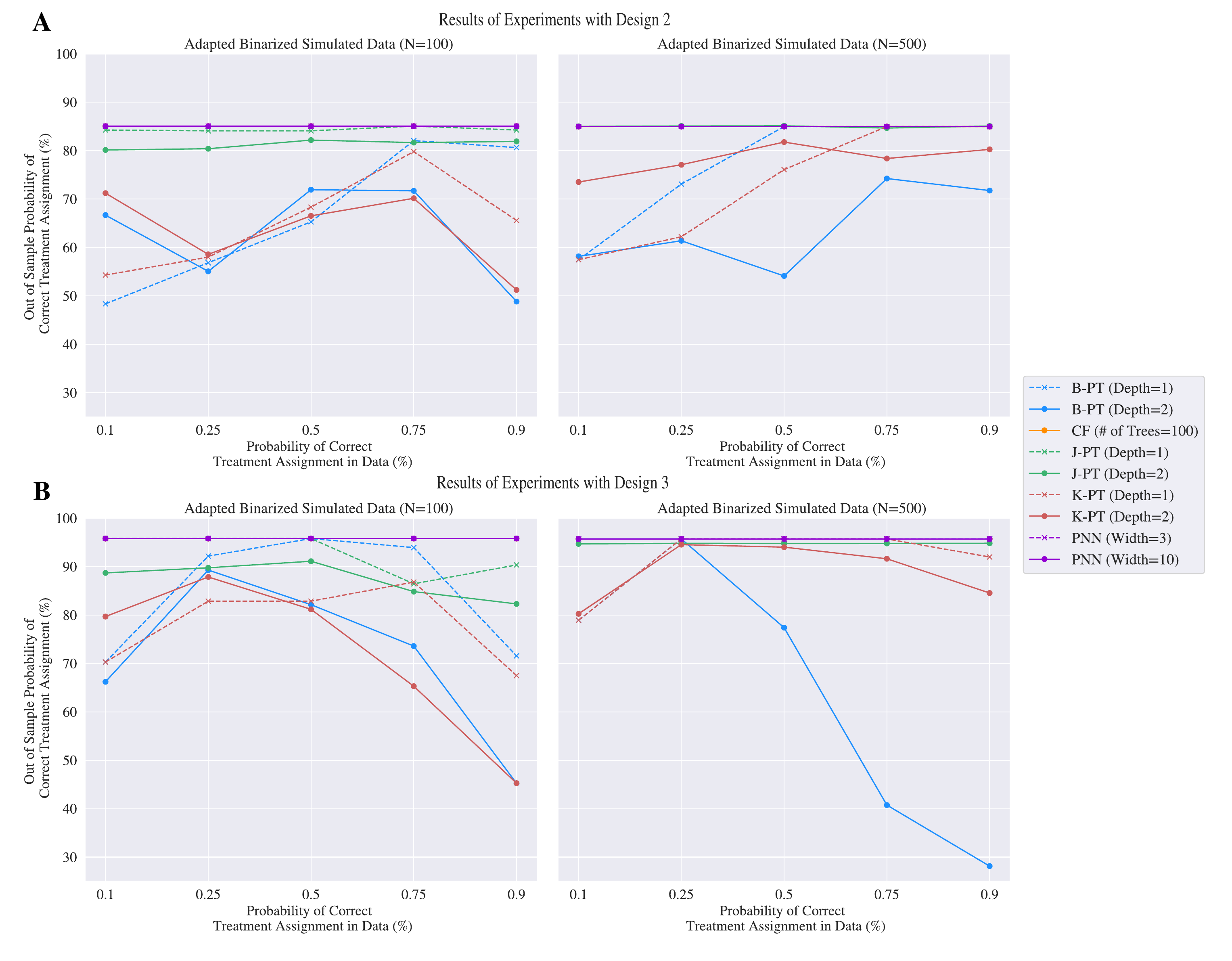}
    \caption{Results of the simulated dataset with experimental design 2 and 3. Panel \textbf{A} visualizes the results when the data is binarized using one-hot encoding with the left learned from 100 datapoints and the right learned from 500 datapoints. Panel \textbf{B} visualizes the results when the data is binarized using adapted binarization with the left learned from 100 datapoints and the right learned from 500 datapoints. 
    }
    \label{simfig_design2&3}
\end{figure*}

Figure \ref{simfig_design2&3} shows the experimental results of designs 2 and 3 on Panels A and B respectively, with the adapted binarization for the prescriptive trees. Recall that for these designs, it is easy to achieve a strong OOSP from a policy that assigns treatments liberally. In fact, the policies designed by the PNN and the causal forest are identical in each experiment for designs 2 and 3, with an average OOSP of around 85\% and 95\% respectively achieved by assigning treatments to everyone. Unsurprisingly, the depth 1 counterpart for each prescriptive tree model achieves a higher peak OOSP given the structure of the datasets. We see that the J-PT model with appropriate depths achieve the same policy as the PNNs and the causal forest when trained on a larger training set. However, there exists a larger variance in the performance of the B-PT and K-PT models, implying limitations in identifying key branching features that support a treatment-forward policy. Tables \ref{simulatedfull2e} and \ref{simulatedfull3e} in Appendix \ref{completedata} show that the policies designed by the B-PT and K-PT models have a standard deviation in the OOSP of up to 36.55\% and 16.16\% respectively. The drop in performance of the B-PT model in the 500-sample training set plot of Panel B in Figure \ref{simfig_design2&3} is attributed to the fact that the policy was designed by the heuristic solution found by the solver; an optimal solution was not found in the allocated time limit. 

\subsubsection{Personalized Postpartum Hypertension Treatments}\label{results:hypertension}

The experiments for the hypertension datasets involved designing policies that reduce the maximum systolic blood pressure (SBP) of postpartum patients in the study. Since the counterfactual outcomes for the patients are unknown, it is difficult to evaluate and compare various policies. Hence, we computed the expected CATE of each policy. Note that we estimate the average CATE for the policy in the data using the doubly robust estimator to allow for an adjusted comparison of the learned policies against the policy in data. we used 10-fold cross validation to tune the $\ell_0$ regularization hyperparameters for the PNN. Results for the hyperparameter tuning can be found in Appendix \ref{pnn_hyperparameter_tuning}.

\begin{figure*}[ht]
    \centering
    \includegraphics[width=1\textwidth]{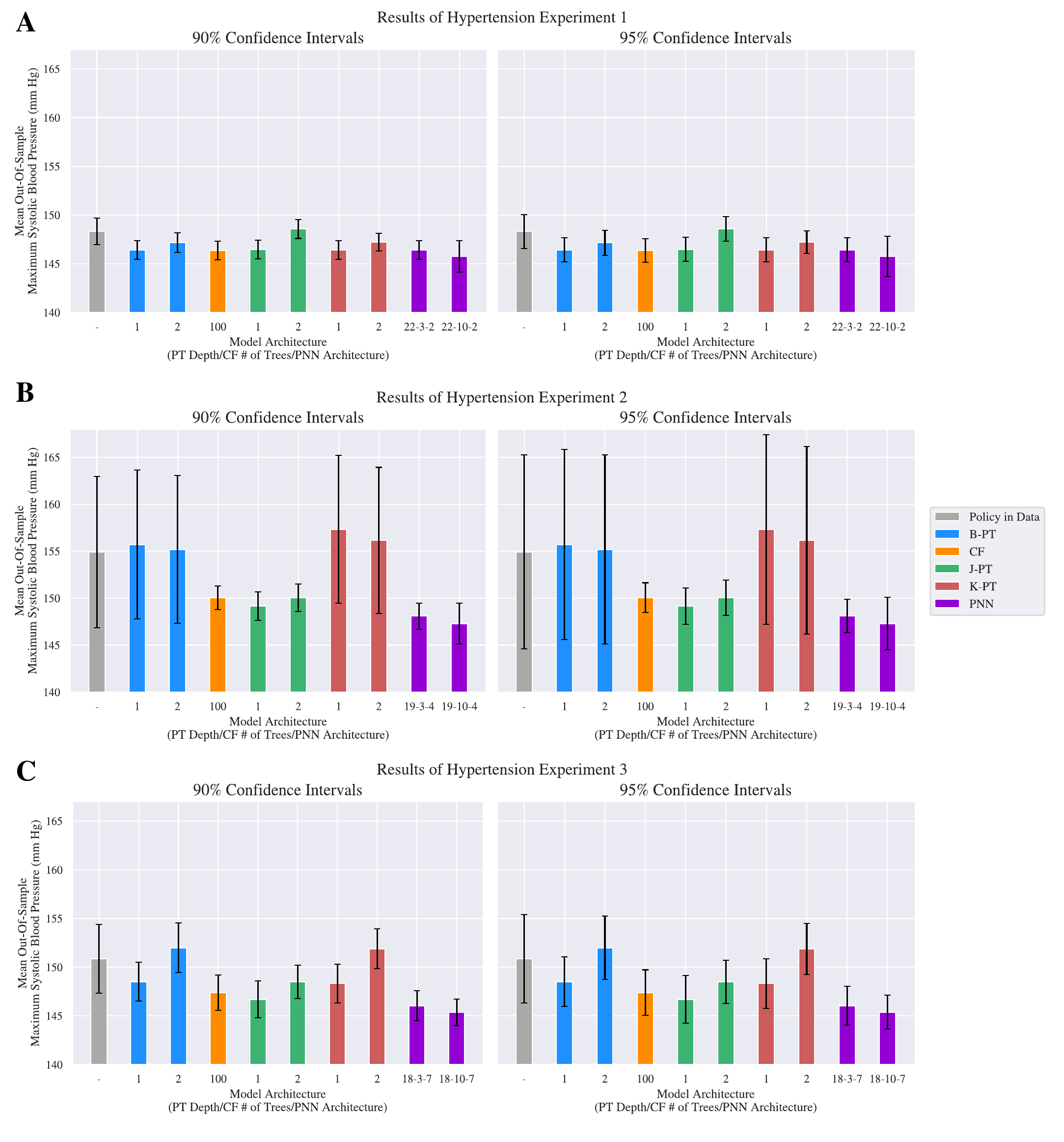}
    \caption{Results of experiments on the hypertension dataset for out-of-sample patients prescribed treatments based on policies learned by different models. Each panel displays the mean out-of-sample maximum systolic blood pressure. The left plots include the 90\% confidence intervals and the right plots include the 95\% confidence intervals. 
    }
    \label{hypertension_barplots}
\end{figure*}

Figure \ref{hypertension_barplots} provides a summary of the results including the confidence intervals for the three experiments. The complete experimental results can be found in Appendix \ref{completedata}. Recall that the experiments differ in terms of how the treatment condition is defined: Experiment 1 considers whether or not a patient receives blood pressure medication, Experiment 2 considers the day a patient is discharged postpartum, and Experiment 3 considers the cross-product of the two. We observe that the PNN model outperforms the decision tree and causal forest benchmarks with regards to the mean out-of-sample maximum SBP in every experiment while maintaining a low variance. 

\begin{figure*}[ht]
    \centering
    \includegraphics[width=1\textwidth]{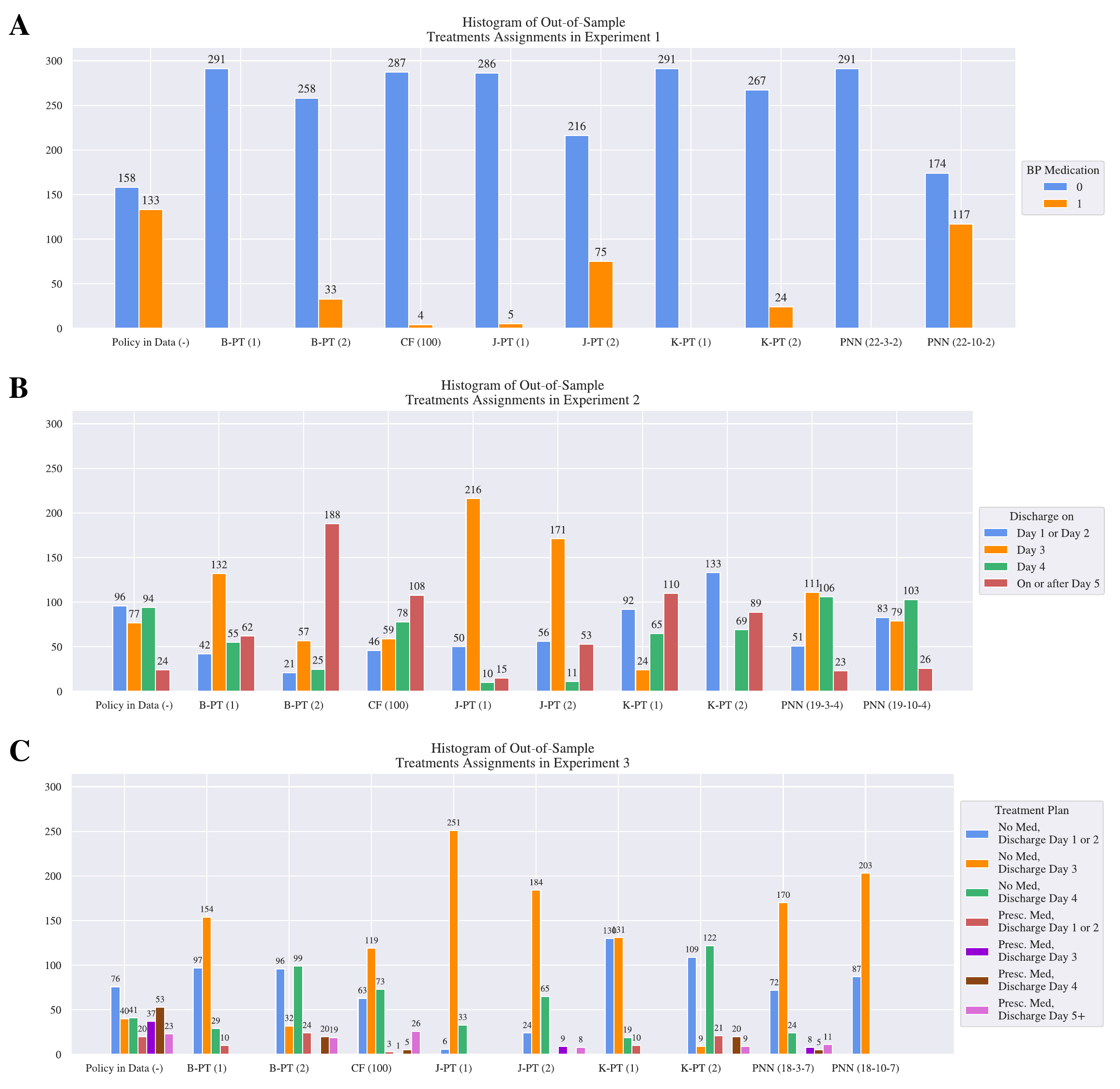}
    \caption{Out-of-sample treatment assignments for policies learned by different models over the hypertension dataset. Each panel displays the number of patients who were prescribed each treatment. 
    }
    \label{hypertension_treatment_assignments}
\end{figure*}

To further analyze the behavior of each model, we refer to Figure \ref{hypertension_treatment_assignments}. We observe that when a model has a smaller architecture, the policies they design are extremely treatment averse in experiment 1 (Panel \textbf{A}). We posit that the complexity of these models is not high enough to capture the causal effect in the data. Note that the depth 2 decision trees design policies with a larger average maximum SBP than the depth 1 trees. This anomaly could be attributed to the fact that the K-PT and J-PT formulations do not regularize against overfitting. Furthermore, while the B-PT formulation does incorporate a regularization term, the depth 2 variant exhibits a large average optimality gap at the time limit. We also note that the 10-width PNN was able to outperform all of the other models by designing a treatment policy that is more willing to prescribe the blood pressure medication.

Panel \textbf{B} of Figure \ref{hypertension_barplots} indicates that policies consider the discharge day as the treatment for postpartum patients lead to a larger mean maximum SBP across all models. The B-PT and K-PT trees perform worse than the policy in data, and reveal a large variance, despite the B-PT formulation incorporating regularization. This may be explained by the large optimality gap in the depth 2 tree, and the notion that the depth 1 was unable to regularize successfully with 4 treatment groups. Panel \textbf{B} of Figure \ref{hypertension_treatment_assignments} additionally shows that the causal forest model recommends a highly risk-averse regiment, with the majority of the patients being discharged on or after day 4. Additionally, the J-PT trees are more inclined to discharge patients at day 3. Considering the performance of the J-PT trees and the PNNs, it can be interpreted that the majority of the patients benefit from being discharged on day 3 and day 4.

Panels \textbf{C} of Figures \ref{hypertension_barplots} and \ref{hypertension_treatment_assignments} indicate the results of experiment 3. The general trends of experiments 1 and 2 are seen to be present in experiment 3 as well, with the majority of smaller architecture models designing policies that are treatment-averse. Note that the 10-width PNN only suggests a treatment plan where patients receive no blood pressure medication, and are discharged either on days 1, 2 or 3. While this may seem irregular, the policy yielded the lowest average maximum SBP. We also see that the J-PT trees maintain their affinity for discharging patients on day 3. We hypothesize that the lack of regularization explains the decline in performance of depth 2 trees in comparison with their depth 1 counterparts, and that the B-PT depth 2 tree failed to achieve a small optimality gap to design a stronger policy.

\begin{figure*}[ht]
    \centering
    \includegraphics[width=1\textwidth]{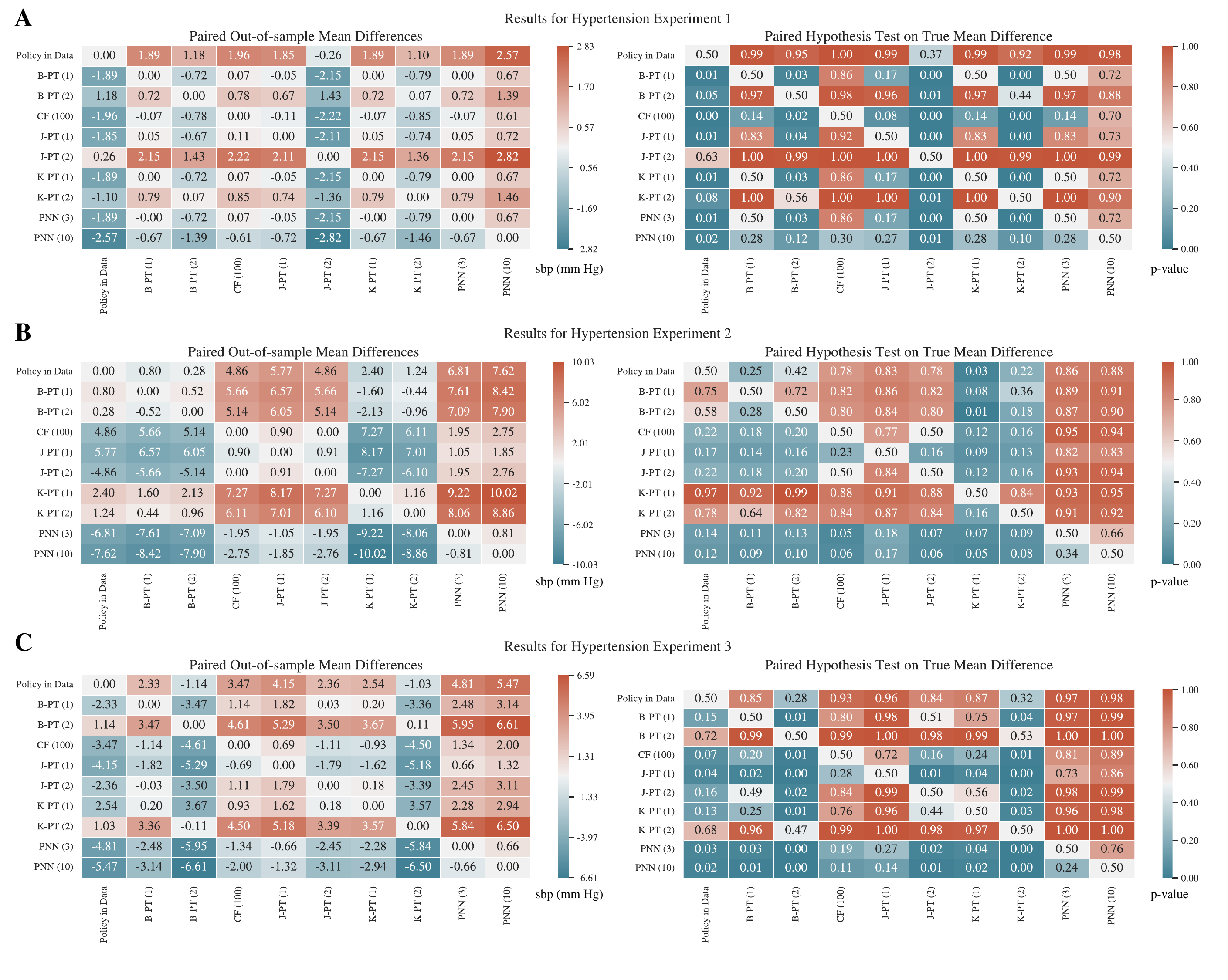}
    \caption{Paired results of experiments on the hypertension dataset for out-of-sample patients prescribed treatments based on policies learned by different models. Each panel displays paired evaluation of the various models. The left heatmaps display the point differences of out-of-sample mean maximum systolic blood pressure between pairs of models. Differences are taken as \textit{row - column}. The right plots display the hypothesis tests on the differences of out-of-sample mean maximum systolic blood pressure between pairs of models. Smaller p-values favor models in the rows. Cool colors denote lower blood pressures (left) or smaller p-values (right). Axes values denote the model (PT depth/\# of trees in CF/PNN width).}
    \label{hypertension_heatmaps}
\end{figure*}

Additionally, we conducted a paired hypothesis test to make statistically significant claims on the performance of each model. The results of the hypothesis test are presented in Figure \ref{hypertension_heatmaps}, which aims to provide a visualization that allows for a fair comparison of the models. The left subplots show the difference in the mean out-of-sample maximum systolic blood pressure (SBP) between each pair. The right subplots show the p-values of the paired hypothesis tests given the null hypothesis $N_0: \mu_{row} - \mu_{column} = 0$, where $\mu_{row}$ is the true mean of the model on the rows and $\mu_{columns}$ is the true mean of the model on the columns. PNNs design policies with a mean out-of-sample maximum SBP that is statistically significantly lower than its competitors' policies at least one of the chosen architecture. We note that the large variance with the B-PT and K-PT models are translated here as relatively large p-values. Overall, our results show that PNNs statistically outperform other causal and prescriptive models in this setting.

\subsection{Model Interpretation}

High stakes domains such as personalized medicine in the context of assigning treatments for postpartum patients 
demand policies that are fair and can be audited for individual safety. Therefore, interpretability and accountability are highly desirable. In this section, we assess each model's ability to discern important features in the personalized postpartum hypertension treatment assignment. 

Analyzing the tree-based approaches is straightforward – features that are deemed more important are branched on higher in the tree. Given that we experiment on trees with a depth of at most 2, these models should therefore choose at most 3 features to branch on. Single-layer PNNs are similarly straightforward to review – features consider more important have larger absolute weights, and less important have weights closer to 0 along the arcs of the network. The inclusion of the $\ell_0$ regularization term in the objective further motivates sparsity in the weights, making the model easier to interpret. However, the causal forest is much less straightforward to interpret. In order to analyze the policies designed by all the models on a level field, we turn to Shapley Additive Explanations, or SHAP values \citep{lundberg2017unified}. We compute the global SHAP values for the features in each model to interpret the prescriptive trees, causal forests PNNs.


To assess the policies designed by the 5 models in the personalized postpartum hypertension treatment experiments, we ranked the features by their SHAP values for each model. Since these experiments followed a 10-fold cross-validation regime, we observed 10 models with different policies. To gain a better understanding of these policies in general, the feature rankings were then averaged across these 10 models. Furthermore, since the prescriptive trees branch on at most 3 features, those that were not chosen to branch on were given the lowest ranking. Recall now that the prescriptive trees optimize on a binarized feature space. To fairly compare the PNN and CF models with the tree models, we took the minimum rank for each binarized feature as the reported rank. Finally, to account for varied number of features due the binarization, the rankings were normalized between 0 and 1. We then subtract these normalized rankings from 1 to maintain an intuitive representation of these rankings. So, features with a larger normalized rank are considered more important by the model. We note that rows with null values represent those models where only treatment was assigned in each fold of the cross-validation; due to this behavior, the results of the SHAP model are inconclusive for these cases. 

We now analyze the models based on their feature rankings. Tables \ref{hypertension shap ranking exp1}-\ref{hypertension shap ranking exp3} in Appendix \ref{interpret_appendix} display these rankings for the three experiments. We see that the features that are commonly ranked highly in the PNNs are prenatal BMI, chronic hypertension, mode of delivery, gestational hypertension, gestational age, and preeclampsia. Additionally, in experiments 1 and 2, discharge day postpartum and blood pressure medication are highly ranked, respectively. These observations align well with medical and clinical preferences \citep{hoffman2021machine,mukhtarova2021evaluation}. We also note that social characteristics such as insurance type is often ranked low by the PNNs. We see that the causal forest ranks similar features highly, with the birthing person's age often considered most important. We also observe that the B-PT and K-PT models frequently consider features such as mode of delivery and prenatal BMI as important. However, the J-PT model finds it more challenging to identify medically relevant features in experiment 1 and 2. We note that ethnicity and insurance are among the highest ranked feature by the depth 2 model in both experiments, which may be a concerning since these attributes are generally considered protected. Note that the J-PT model appropriately identifies important features in experiment 3. This may be because correlations brought on by including the alternate treatments as covariates are negated.

\section{Managerial Implications}\label{managerial}
From our case studies, our implementation of MIP-based ANNs does not only provide strong empirical performance, but also provides additional insights into how our framework can be deployed in healthcare and business settings.

\begin{enumerate}
    \item \textbf{The PNN framework yields consistently strong polices when given limited data and optimized over small networks.} The policies designed by the PNN models used in the experiments consistently achieve the highest out-of-sample probability of correct treatment assignment in the simulated experiments. Moreover, when used in the real-world case study of treating patients with hypertension during pregnancy and postpartum, the PNN models result in the lowest average maximum systolic blood pressure postpartum when considering any of the proposed treatment conditions. In particular, the width 3 and width 10 models were consistently the two best performing models, with the best policy outperforming the policy in the data by 7.62 mmHG (see Figure \ref{hypertension_heatmaps}) on average, in a statistically significant manner. When compared against other prescriptive models, the width-10 PNN designed policies that resulted in lower point average maximum systolic blood pressure in every case. Moreover, results for the join treatment plan in experiment 3 showed a statistically significant improvement, with a $p$-value less than $0.05$, against all but two models: the depth-1 J-PT ($p$-value=$0.14$) and CF with 100 trees ($p$-value=$0.11$). Additionally, the ability to supplement the PNNs with a regularization term leads to lower policy variance and smaller confidence intervals across replications. The PNN models also have a better capacity to personalize treatment as opposed to some other models which tended to provide one treatment to a majority of patients (see Panels B and C in Figure \ref{hypertension_treatment_assignments}). 
    
    \item \textbf{Methods that rely on supplementary models for estimating counterfactual outcomes are sensitive to accuracy of those models.} Our experiments on  warfarin dosing  suggest that accuracy of the inverse propensity weighting method and direct method for estimating counterfactual outcomes significantly affect the performance of the policies of the models that optimize using the doubly robust estimator. In particular, given the warfarin dataset, the logistic regression model, which was deemed the best propensity score estimator via cross-validation, had an average accuracy of 66.77\%. Similarly, the elastic net model, which was the best direct estimator, yielded an average root-mean squared error of 2.056. Given that the policy in the data was ascertained to be optimal, our results show that the models that optimize over the estimated counterfactual outcomes designed sub-optimal policies. It is therefore key for decision-makers to to consider such limitations when using such prescriptive models with any dataset as they be highly sensitive to the performance of the estimation models.

    \item \textbf{The PNN is more likely to identify medically reliable features then other prescriptive methods.} The need to design interpretable policies to solve complex problems gives rise to a tradeoff for prescriptive models. When we have strong counterfactual estimations, a smaller neural network can provide better interpretability on the policies that are designed. Additionally, given that MIP models can be used solved to small optimality gaps when used to trained networks with smaller architectures, these interpretable policies also yield optimal outcomes. The PNN therefore achieves a strong tradeoff since it a complex yet parsimonious training model. As seen in Tables \ref{hypertension shap ranking exp1}-\ref{hypertension shap ranking exp3}, PNNs identify key patient characteristics, such as prenatal BMI and chronic and gestational hypertension as confirmed by our clinical collaborator, and appropriately weigh these features in designing its policy. These results were consistent across all of the experimental designs for these experiments. In comparison, we observe that some benchmarked models consider fail to identify medically reasonable features and instead consider social features such as ethnicity and insurance in their treatment assignments. The interpretability of small PNNs allows for medical professionals and policy-makers to work with this learning models to consider the best approach for assigning treatments.

\end{enumerate}

\section{Conclusions}\label{conclusions}

Our work presents the effectiveness of 0-1 neural networks in addressing the challenges of personalized decision-making, particularly in healthcare domains. By using mixed-integer programming and counterfactual estimation, prescriptive networks offer a valuable tool for optimizing treatment policies when limited patient data is available. The ability to encode complex policies while remaining interpretable sets the PNN apart from existing approaches that employ decision trees. The results obtained from both synthetic data experiments and the real-world case study of postpartum hypertension treatment assignment provide compelling evidence of the advantages offered by prescriptive neural networks towards improving healthcare outcomes.

\bibliography{paper_edit_hmc}

\pagebreak
\clearpage
\begin{appendices}

\section{MIP ANNs and Prediction}
\subsection{ANNs for Prediction: Negative Log Likelihood Loss \label{NLL}}
A common loss for prediction is the negative log likelihood loss (NLL), in this section we will present a MIP formulation for this loss in the case of classification.

Our goal is to represent Equation \ref{loss} in terms of a set of linear constraints and linear objective function such that our ANN makes predictions on data given disjoint classes. Note that the $ L^{th} $ layer outputs are the only ones that interact directly with the loss function, so we focus on these decision variables and leave the discussion of the model parameters when considering individual unit activations. The loss function proposed by \cite{toro2019training} is motivated by maximizing the neuron margin, akin to Support Vector Machines. However, max-margin models are known to under-perform on noisy data with overlapping classes \citep{shalev2014understanding}. Other prediction objectives that have been proposed in the literature are $\ell_0$ or $\ell_1$ losses that can be straightforwardly formulated through MILP linearization techniques \citep{bertsimas1997introduction, wolsey1999integer}. However, in practice a common loss function used for training ANNs in prediction tasks, is the negative log likelihood applied to a soft-max distribution \citep{goodfellow2017deep}. Such a loss can be considered analogous to minimizing the log likelihood of a multinomial distribution across the classes in $J$, thus outputting a predictive distribution over the classes.  This loss is particularly appealing as it harshly penalizes incorrect predictions and is numerically stable \citep{bridle1990probabilistic, goodfellow2017deep}. The equation for the soft-max activation function for the output of the final layer $ h_{i,j,L} $ is $\varphi(h_{i,j,L}) = \frac{\exp(h_{i,j,L})}{\sum_{j' \in J} \exp(h_{i,j',L})} \label{softmax}$, and the resulting negative log likelihood loss is $-\log(\prod_{i=1}^{n}(\varphi(h_{i,j,L})^{\mathbbm{1}[Y_i=j]}))$. We reformulate this loss using the following result:
\begin{proposition}
The negative log likelihood loss applied to the soft-max activation $-\log(\prod_{i=1}^{n}(\varphi(h_{i,j,L})^{\mathbbm{1}[Y_i=j]}))$ is within a constant additive error of the optimal value of the following linear optimization problem:
\begin{align}
 \min_{h_{i,j,L},\omega_i}\big\{ \sum_{i=1}^{n} \sum_{j \in J} \mathbbm{1}[Y_i=j](\omega_{i}-h_{i,j,L}) : \omega_i \geq h_{i,j^\prime,L}, \ j^\prime\in [J]_1,i \in [n]_1 \big\} \label{obj}
\end{align}
\label{prop:nll}
\end{proposition}
Here we present a brief sketch of the proof, for the detailed proof of this proposition see Appendix \ref{proofsappendix}. The main insight used is the fact that the function $\log \sum_{i=1}^n \exp(X_i) = \Theta(\max_{i \in [n]_1} X_i)$ \citep{calafiore2014optimization}, where $\Theta$ is the Big Theta notion that characterizes the function $\log \sum\exp$ to be bounded asymptotically both above and below by the $max$ function \citep{cormen2022introduction}, and the fact that the minimization of a maximum can be written using linear constraints \citep{wolsey1999integer}.

We note that for the output of the final layer to give a unique prediction with the negative log likelihood soft-max loss, we need add constraints which force the optimization problem to assign diverse values to $ h_{i,j,L}, \forall \; i \in [n]_{1},j \in J $:
\begin{align}
 & \omega_{i} \ge h_{i,j,L} \label{omega_define} \\
 & h_{i,j,L} + h_{i,j^\prime,L} - 2h_{i,j,L} \leq - \epsilon + Mr_{i,j,j^\prime}, \ \forall \ j,j^\prime \in [J]^{2}_{1}, j \neq j^\prime \label{diversify C1} \\
 & h_{i,j,L} + h_{i,j^\prime,L} - 2h_{i,j,L} \geq \epsilon - M(1-r_{i,j,j^\prime}), \ \forall \ j,j^\prime \in [J]^{2}_{1}, j \neq j^\prime \label{diversify C2} 
\end{align}

These constraints can be thought of as diversifying constraints, since without them the loss would be  minimized  by setting all the outputs of the final layer, $ h_{i,j,L} $ to be equal. However, owing to the fact that no data point can be labeled with multiple classes, we require unique unit outputs. Constraints \eqref{omega_define} forces $\omega_{i}$ to be the \textit{max} of the components of the output vector. Finally, Constraints \eqref{diversify C1} and \eqref{diversify C2} ensure that no two units in the output layer are equal. The idea behind these two constraints relies on the fact that the average of two variables is equal to one of the variables if and only if the two variables are equal. So, we introduce the binary variables $r_{i,j,j^\prime} \in \mathbbm{B}$ to define Constraints \eqref{diversify C1} and \eqref{diversify C2} as big-M constraints.
 
Considering that the soft-max activation of the output layer is captured by the objective function, we reformulate Constraint \eqref{L output_const} as:
 \begin{flalign}
 & h_{i,j,L} \le \sum_{k^\prime=1}^{K} (z_{i,k^\prime,j,L}) + \beta_{j,L} \label{NLL_hL_define C1} \\
 & h_{i,j,L} \ge \sum_{k^\prime=1}^{K} (z_{i,k^\prime,j,L}) + \beta_{j,L} \label{NLL_hL_define C2}
\end{flalign}
 
Constraints \eqref{NLL_hL_define C1} and \eqref{NLL_hL_define C2} guarantee that the output of the final layer is the linear combination of the units of the activated outputs of the penultimate layer summed with the bias term. 

\subsection{Numerical Experiments on Prediction Problems}\label{predictive experiments}

We now evaluate the MIP formulation for training ANNs through empirical experiments for prediction problems. We aim to show value of incorporating a negative log likelihood loss into the MIP formulation. For the prediction problems, we benchmark against the Stochastic Gradient Descent algorithm with momentum and adaptive learning rates with both binary activations and ReLU in the hidden layer. The SGD models were fit with Cross Entropy Loss which incorporates a negative log likelihood loss over the sigmoid activation in the output layer \cite{zhang2018generalized}. Experiments were done over a subset of the MNIST dataset \cite{lecun1998mnist} in the few-shot learning set-up \cite{ravi2017optimization}. All experiments were run on a high throughput computing cluster systems allocated with 4 CPU Cores, 32GB of RAM, and 4GB of Disk Space, with Python version 3.9.5 \citep{10.5555/1593511}. 

\subsubsection{Experimental Setup}

We use the MNIST handwritten dataset \cite{lecun1998mnist} to evaluate our model's performance in the image recognition problem. The images in this dataset include labelled handwritten digits from 0 through 9 in a grey-scale $28 \times 28$ pixels resolution. We design the experiments as a few-shot learning problem by taking a subset of the labels and images for training; in these experiments, we consider images labeled as $2$ or $7$, and sampled $100$ images for training and $1000$ images for testing. Sampling was conducting with 5 seeds and without replacement between the training and testing sets to evaluate the average out-of-sample testing accuracy of each model. As a result, each model was given $100$ reshaped images of size $28 \times 28 = 784$ and was asked to learn to predict dichotomous labels.

The baseline SGD with momentum models were trained on $50,000$ iterations, an initial learning rate $=0.1$, a constant learning rate decay $=0.5$ every $10,000$ steps, an initial momentum $=0.5$, and final momentum $=0.95$. These hyperparameters were fine tuned through cross-validation for both binary and ReLU activation models. Finally, we used cross-validation to fine tune the $\ell_0$ regularization term in the objective of the MIP ANN model. The results of the cross-validation experiments showed that not having a regularization term yields the best performance for this experimental design.

We assess the performance of the MIP, SGD-Binary and SGD-ReLU ANNs over a series of architectures, reported in the form (number of hidden layers, number of neurons): (1,3), (1,5), (1,10), (2,3), (2,5), (2,10), (3,3), (3,5), (3,10). The results of such a neural network architecture search are reported in this section to better understand the settings in which a MIP-based training regime serves as viable alternative to SGD. 

\subsubsection{Results}

We analyze the results of the prediction problem experiments conducted over the 2-Digit MNIST datasets in this section. All approaches were allotted a run time of 1 hour.

\begin{figure*}[ht]
    \centering
    \includegraphics[width=1\textwidth]{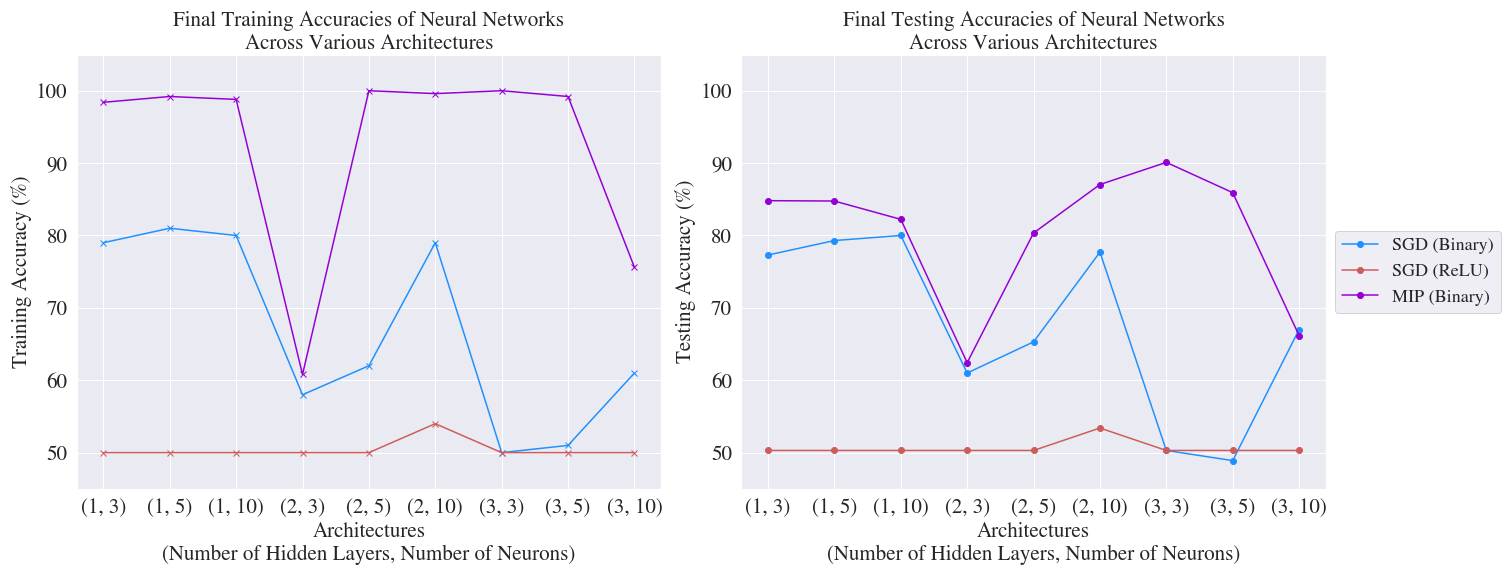}
    \caption{Average Final training (left) and testing (right) accuracies of the three prediction ANNs over the 2-Digit MNIST dataset.}
    \label{2digit prediction plot}
\end{figure*}

Figure \ref{2digit prediction plot} shows a summary of the results averaged over the 5 seeds. The full experimental results can be found in Appendix \ref{completedata}. We observe that the MIP-based training approach outperforms the SGD training approaches in all but one ANN architectures when considering the average final testing accuracy. On average, the MIP-based ANN with 3 hidden layers with 3 neurons in each hidden layer yielded the best final testing accuracy at ~90\%. For the 3 hidden layers and 10 neurons architectures, we found that the MIP model was unable to solve to optimality in the allocated time limit, due to which we associate the deterioration in performs. Additionally, we see that when we use the ReLU activation function in the hidden layers for the SGD approach, the model fails to perform better than random. In fact, our experiments showed that the performs of the ReLU model improves when we train over deeper neural networks; considering the few-shot learning setting, however, the binary models were able to yield results better than random classification with much more parsimonious neural networks. We find that these results are consistent with that which is observed by \cite{toro2019training, kurtz2021efficient, thorbjarnarson2023optimal}.

\section{Proofs}\label{proofsappendix}
\subsection{Proof of Proposition \ref{prop:reform_bin}}
\proof
 There are two key challenges for the reformulation of Constraints \eqref{l_0 output_const},\eqref{l output_const},\eqref{L output_const} first is the binary activation function itself and second are the bi-linear terms present in the hidden and output layers. Without loss of generality, let us consider a particular unit in the neural network such that $h_{n,k,\ell} = \mathbbm{1}[ \sum_{k^\prime =0}^K \alpha_{k,k'}h_{k^\prime,\ell} + \beta_{k,\ell} \geq 0]$. First let us consider the case of the binary activation. This condition can be modeled as a disjunction \citep{wolsey1999integer,conforti2014integer}, a form of constraint that can be reformulated as follows:
 \begin{align}
 &\sum_{k^\prime =0}^K \alpha_{k,k'}h_{k^\prime,\ell} + \beta_{k,\ell} \leq M h_{n,k,\ell} \\
 & \sum_{k^\prime =0}^K \alpha_{k,k'}h_{k^\prime,\ell} + \beta_{k,\ell} \geq \epsilon + (-M-\epsilon)(1- h_{n,k,\ell}) 
 \end{align}
 Where $M$ is a sufficiently large constant, and $\epsilon$ is a small constant. Next let us consider the bi-linear terms $\alpha_{k',k,\ell}h_{k^\prime,\ell-1}$. Here we note that $\alpha_{k',k,\ell} \in [\alpha^L,\alpha^U]$ is a continuous valued variable, while $h_{k^\prime,\ell-1} \in \mathbbm{B}$ is binary valued. This type of product can be reformulated though a standard technique by introducing an additional binary variable $z_{n,k^\prime,k,\ell} = \alpha_{k',k,\ell}h_{k^\prime,\ell-1}$ then these products can be written as:
 \begin{align}
 & z_{n,k^\prime,k,\ell} \le \alpha_{k^\prime,k,\ell} + M(1-h_{n,k^\prime,\ell-1}) \\
 & z_{n,k^\prime,k,\ell} \ge \alpha_{k^\prime,k,\ell} -M(1-h_{n,k^\prime,\ell-1}) \\ 
 & -Mh_{n,k^\prime,\ell-1} \le z_{n,k^\prime,k,\ell} \le Mh_{n,k^\prime,\ell-1} 
 \end{align}
 where again $M$ is an appropriately picked sufficiently large constant. Using these reformulations where appropriate and making the necessary variable substitutions results in the formulations presented above. 
\endproof

\subsection{Proof of Proposition \ref{prop:nll}}
\proof
 First, we can explicitly write the negative log likelihood as: 
\begin{align}
 & -\log(\prod_{n=1}^{N}(\sigma(h_{n,j,L})^{\mathbbm{1}[y_n=j]})) = -\sum_{n=1}^{N} \mathbbm{1}[y_n=j]\log\Big(\frac{\exp(h_{n,j,L})}{\sum_{j' \in J} \exp(h_{n,j',L})}\Big) \label{NLL 1} \\
 & = -\sum_{i=n}^{N} \mathbbm{1}[y_n=j](h_{n,j.L}-\log \sum_{j' \in J} \exp(h_{n,j',L})) \label{NLL 2}
\end{align}
Here we note that the expression $\log \sum_{j' = J} \exp(h_{n,j^\prime,L}) = \Theta(\max_{j^\prime \in j} h_{n,j^\prime,L})$ \cite{calafiore2014optimization}. Thus using this fact we obtain:
\begin{align}
 \eqref{NLL 2} & \leq -\sum_{n=1}^{N} \mathbbm{1}[y_n=j](h_{n,j,L}-\max_{j' \in J}\{h_{n,j',L}\} - \log(|J|)) \label{NLL 3} \\
 & = \sum_{n=1}^{N} \mathbbm{1}[y_n=j](\max_{j' \in J}\{h_{n,j',L}\}-h_{n,j,L} + \log(|J|)) \label{NLL 4}
\end{align}
Using the other side of the big $\Theta$ condition yields that $\eqref{NLL 2} \geq \sum_{n=1}^{N} \mathbbm{1}[y_n=j](\max_{j' \in J}\{h_{n,j',L}\}-h_{n,j,L})$. If we let $\omega_{n} = \max_{j' \in J}\{h_{n,j',L}\}$ we can use standard formulation techniques to reformulate it using linear constraints \cite{wolsey1999integer}. In conjunction with removing the constant terms that do not depend on $\omega_n,\{h_{n,j',L}\}$ yields the desired result. 
\endproof

\subsection{Proof of Proposition \ref{prop:pi obj}}
\proof
Given that we are in the dichotomous treatment case, \cite{farrell2021deep} gives us the following estimators for causal parameters for the full-population averages:
\begin{flalign}
    \hat\tau & = \mathbbm{E}_n[\hat\psi_{1}(z_i) - \hat\psi_{0}(z_i)] \label{proof-tau}\\
    \hat\pi(s) & = \mathbbm{E}_n[s(X_i)\hat\psi_{1}(z_i) + (1-s(X_i))\hat\psi_{0}(z_i)] \label{proof-2treatpi}
\end{flalign}
We can generalize Equation \ref{proof-2treatpi} for more than two treatments by incorporating the indicator function such that the function returns 1 if the policy outcome matching the treatment and 0 otherwise:
\begin{align}
\hat\pi(s) & = \mathbb{E}_{n}[\sum_{t \in T}\mathbbm{1}\{s(X_i)=t\}\hat\psi_{t}(z_i)] \label{proof-pi}
\end{align}
Considering that the output of the MIP-ANN from Proposition \ref{prop:reform_bin} with Constraint \ref{strictly one treatment} gives us one-hot encoded vector where the node associated with the assigned treatment equals 1 and all other nodes are 0, this is precisely the indicator function we use in Equation \ref{proof-pi}. Thus, maximizing over this function under the parameters $z_i$ given in the data, we arrive at the desired objective
\begin{align}
 \max \frac{1}{n}\sum_{i=1}^{n}\sum_{t \in T}h_{i,t,L}\hat\psi(z_i)_{t}
\end{align}
Where $h_{i,t,L}$ is the $t^{\text{th}}$ node of the output layer for the $i^{\text{th}}$ individual in the vector output by the ANN.

The proof of the objective given a baseline policy $s_0$, given by Equation \ref{cate compare obj}, follows identically as above.
\endproof

\subsection{Proof of Proposition \ref{prop:lsc}}
\proof
We note that the argument for the lower semi-continuity of $\varphi_n(\theta,Z)$ and $\varphi(\theta,Z)$ is essentially the same so we will focus on showing $\varphi(\theta,Z)$ is lower semi-continuous.

Using the result of Propositions \ref{prop:reform_bin}-\ref{prop:pi obj}, we can see that $\varphi(\theta,Z)$ can be expressed as the value function of a special case of the formulation with $n=1$ and $\hat{\psi}$ replaced with $\psi$. In particular, suppose that $\theta = (\{\alpha^\theta_{d,k,0}, \alpha^\theta_{k',k,\ell},\alpha^\theta_{k',t,L}\}_{\forall k,k' \in [K]_1, \ell \in [L-1]}, \{\beta^\theta_{k,\ell}\}_{\forall k \in [K]_1, \ell \in [L]})$, then:
\begin{subequations}
    \begin{align}
        \varphi(\theta,Z) = \min - \sum_{t \in \mathcal{T}} h_{t,L}\psi(Z)_t \\
        \text{subject to: } \eqref{h0_define C2} - \eqref{zL_define C3}  \\
        \sum_{t \in \mathcal{T}} h_{t,L} = 1 \\
        \alpha_{d,k,0} = \alpha^\theta_{d,k,0}, \alpha_{k',k,\ell} = \alpha^\theta_{k',k,\ell}, \alpha_{k',t,L} = \alpha^\theta_{k',t,L}, & \forall k,k' \in [K]_1, \ell \in [L-1] \label{eq:alpha_theta_const}\\
        \beta_{k,\ell} = \beta^\theta_{k,\ell}, & \forall k \in [K]_1, \ell \in [L].  
    \end{align}
\end{subequations}
While this formulation is integer linear for fixed $Z$ or $\theta$ note that the way the input layer is processed by Constraints \eqref{h0_define C2} results in bi-linear terms from the products $\alpha_{d,k,0}X_d$. However using Assumption \ref{ass:input_assump} we can reformulate these products to remove the bi-linearity. Let us index the elements of set $\mathcal{A}$ as follows: $\mathcal{A} = \{\alpha_1, \alpha_2, ..., \alpha_m \}$, then let us define a new set of binary variables $\nu_{i,k,d} \in \mathbb{B}$ and new continuous variables $z^0_{d,k,i},z^0_{d,k} \in \mathbb{R}$ for $i \in \{1,...,m\}$ and we can rewrite \eqref{h0_define C2} as:
\begin{align}
    &\sum_{d \in \mathcal{F}} z^0_{d,k} + \beta_{k,0} \leq M h_{k,0} & \forall k\in [K]_1  \\
    &\sum_{d \in \mathcal{F}} z^0_d + \beta_{k,0} \geq \epsilon + (-M-\epsilon)(1- h_{k,0}) & \forall k\in [K]_1 \\
    &-M(1-\nu_{i,k,d}) + \alpha_i x_d \leq z^0_{d,k,i} \leq \alpha_i x_d + M(1-\nu_{i,k,d}), & \forall i \in [m]_1, d \in \mathcal{F}, k \in [K]_1\\ 
    & -M\nu_{i,k,d} \leq z^0_{d,k,i} \leq M\nu_{i,k,d}, & \forall i \in [m]_1, d \in \mathcal{F}, k \in [K]_1\\
    & \sum_{i=1}^m \nu_{i,k,d} = 1, & \forall k \in [K]_1, d \in \mathcal{F}\\ 
    & \sum_{i=1}^m z^0{d,k,i} = z^0_{d,k}, & \forall k \in [K]_1, d \in \mathcal{F}.
\end{align}
Finally we can reformulate the first portion of \eqref{eq:alpha_theta_const} with respect to $\nu_{i,k,d}$ as:
\begin{align}
  \sum_{i=1}^m\alpha_i \nu_{i,k,d} = \alpha_{k,d}^\theta, & \ \forall k \in [K]_1, d \in \mathcal{F}. 
\end{align}

Since these reformulations imply that the corresponding optimization problem to $\varphi(\theta,Z)$ depends on $\theta,Z$ only as affine terms in the constraints and a continuous function in the objective, and that this optimization problem is a mixed integer linear minimization problem applying Berge's Maximum Theorem \citep{berge1963espaces,walker1979generalization} in conjunction with the results from \cite{ralphs2014value} implies that $\varphi(\theta,Z)$ is lower semi-continuous on the domain $\Theta \times \mathcal{Z}$. As previously mentioned a similar reformulation argument can be used to show that $\varphi_n(\theta,Z)$ is also lower semi-continuous on the domain $\Theta \times \mathcal{Z}$.
\endproof
\subsection{Proof of Proposition \ref{prop:pointwise_lln}}
\proof
To proceed with the proof, note that by construction of the training problem in \eqref{eq:general_form} (and thus by extension its reformulation as a MIP), the problem will be feasible for any $\theta \in \Theta$, this means that the first conditions of the proposition are all satisfied. Thus we proceed to show the convergence result.

To prove that $H^\epsilon(\theta,\cdot)$ is a lower semi-continuous approximation to $H^\epsilon(\theta)$ we will first need to show that for any given sequence of data, the two quantities converge in probability. Then we can use the result of Proposition \ref{prop:lsc} in conjunction with Proposition 5.1 from \cite{vogel2003continuous} to turn this initial point-wise condition into a global condition over the set of all possible data observations. 

Thus we proceed to show the point-wise convergence in probability to complete the proof. Using the formal definition of convergence in probability our goal is to show that $|H_n^\epsilon(\theta,Z) - H^\epsilon(\theta)| \overset{p} {\rightarrow}0$. Since the data observations are i.i.d. and $H_n^\epsilon$ is constructed using an arithmetic mean we have the following relationship:
\begin{equation}
    \mathbb{E}[H_n^\epsilon(\theta,Z)] = \frac{1}{n} \sum_{i=1}^n \mathbb{E}[\inf_{\theta' \in U_\epsilon(\theta)} \varphi_n(\theta',z_i)] = \mathbb{E}[\inf_{\theta' \in U_\epsilon(\theta)} \varphi_n(\theta',Z)].
\end{equation}
The last equality follows from the i.i.d. assumption. We can thus upper bound our target expression using the triangle inequality:
\begin{multline}
    |H_n^\epsilon(\theta,Z) - H^\epsilon(\theta)| \leq \\ \underbrace{|H_n^\epsilon(\theta,Z) -\mathbb{E}[\inf_{\theta' \in U_\epsilon(\theta)} \varphi_n(\theta',Z)]|}_{\text{(a)}} + \underbrace{|\mathbb{E}[\inf_{\theta' \in U_\epsilon(\theta)} \varphi_n(\theta',Z)] - H^\epsilon(\theta)| }_{\text{(b)}}
\end{multline}
First consider expression (b), by Assumption \ref{ass:consistancy_assump} this  expression must converge in probability to zero. Thus if we can show that expression (a) also converges in probability to zero we can apply Slutsky's theorem to complete the proof \citep{keener2010theoretical}.

Observe that by assumption $\forall Z \in \mathcal{Z}$ we have that  $|\inf_{\theta' \in U_\epsilon(\theta)} \varphi_n(\theta',z)| \leq \text{diam}(\mathcal{Y})$, where $\text{diam}(\mathcal{Y}) = \max\{ |y-y'| : y,y' \in \mathcal{Y}\}$. Therefore both $\mathbb{E}[ \inf_{\theta' \in U_\epsilon(\theta)} \varphi_n(\theta',Z)] < \infty$ and $\text{Var}(\inf_{\theta' \in U_\epsilon(\theta)} \varphi_n(\theta',Z)) <\infty$, which means we can apply Kolmogorov's law of large numbers \citep{keener2010theoretical} to obtain that (a) must indeed converge in probability to zero.  
\endproof
\subsection{Proof of Proposition \ref{prop:val_func_lsc}}
\proof
 Since the results of Proposition \ref{prop:pointwise_lln} hold $\forall \theta \in \Theta$ and $\forall \epsilon >0$, we have that by Theorem 3.2 from \cite{vogel2003continuouspt2} $\Phi_n \overset{l-prob}{\underset{\{\theta\}}{\rightarrow}} \Phi$, that is it is a point-wise lower semi-continuous approximation in probability $\forall \theta \in \Theta$. Note that as a direct consequence of Proposition \ref{prop:lsc} both $\Phi,\Phi_n$ are lower semi-continuous functions of $\theta$ since semi-continuity is preserved under addition and multiplication by positive scalars. Thus applying Proposition 5.1 from \cite{vogel2003continuous}  proves the desired result.
\endproof
\subsection{Proof of Theorem \ref{thm:main_consistant}}
\proof
Because Proposition \ref{prop:val_func_lsc} implies that the empirical training problem's value function is lower semi-continuous approximation in probability to the target problem, we can apply Theorem 4.3 from \citep{vogel2003continuous} to obtain the main result of the proposition.
\endproof

\section{Prescriptive ANN Experiments on Warfarin Dosing}
\subsection{Experimental Setup}\label{setup:warfarin}

We now employ the prescriptive models to learn personalized warfarin dosing. Warfarin is an oral anticoagulant prescribed to reduce the risk of blood clots. While its usage is commonplace, prescribing the appropriate dosage can be difficult because an individual's response to the anticoagulant varies greatly based on their genetic, clinical, and demographic characteristics \citep{international2009estimation}. 

We use the publicly available dataset curated by the International Warfarin Pharmacogenetics Consortium and published by the Pharmacogenetics and Pharmacogenomics Knowledge Base. The dataset comprises of information for 5700 individuals including single nucleotide polymorphisms (SNPs) in genes for cytochrome P450 2C9 (CYP2C9) and vitamin K epoxide reductase (VKORC1), genetic information known to correlate with warfarin requirements, demographic information such as gender, race, and age, medical information such as which medication they were on at the time of the study and their reason for receiving warfarin as a treatment, treatment information regarding the patient's target international normalised ratio (INR), the therapeutic dose of warfarin, and whether or not their INR on receiving said dose \citep{international2009estimation}. The therapeutic warfarin doses reported in the dataset are the optimal stable doses for each patient, found by controlled experimentation conducted by physicians, which allows us to compare the learned policies' performances against the true optimal policy given. For our analysis, we remove patients that had missing entries in any of the covariates listed above. The final dataset included patient information for 3398 individuals. We used a discretization of the dosage as the treatment, as is standard with literature that study the same dataset \citep{international2009estimation, jo2021learning, bertsimas2019optimal, kallus2017recursive}: $t_i = 0 \ (\leq 3 \text{ mg/day}), \ 1 \ (> 3 \text{ and} < 5 \text{ mg/day}), 2 \ (\geq 5 \text{ mg/day})$. As for the observed outcome, we consider the absolute different between INR on Reported Therapeutic Dose of Warfarin from Target INR. 
Any continuous covariate was binarized and any categorical covariate
was split into one-hot encoded binary vectors for the prescriptive tree models. Furthermore, we performed the data imputation steps recommended by \cite{international2009estimation}. The dataset was finally split on 5 random seeds with a 70:30 train:test ratio. We used a 5-fold cross validation scheme to fine-tune the $\ell_0$ regularization hyperparameter for the PNN models.

In order to estimate the propensity scores for treatments in this dataset, we used 10-fold cross validation to evaluate logistic regression models, decision trees and random forests. Similarly, to estimating the counnterfactual outcomes for patients in the dataset and compare elastic net rergessors, random forests and multilayer perceptrons. We conclude that the logistic regression is ideal to estimate the propensity scores, and that elastic net is best for estimating the counterfactual outcomes of the patients.

The policies constructed by each prescriptive model are evaluated using $\pi(s)$, where a function output closer to 0 corresponds to a policy where most patients were prescribed a dosage that led to their INR being close to their target INR. The results reported in the following section are the averages of $\pi(s)$ over the 5 seeds.

 \subsection{Results}\label{results:warfarin}

 We present a summary of our results for the personalized warfarin dosing experiments in Figures \ref{warfarin_barplots} and \ref{warfarin_heatmaps} and a complete table of results can be found in Appendix \ref{completedata}. 

\begin{figure*}[ht]
    \centering
    \includegraphics[width=1\textwidth]{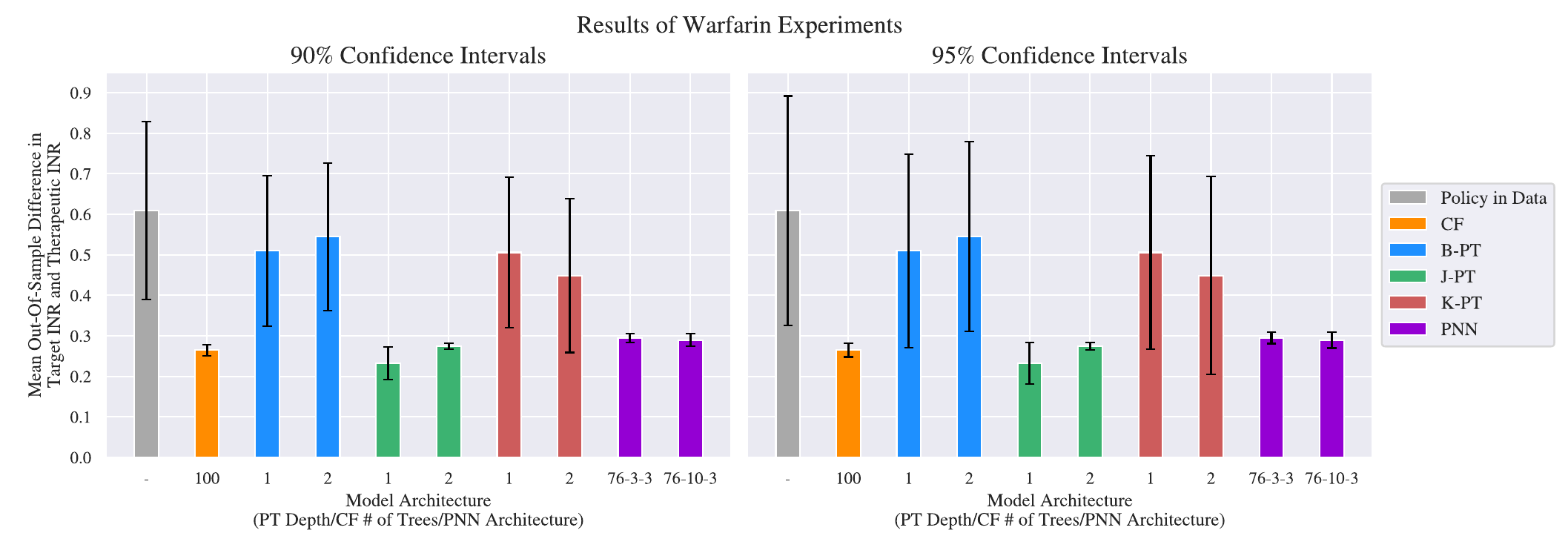}
    \caption{Results of experiments on the warfarin dataset for out-of-sample patients prescribed treatments based on policies learned by different models. The plots display the average outcome. The left plot include the 90\% confidence intervals and the right plot include the 95\% confidence intervals.
    }
    \label{warfarin_barplots}
\end{figure*}
 
 \begin{figure*}[ht]
    \centering
    \includegraphics[width=1\textwidth]{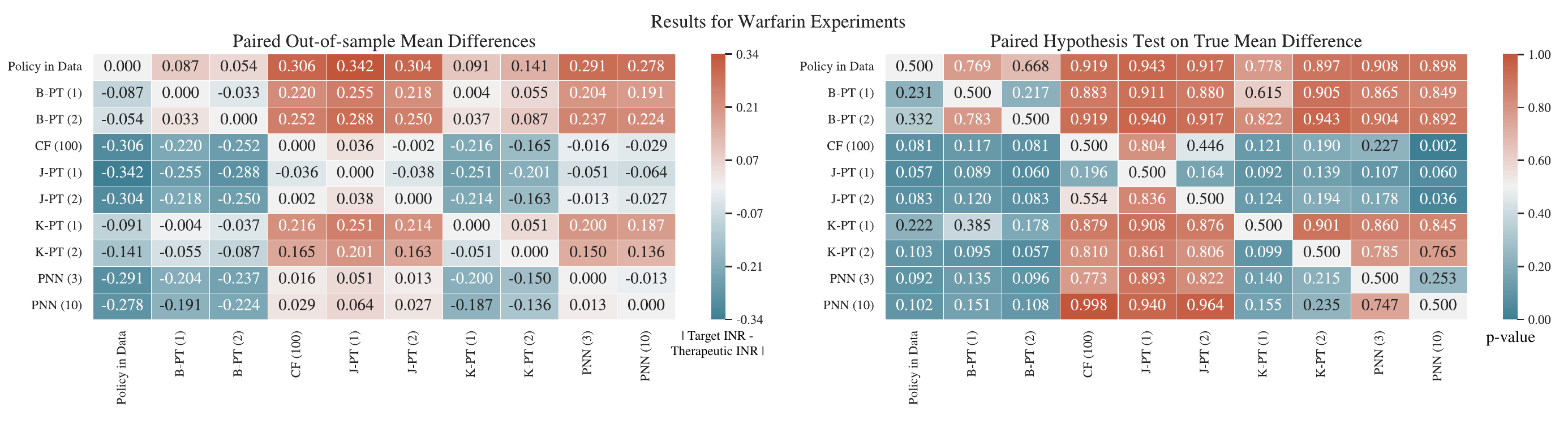}
    \caption{Paired results of experiments on the warfarin dataset for out-of-sample patients prescribed treatments based on policies learned by different models. Each panel displays paired evaluation of the various models. Left display the point differences of out-of-sample average outcome. Differences are taken as \textit{row - column}. Right displays the hypothesis tests on the differences of out-of-sample average outcome between every pair of models. Smaller p-values favor models in the rows. Cool colors denote lower blood pressures (left) or smaller p-values (right). Axes values denote the model (PT depth/\# of trees in CF/PNN width).}
    \label{warfarin_heatmaps}
\end{figure*}
 
 Recall that in these experiments, the models work to find the optimal warfarin dosing policy such that the difference between an individual's target INR and their INR post receiving warfarin is minimized. We see that both figures indicate that every learning model outperforms the policy in the data evaluated with the doubly robust estimand for the counterfactual outcome. However, we know that the policy in the data contains the true optimal dosage for each patient. We note that the best performing models in Figures \ref{warfarin_barplots} and \ref{warfarin_heatmaps}, the CF, J-PT and PNN, are those models that incorporate the doubly robust estimator in their optimization problem. By minimizing over this estimate, these models heavily rely on the doubly robust estimator in estimating the counterfactual outcomes in the data. In this case, while a logistic regression model yielded the best score for estimating the propensity scores in our cross-validation regime, the accuracy of the model was still only 66.77\%. Similarly, the root mean squared error of the elastic net model that was used to estimate the counterfactual outcomes with the DM method was 2.0558 on average across the three treatments (the typical effective therapeutic INR range for individuals taking warfarin is between 2.0–3.0 and 2.5–3.5 \citep{dumont2013warfarin}). This observation is maintained considering that the B-PT and K-PT models estimate the counterfactual independent from the doubly robust estimation model. Nevertheless, the left plot in Figure \ref{warfarin_heatmaps} indicates that the paired out-of-sample mean differences for every pair is within  $\pm 0.35$, and so, given the typical therapeutic INR range, the policies produced by every model are relatively similar to the optimal policy in the data. 

 These results shed light on the limitations of the doubly robust estimation technique, and the models that utilize it their optimization problem. We also observed that the 3- and 10-width PNN models terminated with an average of 74.6\% and 92.78\% optimality gap, respectively. This means PNNs would require addtional computational resources to provide effectiver policies given the high-dimensional feature vector and the large number of samples in this setting. Considering that the performance of the PNN models are competitive given current resources, we expect even stronger performance when additional resources are available.

\section{Additional Experimental Results}\label{resultssappendix}

In this section, we provide additional experimental results that supplement our analysis.

\subsection{Complete Experiment Results}\label{completedata}

\subsubsection{Simulated Data}
The table in this section display the average out-of-sample probability of correct treatment assignment, average optimality gap, and average running time in seconds of the models in the simulated data experiments. The averages are taken over the 5 seeds in the experiments, with the standard deviation given parenthetically. The results of experiments with 100 and 500 training points are included. 

\begin{table*}[ht]
  \centering
  \resizebox{15cm}{!}{%
    \begin{tabular}{ccclllllllllllllll}
    \toprule
    \multicolumn{1}{c}{\multirow{3}[6]{1cm}{\centering Train Data Length}} & \multicolumn{1}{c}{\multirow{3}[6]{1cm}{\centering Model}} & \multicolumn{1}{c}{\multirow{3}[6]{1cm}{\centering Architec -ture}} & \multicolumn{15}{c}{\centering Probability of Correct Assignments} \\
\cmidrule{4-18}          &       &       & \multicolumn{3}{c}{0.1} & \multicolumn{3}{c}{0.25} & \multicolumn{3}{c}{0.5} & \multicolumn{3}{c}{0.75} & \multicolumn{3}{c}{0.9} \\
\cmidrule{4-18}          &       &       & \multicolumn{1}{c}{OOSP (\%)} & \multicolumn{1}{c}{Gap (\%)} & \multicolumn{1}{c}{Runtime (s)} & \multicolumn{1}{c}{OOSP (\%)} & \multicolumn{1}{c}{Gap (\%)} & \multicolumn{1}{c}{Runtime (s)} & \multicolumn{1}{c}{OOSP (\%)} & \multicolumn{1}{c}{Gap (\%)} & \multicolumn{1}{c}{Runtime (s)} & \multicolumn{1}{c}{OOSP (\%)} & \multicolumn{1}{c}{Gap (\%)} & \multicolumn{1}{c}{Runtime (s)} & \multicolumn{1}{c}{OOSP (\%)} & \multicolumn{1}{c}{Gap (\%)} & \multicolumn{1}{c}{Runtime (s)} \\
    \midrule
    \multirow{9}[10]{*}{100} & \multirow{2}[2]{*}{B-PT} & 1     & 58.34 (18.42) & 0 (0) & 1.51 (0.22) & 62.37 (18.27) & 0 (0) & 1.08 (0.59) & 53.99 (8.9) & 0 (0) & 1.89 (0.38) & 50.09 (0.59) & 0 (0) & 1.27 (0.32) & 44.16 (13.49) & 0 (0) & 1.02 (0.33) \\
          &       & 2     & 81.15 (12.77) & 0 (0) & 45.27 (3.42) & 69.44 (17.74) & 0 (0) & 64.91 (15.69) & 75.58 (14.57) & 0 (0) & 90.23 (12.92) & 53.4 (10.66) & 0 (0) & 71.5 (11.76) & 39.86 (7.86) & 0 (0) & 52.97 (12.99) \\
\cmidrule{2-18}          & CF    & 100   & 78.59 (16.92) & -     & 0.65 (0.01) & 89.96 (1.48) & -     & 0.63 (0.01) & 90.46 (1.1) & -     & 0.63 (0.01) & 90.48 (0.57) & -     & 0.62 (0.01) & 84.03 (14.95) & -     & 0.64 (0.02) \\
\cmidrule{2-18}          & \multirow{2}[2]{*}{J-PT} & 1     & 86.99 (4.4) & 0 (0) & 0.11 (0.02) & 89.3 (1.9) & 0 (0) & 0.1 (0.02) & 89.3 (1.9) & 0 (0) & 0.09 (0.01) & 89.97 (1.86) & 0 (0) & 0.09 (0.01) & 89.97 (1.77) & 0 (0) & 0.09 (0.02) \\
          &       & 2     & 87.33 (4.44) & 0 (0) & 1.38 (1) & 88.29 (2.02) & 0 (0) & 2.75 (1.42) & 88.76 (1.19) & 0 (0) & 1.6 (1.06) & 87.88 (1.98) & 0 (0) & 2.27 (1.57) & 88.2 (4.39) & 0 (0) & 1.7 (1.58) \\
\cmidrule{2-18}          & \multirow{2}[2]{*}{K-PT} & 1     & 73.33 (15.59) & 0 (0) & 0.39 (0.11) & 73.98 (16.4) & 0 (0) & 0.47 (0.08) & 76.51 (16.91) & 0 (0) & 1.06 (0.41) & 56.03 (8.95) & 0 (0) & 0.84 (0.52) & 42.07 (12.99) & 0 (0) & 0.44 (0.27) \\
          &       & 2     & 83.58 (8.02) & 0 (0) & 16.83 (5.82) & 79.02 (12.81) & 0 (0) & 17.85 (5.36) & 83.96 (5.49) & 0 (0) & 29.77 (6.66) & 60.08 (14.19) & 0 (0) & 26.82 (3.57) & 35.37 (13.88) & 0 (0) & 14.52 (3.51) \\
\cmidrule{2-18}          & \multirow{2}[2]{*}{PNN} & 2-3-2 & 88.93 (3.42) & 0 (0) & 22.56 (14.52) & 89.75 (3.39) & 0 (0) & 16.02 (10.87) & 91.27 (0.15) & 0 (0) & 11.61 (2.69) & 90.08 (2.03) & 0 (0) & 14.85 (5.4) & 91.28 (0.15) & 0 (0) & 14.79 (10.02) \\
          &       & 2-10-2 & 88.93 (3.43) & 0.02 (0.02) & 3232.61 (821.74) & 89.8 (3.32) & 0.01 (0.01) & 2885.75 (1018.32) & 91.25 (0.15) & 0.01 (0.01) & 3080.71 (1161.3) & 90.14 (1.89) & 0.02 (0.02) & 3268.12 (587.45) & 91.26 (0.15) & 0.02 (0.02) & 2193.95 (1369.05) \\
    \midrule
    \multirow{9}[10]{*}{500} & \multirow{2}[2]{*}{B-PT} & 1     & 49.8 (0.61) & 0 (0) & 4.77 (1.97) & 57.7 (17.1) & 0 (0) & 7.13 (3.13) & 58.53 (18.56) & 0 (0) & 5.11 (4.71) & 50.2 (0.61) & 0 (0) & 5.54 (3.55) & 50.2 (0.61) & 0 (0) & 4.04 (2.61) \\
          &       & 2     & 74.33 (20.4) & 0 (0) & 502.1 (92.75) & 83.2 (12.61) & 0 (0) & 526.6 (146.83) & 85.52 (8.89) & 0 (0) & 570.98 (78.21) & 53.34 (2.53) & 0 (0) & 588.45 (97.26) & 49.36 (1.96) & 0 (0) & 562.64 (83.7) \\
\cmidrule{2-18}          & CF    & 100   & 90.95 (0.45) & -     & 0.69 (0.02) & 91.21 (0.4) & -     & 0.68 (0.01) & 91.17 (0.3) & -     & 0.66 (0.02) & 90.9 (0.85) & -     & 0.69 (0.02) & 89.96 (1.56) & -     & 0.7 (0.01) \\
\cmidrule{2-18}          & \multirow{2}[2]{*}{J-PT} & 1     & 91.41 (0.24) & 0 (0) & 1.13 (0.17) & 91.41 (0.24) & 0 (0) & 0.98 (0.18) & 91.41 (0.24) & 0 (0) & 0.84 (0.03) & 90.66 (1.85) & 0 (0) & 1.03 (0.17) & 90.66 (1.85) & 0 (0) & 1.21 (0.11) \\
          &       & 2     & 90.89 (0.44) & 0 (0) & 35.2 (8.09) & 90.95 (0.52) & 0 (0) & 32.71 (14.9) & 90.79 (0.38) & 0 (0) & 27.27 (14.23) & 90.58 (0.94) & 0 (0) & 35.11 (12.88) & 89.75 (0.72) & 0 (0) & 39.09 (3.29) \\
\cmidrule{2-18}          & \multirow{2}[2]{*}{K-PT} & 1     & 80.73 (17.71) & 0 (0) & 2.67 (0.68) & 88.39 (4.91) & 0 (0) & 2.4 (0.85) & 84.79 (9.71) & 0 (0) & 3.4 (0.62) & 52.17 (4.44) & 0 (0) & 3 (0.26) & 50.2 (0.61) & 0 (0) & 1.83 (1.02) \\
          &       & 2     & 87.93 (4.11) & 0 (0) & 134.46 (47.93) & 87.71 (4.63) & 0 (0) & 148.33 (35.1) & 83.78 (9.52) & 0 (0) & 146.02 (20.47) & 58.5 (7.11) & 0 (0) & 157.8 (46.88) & 48.92 (2.04) & 0 (0) & 140.41 (34.6) \\
\cmidrule{2-18}          & \multirow{2}[2]{*}{PNN} & 37290 & 90.79 (0.58) & 0.05 (0.04) & 2996.29 (1350.27) & 91.21 (0.1) & 0.02 (0.03) & 2721.05 (1120.22) & 91.21 (0.08) & 0.02 (0.02) & 2764.36 (1144.47) & 91.28 (0.15) & 0.05 (0.04) & 3514.17 (192.23) & 91.21 (0.13) & 0.11 (0.04) & 3600.13 (0.1) \\
          &       & 2-10-2 & 91 (0.34) & 0.06 (0.01) & 3600.15 (0.07) & 91.21 (0.09) & 0.06 (0.01) & 3601.7 (3.09) & 91.2 (0.08) & 0.06 (0.01) & 3600.47 (0.27) & 90.96 (0.69) & 0.06 (0.01) & 3600.93 (1.66) & 91.2 (0.12) & 0.08 (0.01) & 3600.6 (0.48) \\
    \bottomrule
    \end{tabular}%
    }
    \caption{Complete Results Simulated Experiments with Athey Model 1 (with Adapted Binarization).}
  \label{simulatedfull1e}%
\end{table*}%

\begin{table*}[ht]
  \centering
  \resizebox{15cm}{!}{%
    \begin{tabular}{ccclllllllllllllll}
    \toprule
    \multicolumn{1}{c}{\multirow{3}[6]{1cm}{\centering Train Data Length}} & \multicolumn{1}{c}{\multirow{3}[6]{1cm}{\centering Model}} & \multicolumn{1}{c}{\multirow{3}[6]{1cm}{\centering Architec -ture}} & \multicolumn{15}{c}{\centering Probability of Correct Assignments} \\
\cmidrule{4-18}          &       &       & \multicolumn{3}{c}{0.1} & \multicolumn{3}{c}{0.25} & \multicolumn{3}{c}{0.5} & \multicolumn{3}{c}{0.75} & \multicolumn{3}{c}{0.9} \\
\cmidrule{4-18}          &       &       & \multicolumn{1}{c}{OOSP (\%)} & \multicolumn{1}{c}{Gap (\%)} & \multicolumn{1}{c}{Runtime (s)} & \multicolumn{1}{c}{OOSP (\%)} & \multicolumn{1}{c}{Gap (\%)} & \multicolumn{1}{c}{Runtime (s)} & \multicolumn{1}{c}{OOSP (\%)} & \multicolumn{1}{c}{Gap (\%)} & \multicolumn{1}{c}{Runtime (s)} & \multicolumn{1}{c}{OOSP (\%)} & \multicolumn{1}{c}{Gap (\%)} & \multicolumn{1}{c}{Runtime (s)} & \multicolumn{1}{c}{OOSP (\%)} & \multicolumn{1}{c}{Gap (\%)} & \multicolumn{1}{c}{Runtime (s)} \\
    \midrule
    \multirow{9}[10]{*}{100} & \multirow{2}[2]{*}{B-PT} & 1     & 51.56 (3.63) & 0 (0) & 0.93 (0.26) & 51.39 (3.99) & 0 (0) & 1.13 (0.32) & 51.83 (4.38) & 0 (0) & 0.84 (0.11) & 48.13 (4.4) & 0 (0) & 0.85 (0.06) & 43.81 (4.9) & 0 (0) & 0.85 (0.27) \\
          &       & 2     & 58.38 (0.85) & 0 (0) & 19.41 (4.42) & 56.68 (7.8) & 0 (0) & 22.86 (3.88) & 53.85 (4.53) & 0 (0) & 24.8 (8.13) & 47.69 (4.82) & 0 (0) & 21.75 (5.7) & 35.27 (4.59) & 0 (0) & 19.8 (3.69) \\
\cmidrule{2-18}          & CF    & 100   & 78.59 (16.92) & -     & 0.64 (0.01) & 89.96 (1.48) & -     & 0.63 (0.01) & 90.46 (1.1) & -     & 0.62 (0.02) & 90.48 (0.57) & -     & 0.62 (0.01) & 84.03 (14.95) & -     & 0.65 (0.01) \\
\cmidrule{2-18}          & \multirow{2}[2]{*}{J-PT} & 1     & 60.21 (0.59) & 0 (0) & 0.09 (0.02) & 59.81 (0.45) & 0 (0) & 0.09 (0.01) & 60.07 (0.65) & 0 (0) & 0.08 (0) & 59.67 (0.36) & 0 (0) & 0.09 (0.01) & 60.07 (0.65) & 0 (0) & 0.08 (0) \\
          &       & 2     & 70 (0.88) & 0 (0) & 0.52 (0.19) & 69.79 (0.82) & 0 (0) & 0.43 (0.15) & 70.18 (0.69) & 0 (0) & 0.46 (0.2) & 69.83 (1.05) & 0 (0) & 0.41 (0.15) & 70.26 (0.62) & 0 (0) & 0.41 (0.14) \\
\cmidrule{2-18}          & \multirow{2}[2]{*}{K-PT} & 1     & 54 (5.17) & 0 (0) & 0.21 (0.2) & 56.01 (4.84) & 0 (0) & 0.4 (0.19) & 58.28 (3.39) & 0 (0) & 0.74 (0.43) & 49.46 (7.27) & 0 (0) & 0.36 (0.39) & 41.8 (2.97) & 0 (0) & 0.07 (0.01) \\
          &       & 2     & 58.83 (7.29) & 0 (0) & 5.49 (0.85) & 61.01 (4.17) & 0 (0) & 6.02 (1.48) & 59.54 (1.87) & 0 (0) & 7.54 (1.55) & 48.7 (5.54) & 0 (0) & 7.95 (1.41) & 38.27 (2.83) & 0 (0) & 4.63 (1.66) \\
\cmidrule{2-18}          & \multirow{2}[2]{*}{PNN} & 2-3-2 & 88.93 (3.42) & 0 (0) & 22.56 (14.52) & 89.75 (3.39) & 0 (0) & 16.02 (10.87) & 91.27 (0.15) & 0 (0) & 11.61 (2.69) & 90.08 (2.03) & 0 (0) & 14.85 (5.4) & 91.28 (0.15) & 0 (0) & 14.79 (10.02) \\
          &       & 2-10-2 & 88.93 (3.43) & 0.02 (0.02) & 3232.61 (821.74) & 89.8 (3.32) & 0.01 (0.01) & 2885.75 (1018.32) & 91.25 (0.15) & 0.01 (0.01) & 3080.71 (1161.3) & 90.14 (1.89) & 0.02 (0.02) & 3268.12 (587.45) & 91.26 (0.15) & 0.02 (0.02) & 2193.95 (1369.05) \\
    \midrule
    \multirow{9}[10]{*}{500} & \multirow{2}[2]{*}{B-PT} & 1     & 49.8 (0.61) & 0 (0) & 20.42 (1.98) & 49.93 (0.33) & 0 (0) & 23.71 (4.34) & 50.5 (0.51) & 0 (0) & 24.99 (2.45) & 50.2 (0.61) & 0 (0) & 21.33 (2.91) & 50.2 (0.61) & 0 (0) & 17.34 (2.33) \\
          &       & 2     & 52.03 (3.89) & 0 (0) & 718.07 (166.84) & 53.89 (4.94) & 0 (0) & 821.48 (52.94) & 54.96 (5.17) & 0 (0) & 753.47 (277.19) & 49.79 (0.51) & 0 (0) & 632.35 (67.21) & 49.89 (1.17) & 0 (0) & 614.77 (137) \\
\cmidrule{2-18}          & CF    & 100   & 90.95 (0.45) & -     & 0.69 (0.01) & 91.21 (0.4) & -     & 0.68 (0.02) & 91.17 (0.3) & -     & 0.67 (0.01) & 90.9 (0.85) & -     & 0.67 (0.02) & 89.96 (1.56) & -     & 0.69 (0.01) \\
\cmidrule{2-18}          & \multirow{2}[2]{*}{J-PT} & 1     & 60.12 (0.64) & 0 (0) & 1.02 (0.17) & 60.3 (0.69) & 0 (0) & 0.94 (0.04) & 60.12 (0.64) & 0 (0) & 0.95 (0.03) & 60.35 (0.64) & 0 (0) & 0.95 (0.03) & 60.35 (0.64) & 0 (0) & 1.41 (0.48) \\
          &       & 2     & 70.36 (0.64) & 0 (0) & 7.49 (2.13) & 70.02 (0.62) & 0 (0) & 5.39 (2.57) & 70.36 (0.64) & 0 (0) & 4.1 (1.18) & 69.91 (0.74) & 0 (0) & 5.91 (3.57) & 69.91 (0.74) & 0 (0) & 7.07 (3.76) \\
\cmidrule{2-18}          & \multirow{2}[2]{*}{K-PT} & 1     & 58.2 (2.72) & 0 (0) & 5.81 (0.75) & 58.24 (2.73) & 0 (0) & 5.83 (0.55) & 59.89 (0.59) & 0 (0) & 5.15 (1.04) & 50.85 (5.97) & 0 (0) & 5.58 (1.3) & 48.22 (2.31) & 0 (0) & 5.17 (1.04) \\
          &       & 2     & 61.83 (6.67) & 0 (0) & 138.4 (31.34) & 67.94 (2.83) & 0 (0) & 138.91 (27.41) & 69.57 (1.38) & 0 (0) & 149.05 (24.19) & 50.3 (8.92) & 0 (0) & 135.3 (11.87) & 48.5 (2.24) & 0 (0) & 144.82 (28.23) \\
\cmidrule{2-18}          & \multirow{2}[2]{*}{PNN} & 2-3-2 & 90.79 (0.58) & 0.05 (0.04) & 2996.29 (1350.27) & 91.21 (0.1) & 0.02 (0.03) & 2721.05 (1120.22) & 91.21 (0.08) & 0.02 (0.02) & 2764.36 (1144.47) & 91.28 (0.15) & 0.05 (0.04) & 3514.17 (192.23) & 91.21 (0.13) & 0.11 (0.04) & 3600.13 (0.1) \\
          &       & 2-10-2 & 91 (0.34) & 0.06 (0.01) & 3600.15 (0.07) & 91.21 (0.09) & 0.06 (0.01) & 3601.7 (3.09) & 91.2 (0.08) & 0.06 (0.01) & 3600.47 (0.27) & 90.96 (0.69) & 0.06 (0.01) & 3600.93 (1.66) & 91.2 (0.12) & 0.08 (0.01) & 3600.6 (0.48) \\
    \bottomrule
    \end{tabular}%
    }
    \caption{Complete Results of Simulated Experiments with Athey Model 1 (with One-Hot Binarization).}
  \label{simulatedfull1h}%
\end{table*}%

\begin{table*}[ht]
  \centering
  \resizebox{15cm}{!}{%
    \begin{tabular}{ccclllllllllllllll}
    \toprule
    \multicolumn{1}{c}{\multirow{3}[6]{1cm}{\centering Train Data Length}} & \multicolumn{1}{c}{\multirow{3}[6]{1cm}{\centering Model}} & \multicolumn{1}{c}{\multirow{3}[6]{1cm}{\centering Architec -ture}} & \multicolumn{15}{c}{\centering Probability of Correct Assignments} \\
\cmidrule{4-18}          &       &       & \multicolumn{3}{c}{0.1} & \multicolumn{3}{c}{0.25} & \multicolumn{3}{c}{0.5} & \multicolumn{3}{c}{0.75} & \multicolumn{3}{c}{0.9} \\
\cmidrule{4-18}          &       &       & \multicolumn{1}{c}{OOSP (\%)} & \multicolumn{1}{c}{Gap (\%)} & \multicolumn{1}{c}{Runtime (s)} & \multicolumn{1}{c}{OOSP (\%)} & \multicolumn{1}{c}{Gap (\%)} & \multicolumn{1}{c}{Runtime (s)} & \multicolumn{1}{c}{OOSP (\%)} & \multicolumn{1}{c}{Gap (\%)} & \multicolumn{1}{c}{Runtime (s)} & \multicolumn{1}{c}{OOSP (\%)} & \multicolumn{1}{c}{Gap (\%)} & \multicolumn{1}{c}{Runtime (s)} & \multicolumn{1}{c}{OOSP (\%)} & \multicolumn{1}{c}{Gap (\%)} & \multicolumn{1}{c}{Runtime (s)} \\
    \midrule
    \multirow{9}[10]{*}{100} & \multirow{2}[2]{*}{B-PT} & 1     & 48.35 (21) & 0 (0) & 17.6 (4.21) & 56.84 (24.71) & 0 (0) & 15.5 (3.94) & 65.29 (28.6) & 0 (0) & 15.84 (1.91) & 82.06 (3.67) & 0 (0) & 16.53 (3.63) & 80.62 (9.47) & 0 (0) & 17.25 (3.96) \\
          &       & 2     & 66.67 (16.28) & 0.9 (0.19) & 3600.05 (0.03) & 55.06 (23.48) & 1.38 (0.56) & 3600.05 (0.02) & 71.92 (23.12) & 1.58 (0.21) & 3600.04 (0.02) & 71.69 (3.79) & 1.55 (0.32) & 3600.02 (0.01) & 48.82 (11.58) & 1.28 (0.36) & 3600.03 (0.01) \\
\cmidrule{2-18}          & CF    & 100   & 85.05 (0.55) & -     & 0.8 (0.03) & 85.05 (0.54) & -     & 0.77 (0.02) & 85.05 (0.54) & -     & 0.71 (0.01) & 85.05 (0.54) & -     & 0.72 (0.01) & 85.05 (0.54) & -     & 0.78 (0.02) \\
\cmidrule{2-18}          & \multirow{2}[2]{*}{J-PT} & 1     & 84.25 (2.09) & 0 (0) & 0.28 (0.02) & 84.09 (2.18) & 0 (0) & 0.34 (0.08) & 84.09 (2.18) & 0 (0) & 0.34 (0.08) & 85.05 (0.54) & 0 (0) & 0.35 (0.06) & 84.25 (2.09) & 0 (0) & 0.41 (0.06) \\
          &       & 2     & 80.13 (3.2) & 0 (0) & 75.39 (32.19) & 80.39 (3.37) & 0 (0) & 105.34 (24.29) & 82.17 (2.4) & 0 (0) & 83.48 (16.31) & 81.66 (2.24) & 0 (0) & 95.21 (35.06) & 81.9 (1.78) & 0 (0) & 76.47 (28.1) \\
\cmidrule{2-18}          & \multirow{2}[2]{*}{K-PT} & 1     & 54.3 (22.95) & 0 (0) & 3.03 (1.15) & 58 (28.14) & 0 (0) & 2.4 (0.49) & 68.32 (22.86) & 0 (0) & 2.39 (0.63) & 79.79 (2.96) & 0 (0) & 2.8 (0.59) & 65.58 (18.83) & 0 (0) & 2.77 (0.95) \\
          &       & 2     & 71.21 (14.69) & 0 (0) & 689.69 (285.66) & 58.57 (20.66) & 0 (0) & 1415.73 (537.86) & 66.52 (15.66) & 0 (0) & 1913.22 (247.75) & 70.15 (5.69) & 0 (0) & 1635.12 (436.59) & 51.22 (4.96) & 0 (0) & 849.38 (320.42) \\
\cmidrule{2-18}          & \multirow{2}[2]{*}{PNN} & 10-3-2 & 85.05 (0.54) & 0 (0.01) & 888.36 (1517.13) & 85.05 (0.54) & 0.01 (0.02) & 1092.54 (1438.89) & 85.05 (0.54) & 0 (0) & 725.32 (1607.54) & 85.05 (0.54) & 0 (0) & 246.13 (313.25) & 85.05 (0.54) & 0 (0.01) & 853.77 (1546.05) \\
          &       & 10-10-2 & 85.05 (0.54) & 0.01 (0.01) & 3600.1 (0.05) & 85.05 (0.54) & 0.02 (0.01) & 3600.09 (0.09) & 85.05 (0.54) & 0.01 (0.01) & 3085.44 (1150.71) & 85.05 (0.54) & 0.01 (0.01) & 2483.05 (1580.02) & 85.05 (0.54) & 0.01 (0.01) & 3600.34 (0.61) \\
    \midrule
    \multirow{9}[10]{*}{500} & \multirow{2}[2]{*}{B-PT} & 1     & 57.47 (7.74) & 0 (0) & 243.19 (21.18) & 73.09 (16.05) & 0 (0) & 247.53 (48.23) & 84.97 (0.4) & 0 (0) & 218.83 (36.82) & 84.97 (0.4) & 0 (0) & 204.44 (14.8) & 84.97 (0.4) & 0 (0) & 238.79 (28.57) \\
          &       & 2     & 58.15 (36.55) & 8.64 (5.4) & 3600.02 (0.01) & 61.38 (33.29) & 7.58 (3.57) & 3600.02 (0.01) & 54.11 (36.02) & 9.43 (7.24) & 3600.02 (0.01) & 74.23 (20.98) & 8.39 (5.09) & 3600.01 (0) & 71.75 (25.99) & 7.07 (4.15) & 3600.07 (0.08) \\
\cmidrule{2-18}          & CF    & 100   & 84.97 (0.4) & -     & 0.76 (0.03) & 84.99 (0.4) & -     & 0.73 (0) & 84.99 (0.38) & -     & 0.72 (0.01) & 84.97 (0.4) & -     & 0.73 (0.02) & 84.97 (0.4) & -     & 0.74 (0.01) \\
\cmidrule{2-18}          & \multirow{2}[2]{*}{J-PT} & 1     & 84.97 (0.4) & 0 (0) & 3.5 (0.63) & 84.97 (0.4) & 0 (0) & 2.82 (0.17) & 84.97 (0.4) & 0 (0) & 3.65 (0.92) & 84.97 (0.4) & 0 (0) & 3.13 (0.21) & 84.97 (0.4) & 0 (0) & 3.64 (0.8) \\
          &       & 2     & 85 (0.17) & 0 (0) & 1088.41 (533.15) & 85.08 (0.34) & 0 (0) & 1221.06 (217.82) & 85.12 (0.37) & 0 (0) & 1730.08 (649.18) & 84.66 (0.71) & 0 (0) & 1242.12 (388.31) & 85.07 (0.4) & 0 (0) & 882.05 (347.48) \\
\cmidrule{2-18}          & \multirow{2}[2]{*}{K-PT} & 1     & 57.47 (7.74) & 0 (0) & 38.88 (2.61) & 62.21 (14.59) & 0 (0) & 33.48 (2.7) & 76.08 (11.64) & 0 (0) & 40.15 (10.33) & 84.97 (0.4) & 0 (0) & 35.37 (2.86) & 84.97 (0.4) & 0 (0) & 30.43 (4.12) \\
          &       & 2     & 73.51 (4.24) & 0.4 (0.08) & 3600.5 (0.11) & 77.08 (7.47) & 0.43 (0.07) & 3600.48 (0.2) & 81.76 (1.98) & 0.4 (0.1) & 3600.68 (0.42) & 78.37 (1.39) & 0.47 (0.09) & 3600.39 (0.19) & 80.25 (4.67) & 0.48 (0.1) & 3600.32 (0.06) \\
\cmidrule{2-18}          & \multirow{2}[2]{*}{PNN} & 10-3-2 & 84.97 (0.4) & 0.04 (0.01) & 3600.09 (0.03) & 84.97 (0.4) & 0.02 (0.01) & 3600.24 (0.23) & 84.97 (0.4) & 0 (0) & 2197.94 (1650.17) & 84.97 (0.4) & 0.02 (0) & 3600.12 (0.03) & 84.97 (0.4) & 0.04 (0.01) & 3600.2 (0.17) \\
          &       & 10-10-2 & 84.97 (0.4) & 0.03 (0.01) & 3600.7 (0.42) & 84.97 (0.4) & 0.02 (0.01) & 3605.46 (11.01) & 84.97 (0.4) & 0.01 (0) & 3604.76 (9.72) & 84.97 (0.4) & 0.02 (0) & 3601.56 (1.65) & 84.97 (0.4) & 0.03 (0.01) & 3605.48 (11.47) \\
    \bottomrule
    \end{tabular}%
    }
    \caption{Complete Results of Simulated Experiments with Athey Model 2 (with Adapted Binarization).}
  \label{simulatedfull2e}%
\end{table*}%

\begin{table*}[ht]
  \centering
  \resizebox{15cm}{!}{%
    \begin{tabular}{ccclllllllllllllll}
    \toprule
    \multicolumn{1}{c}{\multirow{3}[6]{1cm}{\centering Train Data Length}} & \multicolumn{1}{c}{\multirow{3}[6]{1cm}{\centering Model}} & \multicolumn{1}{c}{\multirow{3}[6]{1cm}{\centering Architec -ture}} & \multicolumn{15}{c}{\centering Probability of Correct Assignments} \\
\cmidrule{4-18}          &       &       & \multicolumn{3}{c}{0.1} & \multicolumn{3}{c}{0.25} & \multicolumn{3}{c}{0.5} & \multicolumn{3}{c}{0.75} & \multicolumn{3}{c}{0.9} \\
\cmidrule{4-18}          &       &       & \multicolumn{1}{c}{OOSP (\%)} & \multicolumn{1}{c}{Gap (\%)} & \multicolumn{1}{c}{Runtime (s)} & \multicolumn{1}{c}{OOSP (\%)} & \multicolumn{1}{c}{Gap (\%)} & \multicolumn{1}{c}{Runtime (s)} & \multicolumn{1}{c}{OOSP (\%)} & \multicolumn{1}{c}{Gap (\%)} & \multicolumn{1}{c}{Runtime (s)} & \multicolumn{1}{c}{OOSP (\%)} & \multicolumn{1}{c}{Gap (\%)} & \multicolumn{1}{c}{Runtime (s)} & \multicolumn{1}{c}{OOSP (\%)} & \multicolumn{1}{c}{Gap (\%)} & \multicolumn{1}{c}{Runtime (s)} \\
    \midrule
    \multirow{9}[10]{*}{100} & \multirow{2}[2]{*}{B-PT} & 1     & 70.28 (24.1) & 0 (0) & 31.27 (3.51) & 92.16 (8.23) & 0 (0) & 44.32 (5.82) & 95.79 (0.17) & 0 (0) & 43.6 (11.36) & 93.92 (4.22) & 0 (0) & 34.86 (9.49) & 71.56 (21.71) & 0 (0) & 32.34 (6.59) \\
          &       & 2     & 66.21 (18.77) & 1.59 (0.62) & 3600.06 (0.01) & 89.28 (10.94) & 2.77 (1.65) & 3600.07 (0.04) & 82.13 (9.6) & 4.18 (2.31) & 3600.12 (0.19) & 73.58 (27.11) & 3.68 (1.99) & 3600.07 (0.09) & 45.24 (12.25) & 1.82 (0.93) & 3600.1 (0.06) \\
\cmidrule{2-18}          & CF    & 100   & 95.79 (0.17) & -     & 0.88 (0.04) & 95.79 (0.17) & -     & 0.99 (0.08) & 95.79 (0.17) & -     & 1.09 (0.12) & 95.79 (0.17) & -     & 0.92 (0.04) & 95.79 (0.17) & -     & 0.85 (0.05) \\
\cmidrule{2-18}          & \multirow{2}[2]{*}{J-PT} & 1     & 95.79 (0.17) & 0 (0) & 0.49 (0.08) & 95.79 (0.17) & 0 (0) & 0.41 (0.03) & 95.79 (0.17) & 0 (0) & 0.39 (0.02) & 86.44 (11.51) & 0 (0) & 0.4 (0.01) & 90.34 (5.06) & 0 (0) & 0.42 (0.05) \\
          &       & 2     & 88.69 (3.17) & 0 (0) & 158.63 (66.87) & 89.75 (2.77) & 0 (0) & 110.64 (64.14) & 91.1 (4.14) & 0 (0) & 97 (55.53) & 84.83 (5.74) & 0 (0) & 126.96 (54.38) & 82.29 (8.02) & 0 (0) & 123.41 (42.14) \\
\cmidrule{2-18}          & \multirow{2}[2]{*}{K-PT} & 1     & 70.28 (24.1) & 0 (0) & 5.84 (1.02) & 82.85 (18.07) & 0 (0) & 6 (1.72) & 82.85 (4.96) & 0 (0) & 6.16 (1.61) & 86.8 (0.5) & 0 (0) & 7.11 (0.62) & 67.49 (17.58) & 0 (0) & 4.31 (1.18) \\
          &       & 2     & 79.68 (14.99) & 0.02 (0.05) & 2821.78 (738.69) & 87.85 (13.01) & 0.18 (0.11) & 3383.59 (484.22) & 81.17 (5.06) & 0.27 (0.08) & 3600.23 (0.11) & 65.29 (14.41) & 0.13 (0.12) & 3244.13 (682.43) & 45.26 (12.04) & 0 (0) & 1356.21 (771.37) \\
\cmidrule{2-18}          & \multirow{2}[2]{*}{PNN} & 20-3-2 & 95.79 (0.17) & 0 (0) & 0.53 (0.04) & 95.79 (0.17) & 0 (0) & 0.59 (0.08) & 95.79 (0.17) & 0 (0) & 0.54 (0.06) & 95.79 (0.17) & 0 (0) & 0.68 (0.31) & 95.79 (0.17) & 0 (0) & 0.72 (0.31) \\
          &       & 20-10-2 & 95.79 (0.17) & 0 (0) & 3.35 (1.47) & 95.79 (0.17) & 0 (0) & 3.57 (1.24) & 95.79 (0.17) & 0 (0) & 3.57 (1.89) & 95.79 (0.17) & 0 (0) & 5.74 (3.16) & 95.79 (0.17) & 0 (0) & 4.61 (2.31) \\
    \midrule
    \multirow{9}[10]{*}{500} & \multirow{2}[2]{*}{B-PT} & 1     & 78.92 (23.01) & 0 (0) & 660.69 (255.35) & 95.71 (0.15) & 0 (0) & 722.24 (236.38) & 95.71 (0.15) & 0 (0) & 869.83 (285.7) & 95.71 (0.15) & 0 (0) & 541.49 (132.78) & 95.71 (0.15) & 0 (0) & 695.11 (253.31) \\
          &       & 2     & 95.71 (0.15) & inf   & 3600.06 (0.02) & 95.71 (0.15) & inf   & 3600.08 (0.04) & 77.38 (40.93) & inf   & 3600.04 (0.04) & 40.74 (50.05) & inf   & 3600.02 (0.02) & 28.13 (38.46) & inf   & 3600.03 (0.03) \\
\cmidrule{2-18}          & CF    & 100   & 95.71 (0.15) & -     & 0.81 (0.03) & 95.71 (0.15) & -     & 0.8 (0.02) & 95.71 (0.15) & -     & 0.79 (0.01) & 95.71 (0.15) & -     & 0.79 (0.02) & 95.71 (0.15) & -     & 0.82 (0.02) \\
\cmidrule{2-18}          & \multirow{2}[2]{*}{J-PT} & 1     & 95.71 (0.15) & 0 (0) & 4.71 (0.76) & 95.71 (0.15) & 0 (0) & 4.25 (0.37) & 95.71 (0.15) & 0 (0) & 4.39 (0.57) & 95.71 (0.15) & 0 (0) & 3.97 (0.13) & 95.71 (0.15) & 0 (0) & 3.94 (0.52) \\
          &       & 2     & 94.68 (0.4) & 0 (0) & 1505.45 (387.84) & 94.8 (0.26) & 0 (0) & 2741.94 (865.17) & 94.77 (0.26) & 0 (0) & 2781.14 (464.43) & 94.78 (0.23) & 0 (0) & 2685.15 (626.67) & 94.81 (0.25) & 0 (0) & 2081.06 (533.69) \\
\cmidrule{2-18}          & \multirow{2}[2]{*}{K-PT} & 1     & 78.92 (23.01) & 0 (0) & 73.35 (12.37) & 95.71 (0.15) & 0 (0) & 81.45 (7.94) & 95.71 (0.15) & 0 (0) & 73.62 (20.97) & 95.71 (0.15) & 0 (0) & 66.69 (15.77) & 91.98 (5.19) & 0 (0) & 78.72 (25.57) \\
          &       & 2     & 80.25 (16.15) & 0.54 (0.07) & 3600.58 (0.17) & 94.53 (2.74) & 0.62 (0.19) & 3600.39 (0.09) & 94.01 (2.27) & 0.62 (0.15) & 3600.37 (0.12) & 91.61 (2.83) & 0.73 (0.13) & 3600.42 (0.16) & 84.54 (2.82) & 0.76 (0.13) & 3601.19 (1.15) \\
\cmidrule{2-18}          & \multirow{2}[2]{*}{PNN} & 20-3-2 & 95.71 (0.15) & 0.01 (0) & 3600.37 (0.5) & 95.71 (0.15) & 0 (0) & 311.09 (280.68) & 95.71 (0.15) & 0 (0) & 23.81 (10.85) & 95.71 (0.15) & 0 (0) & 653.81 (775.29) & 95.71 (0.15) & 0.01 (0.01) & 3600.32 (0.36) \\
          &       & 20-10-2 & 95.71 (0.15) & 0.01 (0) & 3600.49 (0.39) & 95.71 (0.15) & 0.01 (0) & 3600.58 (0.53) & 95.71 (0.15) & 0 (0) & 1529.09 (1891.52) & 95.71 (0.15) & 0 (0) & 3603.89 (7.25) & 95.71 (0.15) & 0.01 (0) & 3600.49 (0.5) \\
    \bottomrule
    \end{tabular}%
    }
    \caption{Complete Results of Simulated Experiments with Athey Model 3 (with Adapted Binarization).}
  \label{simulatedfull3e}%
\end{table*}%


\subsubsection{Personalized Postpartum Hypertension Treatments}

Here we display the average average optimality gap, average running time in seconds, and the average $\pi$ value for the polices derived by the models in the postpartum hypertension experiments. The averages are taken over the 10 cross-validation folds in the experiments, with the standard deviation given parenthetically. The 90\% and 95\% confidence intervals for the $\pi$ value are also included. The results of all three experiments are given.

\begin{table*}[ht]
  \centering
  \resizebox{15cm}{!}{%
    \begin{tabular}{cccllrll}
    \toprule
    \multicolumn{1}{l}{\centering Experiment} & \multicolumn{1}{p{5.665em}}{\centering Model} & \multicolumn{1}{p{5.25em}}{\centering Model Architecture} & \multicolumn{1}{p{5.085em}}{\centering Average Gap (\%) with st.dev} & \multicolumn{1}{p{7.585em}}{\centering Average Runtime (s) with st.dev} & \multicolumn{1}{p{3.165em}}{\centering $\pi$} & \multicolumn{1}{p{6.835em}}{\centering 95\% Confidence Interval} & \multicolumn{1}{p{6.835em}}{\centering 90\% Confidence Interval} \\
    \midrule
    \multirow{10}[12]{*}{1} & Policy in Data & -     & \multicolumn{1}{c}{-} & \multicolumn{1}{c}{-} & 148.32 & [146.59, 150.06] & [146.97, 149.67] \\
\cmidrule{2-8}          & \multirow{2}[2]{*}{B-PT} & 1     & 0.0 (0.0) & 21.74 (1.89) & 146.43 & [145.2, 147.66] & [145.47, 147.39] \\
          &       & 2     & 74.84 (13.44) & 3600.15 (0.05) & 147.15 & [145.85, 148.44] & [146.14, 148.16] \\
\cmidrule{2-8}          & CF    & 100   & \multicolumn{1}{c}{-} & 0.14 (0.01) & 146.37 & [145.14, 147.59] & [145.41, 147.32] \\
\cmidrule{2-8}          & \multirow{2}[2]{*}{J-PT} & 1     & 0.05 (0.03) & 3.11 (1.33) & 146.48 & [145.25, 147.71] & [145.52, 147.44] \\
          &       & 2     & 0.01 (0.02) & 285.77 (28.25) & 148.58 & [147.32, 149.84] & [147.6, 149.56] \\
\cmidrule{2-8}          & \multirow{2}[2]{*}{K-PT} & 1     & 0.0 (0.0) & 3.96 (0.47) & 146.43 & [145.2, 147.66] & [145.47, 147.39] \\
          &       & 2     & 2.32 (5.48) & 2620.15 (519.68) & 147.22 & [146.04, 148.4] & [146.3, 148.14] \\
\cmidrule{2-8}          & \multirow{2}[2]{*}{PNN} & 22-3-2 & 0.03 (0.0) & 3600.42 (0.49) & 146.43 & [145.2, 147.66] & [145.47, 147.39] \\
          &       & 22-10-2 & 0.0 (0.0) & 2333.51 (1156.53) & 145.76 & [143.71, 147.81] & [144.16, 147.36] \\
    \midrule
    \multirow{10}[12]{*}{2} & Policy in Data & -     & \multicolumn{1}{c}{-} & \multicolumn{1}{c}{-} & 154.92 & [144.58, 165.25] & [146.87, 162.97] \\
\cmidrule{2-8}          & \multirow{2}[2]{*}{B-PT} & 1     & 0.02 (0.04) & 33.21 (13.37) & 155.72 & [145.57, 165.86] & [147.81, 163.62] \\
          &       & 2     & 40.18 (21.48) & 3600.15 (0.07) & 155.19 & [145.12, 165.27] & [147.34, 163.05] \\
\cmidrule{2-8}          & CF    & 100   & \multicolumn{1}{c}{-} & 0.17 (0.0) & 150.05 & [148.47, 151.64] & [148.82, 151.29] \\
\cmidrule{2-8}          & \multirow{2}[2]{*}{J-PT} & 1     & 0.03 (0.04) & 8.94 (4.02) & 149.15 & [147.19, 151.1] & [147.62, 150.67] \\
          &       & 2     & 0.0 (0.01) & 434.55 (41.5) & 150.05 & [148.17, 151.94] & [148.58, 151.52] \\
\cmidrule{2-8}          & \multirow{2}[2]{*}{K-PT} & 1     & 0.0 (0.0) & 5.61 (1.51) & 157.32 & [147.22, 167.42] & [149.45, 165.19] \\
          &       & 2     & 0.0 (0.0) & 1507.98 (483.29) & 156.16 & [146.19, 166.13] & [148.39, 163.93] \\
\cmidrule{2-8}          & \multirow{2}[2]{*}{PNN} & 19-3-4 & 0.03 (0.0) & 3600.14 (0.13) & 148.1 & [146.32, 149.88] & [146.71, 149.49] \\
          &       & 19-10-4 & 0.01 (0.0) & 3600.97 (1.5) & 147.3 & [144.48, 150.11] & [145.1, 149.49] \\
    \midrule
    \multirow{10}[12]{*}{3} & Policy in Data & -     & \multicolumn{1}{c}{-} & \multicolumn{1}{c}{-} & 150.86 & [146.34, 155.38] & [147.34, 154.38] \\
\cmidrule{2-8}          & \multirow{2}[2]{*}{B-PT} & 1     & 0.01 (0.03) & 37.02 (15.84) & 148.52 & [145.98, 151.07] & [146.54, 150.51] \\
          &       & 2     & 0.53 (0.9) & 3156.02 (416.63) & 151.99 & [148.74, 155.25] & [149.46, 154.53] \\
\cmidrule{2-8}          & CF    & 100   & \multicolumn{1}{c}{-} & 0.17 (0.01) & 147.39 & [145.05, 149.73] & [145.56, 149.21] \\
\cmidrule{2-8}          & \multirow{2}[2]{*}{J-PT} & 1     & 0.02 (0.03) & 12.77 (5.65) & 146.7 & [144.27, 149.14] & [144.8, 148.6] \\
          &       & 2     & 0.01 (0.02) & 665.01 (165.11) & 148.5 & [146.28, 150.72] & [146.77, 150.23] \\
\cmidrule{2-8}          & \multirow{2}[2]{*}{K-PT} & 1     & 0.0 (0.0) & 6.39 (3.89) & 148.32 & [145.78, 150.86] & [146.34, 150.3] \\
          &       & 2     & 21.38 (33.41) & 1855.24 (1507.95) & 151.89 & [149.26, 154.51] & [149.84, 153.93] \\
\cmidrule{2-8}          & \multirow{2}[2]{*}{PNN} & 18-3-7 & 0.09 (0.01) & 3600.16 (0.16) & 146.05 & [144.07, 148.03] & [144.5, 147.59] \\
          &       & 18-10-7 & 0.09 (0.01) & 3600.34 (0.1) & 145.38 & [143.65, 147.12] & [144.03, 146.74] \\
    \bottomrule
    \end{tabular}%
    }
    \caption{Complete Results of Postpartum Hypertension Experiments.}
  \label{hypertensionfull}%
\end{table*}%


\subsubsection{Personalized Warfarin Dosing}

Table \ref{warfarin_complete} displays the average average optimality gap, average running time in seconds, and the average $\pi$ value for the polices derived by the models in the warfarin dosing experiments. The averages are taken over the 5 seeds of the experiments, with the standard deviation given parenthetically. The 90\% and 95\% confidence intervals for the $\pi$ value are also included.

\begin{table*}[ht]
  \centering
  \resizebox{15cm}{!}{%
    \begin{tabular}{ccllrll}
    \toprule
    \multicolumn{1}{p{5.665em}}{\centering Method} & \multicolumn{1}{p{5.25em}}{\centering Model Architecture} & \multicolumn{1}{p{6.085em}}{\centering Average Gap (\%) with st.dev} & \multicolumn{1}{p{7.585em}}{\centering Average Runtime (s) with st.dev} & \multicolumn{1}{p{3.165em}}{\centering $\pi$} & \multicolumn{1}{p{6.835em}}{\centering 95\% Confidence Interval} & \multicolumn{1}{p{6.835em}}{\centering 90\% Confidence Interval} \\
    \midrule
    Policy in Data & -     & \multicolumn{1}{c}{-} & \multicolumn{1}{c}{-} & 0.609 & [0.325, 0.892] & [0.388, 0.829] \\
    \midrule
    \multirow{2}[2]{*}{B-PT} & 1     & 0.0 (0.0) & 1185.55 (146.99) & 0.510  & [0.271, 0.749] & [0.324, 0.696] \\
          & 2     & 100.0 (0.0) & 3600.05 (0.01) & 0.545 & [0.312, 0.779] & [0.363, 0.727] \\
    \midrule
    CF    & 100   & \multicolumn{1}{c}{-} & 0.85 (0.02) & 0.265 & [0.248, 0.282] & [0.252, 0.279] \\
    \midrule
    \multirow{2}[2]{*}{J-PT} & 1     & 0.0 (0.0) & 89.52 (32.72) & 0.232 & [0.182, 0.283] & [0.193, 0.272] \\
          & 2     & 146.35 (46.04) & 3600.38 (0.4) & 0.275 & [0.265, 0.284] & [0.267, 0.282] \\
    \midrule
    \multirow{2}[2]{*}{K-PT} & 1     & 0.0 (0.0) & 211.66 (63.94) & 0.506 & [0.267, 0.744] & [0.32, 0.692] \\
          & 2     & 100.0 (0.0) & 3600.3 (0.08) & 0.449 & [0.205, 0.693] & [0.259, 0.639] \\
    \midrule
    \multirow{2}[2]{*}{PNN} & 76-3-2 & 73.6 (82.02) & 3170.79 (867.48) & 0.295 & [0.281, 0.309] & [0.284, 0.306] \\
          & 76-10-2 & 92.78 (47.38) & 3600.79 (0.9) & 0.290  & [0.271, 0.310] & [0.275, 0.305] \\
    \bottomrule
    \end{tabular}%
    }    
    \caption{Complete Results of Warfarin Dosing Experiments.}    
  \label{warfarin_complete}%
\end{table*}%


\subsubsection{Prediction Problem Results}

We provide the complete results of the prediction experiments conducted over the 2-digit MNIST problem. Table \ref{predictionfull} shows the results of training ANNs over various architectures for SGD with binary and ReLU activations, and the MIP with binary activations. The architecture is denoted by (Number of Hidden Layers, Number of Neurons in Each Layer). So, for example, (2,10) indicates 2 hidden layers with 10 neurons in each layer.

\begin{table*}[ht]
  \centering
  \resizebox{15cm}{!}{%
    \begin{tabular}{ccccccc}
    \toprule
    \multicolumn{1}{p{5em}}{\centering Training Approach} & \multicolumn{1}{p{5em}}{\centering Hidden Layer Activation} & \multicolumn{1}{p{5em}}{\centering Architecture} & \multicolumn{1}{p{5em}}{\centering Average Runtime (s)} & \multicolumn{1}{p{5em}}{\centering Average Optimality Gap (\%)} & \multicolumn{1}{p{5em}}{\centering Aveage Final Training Accuracy (\%)} & \multicolumn{1}{p{5em}}{\centering Final Testing Average Accuracy (\%)} \\
    \midrule
    \multirow{18}[4]{*}{SGD} & \multirow{9}[2]{*}{Binary} & (1, 3) & 165.31 & -     & 79.0  & 77.3 \\
          &       & (1, 5) & 195.42 & -     & 81.0  & 79.3 \\
          &       & (1, 10) & 181.83 & -     & 80.0  & 80.00 \\
          &       & (2, 3) & 197.65 & -     & 58.0  & 61.00 \\
          &       & (2, 5) & 207.99 & -     & 62.0  & 65.3 \\
          &       & (2, 10) & 207.49 & -     & 79.0  & 77.7 \\
          &       & (3, 3) & 223.53 & -     & 50.0  & 50.3 \\
          &       & (3, 5) & 239.01 & -     & 51.0  & 48.9 \\
          &       & (3, 10) & 242.44 & -     & 61.0  & 67.0 \\
\cmidrule{2-7}          & \multirow{9}[2]{*}{ReLU} & (1, 3) & 125.49 & -     & 50.0  & 50.3 \\
          &       & (1, 5) & 132.43 & -     & 50.0  & 50.3 \\
          &       & (1, 10) & 138.27 & -     & 50.0  & 50.3 \\
          &       & (2, 3) & 149.75 & -     & 50.0  & 50.3 \\
          &       & (2, 5) & 155.35 & -     & 50.0  & 50.3 \\
          &       & (2, 10) & 159.56 & -     & 54.0  & 53.4 \\
          &       & (3, 3) & 169.57 & -     & 50.0  & 50.3 \\
          &       & (3, 5) & 174.85 & -     & 50.0  & 50.3 \\
          &       & (3, 10) & 181.34 & -     & 50.0  & 50.3 \\
    \midrule
    \multirow{9}[2]{*}{MIP} & \multirow{9}[2]{*}{Binary} & (1, 3) & 68.88 & 0.00  & 98.4  & 84.8 \\
          &       & (1, 5) & 37.63 & 0.00  & 99.2  & 84.8 \\
          &       & (1, 10) & 34.08 & 0.00  & 98.8  & 82.2 \\
          &       & (2, 3) & 125.75 & 0.00  & 60.8  & 62.4 \\
          &       & (2, 5) & 431.15 & 0.00  & 100.0 & 80.3 \\
          &       & (2, 10) & 1056.70 & 0.00  & 99.6  & 87.0 \\
          &       & (3, 3) & 238.75 & 0.00  & 100.0 & 90.1 \\
          &       & (3, 5) & 135.02 & 0.00  & 99.2  & 85.9 \\
          &       & (3, 10) & 3601.15 & 100.00 & 75.6  & 66.2 \\
    \bottomrule
    \end{tabular}%
    }
    \caption{Complete Results of Prediction Experiments.}
  \label{predictionfull}%
\end{table*}%


\subsection{Hyperparameter Tuning Results for Prescriptive Neural Networks}\label{pnn_hyperparameter_tuning}

This sections provides the complete experimental results of the hyperparameter tuning conducted for the PNN models simulated, postpartum hypertension, and warfarin dosing experiments. These hyperparameters were tuned with a 10-fold cross validation set-up, where we considered $\lambda \in \{0, 0.01, 0.1, 1, 10\}$. The hyperparameter with the best out-of-sample probability of correct treatment assignment or $\pi$ was chosen for the final experiments. The tables are set up in this section are similarly to the previous section.

\subsubsection{Simulated Data}
Please see Tables \ref{l0athey1results}-\ref{l1athey3results}.
\begin{table*}[ht]
  \centering
  \resizebox{15cm}{!}{%
    \begin{tabular}{ccclllllllllllllll}
    \toprule
    \multicolumn{1}{c}{\multirow{3}[6]{1cm}{\centering Train Data Length}} & \multicolumn{1}{c}{\multirow{3}[6]{1cm}{\centering Width}} & \multicolumn{1}{c}{\multirow{3}[6]{1.5cm}{\centering Hyper parameter}} & \multicolumn{15}{c}{Probability of Correct Assignments} \\
\cmidrule{4-18} \multicolumn{1}{c}{} & \multicolumn{1}{c}{} &       & \multicolumn{3}{c}{0.1} & \multicolumn{3}{c}{0.25} & \multicolumn{3}{c}{0.5} & \multicolumn{3}{c}{0.75} & \multicolumn{3}{c}{0.9} \\
\cmidrule{4-18}    \multicolumn{1}{c}{} & \multicolumn{1}{c}{} &       & \multicolumn{1}{c}{OOSP (\%)} & \multicolumn{1}{c}{Gap (\%)} & \multicolumn{1}{c}{Runtime (s)} & \multicolumn{1}{c}{OOSP (\%)} & \multicolumn{1}{c}{Gap (\%)} & \multicolumn{1}{c}{Runtime (s)} & \multicolumn{1}{c}{OOSP (\%)} & \multicolumn{1}{c}{Gap (\%)} & \multicolumn{1}{c}{Runtime (s)} & \multicolumn{1}{c}{OOSP (\%)} & \multicolumn{1}{c}{Gap (\%)} & \multicolumn{1}{c}{Runtime (s)} & \multicolumn{1}{c}{OOSP (\%)} & \multicolumn{1}{c}{Gap (\%)} & \multicolumn{1}{c}{Runtime (s)} \\
    \midrule
    \multirow{10}[4]{*}{100} & \multirow{5}[2]{*}{3} & 0     & 87.6 (2.6) & 0.8 (1) & 2165.3 (1597.7) & 87.1 (1.2) & 2.5 (1.6) & 2881.4 (1607.2) & 87.1 (1.5) & 1.5 (1.4) & 2701.7 (1315.7) & 86.9 (3.2) & 2.8 (1.5) & 3600.4 (0.5) & 86.9 (1.8) & 1.6 (3.5) & 1925.3 (1390.7) \\
          &       & 0.01  & 88.9 (3.4) & 0 (0) & 22.6 (14.5) & 89.7 (3.4) & 0 (0) & 16 (10.9) & 91.3 (0.1) & 0 (0) & 11.6 (2.7) & 90.1 (2) & 0 (0) & 14.8 (5.4) & 91.3 (0.2) & 0 (0) & 14.8 (10) \\
          &       & 0.1   & 49.6 (0.4) & 0 (0) & 3.2 (0.9) & 49.5 (0.3) & 0 (0) & 2.7 (0.7) & 49.6 (0.4) & 0 (0) & 2.9 (0.8) & 49.5 (0.3) & 0 (0) & 2.9 (0.5) & 50.4 (0.4) & 0 (0) & 3.1 (0.5) \\
          &       & 1     & 49.6 (0.4) & 0 (0) & 1.3 (0.3) & 49.5 (0.3) & 0 (0) & 1.1 (0.2) & 49.6 (0.4) & 0 (0) & 1 (0.2) & 49.5 (0.3) & 0 (0) & 0.7 (0.3) & 50.4 (0.4) & 0 (0) & 1.2 (0.2) \\
          &       & 10    & 49.7 (0.5) & 0.1 (0) & 0.9 (0.2) & 49.5 (0.3) & 0.1 (0) & 0.9 (0.4) & 49.6 (0.4) & 0.1 (0) & 0.9 (0.1) & 49.5 (0.3) & 0 (0) & 0.8 (0.2) & 50.4 (0.4) & 0.1 (0) & 0.9 (0.1) \\
\cmidrule{2-18}          & \multirow{5}[2]{*}{10} & 0     & 86.2 (1.8) & 0 (0) & 61.1 (78.8) & 85.4 (1.5) & 0 (0) & 310 (394) & 84.6 (2.3) & 0 (0) & 137.9 (114.5) & 83 (3.9) & 0.1 (0.1) & 930.6 (1518.6) & 86.8 (2.9) & 0 (0) & 134 (114.4) \\
          &       & 0.01  & 88.9 (3.4) & 2.4 (2) & 3232.6 (821.7) & 89.8 (3.3) & 1.5 (1.5) & 2885.7 (1018.3) & 91.2 (0.2) & 1.2 (1.1) & 3080.7 (1161.3) & 90.1 (1.9) & 1.8 (1.7) & 3268.1 (587.5) & 91.3 (0.1) & 1.7 (2.3) & 2193.9 (1369) \\
          &       & 0.1   & 49.6 (0.4) & 0 (0) & 20.2 (12.4) & 49.5 (0.3) & 0 (0) & 15.7 (4.5) & 49.6 (0.4) & 0 (0) & 15.9 (6.6) & 49.5 (0.3) & 0 (0) & 12.2 (4.2) & 50.4 (0.4) & 0 (0) & 21.2 (3.5) \\
          &       & 1     & 49.6 (0.4) & 0.1 (0) & 4.4 (0.6) & 49.5 (0.3) & 0.1 (0) & 4.2 (1.1) & 49.5 (0.3) & 0.1 (0) & 4 (1.2) & 49.6 (0.4) & 0.1 (0) & 4.6 (0.2) & 50 (0.6) & 0.1 (0) & 4 (0.2) \\
          &       & 10    & 49.6 (0.4) & 0 (0) & 2.2 (0.1) & 49.5 (0.3) & 0 (0) & 2.2 (0.5) & 49.5 (0.3) & 0 (0) & 2 (0.5) & 49.6 (0.4) & 0 (0) & 2.1 (0.2) & 50 (0.6) & 0 (0) & 2.3 (0.4) \\
    \midrule
    \multirow{10}[4]{*}{500} & \multirow{5}[2]{*}{3} & 0     & 90 (0.9) & 8.9 (1.1) & 3600.3 (0.2) & 90.5 (0.7) & 6.8 (1.3) & 3600.2 (0.1) & 90.3 (0.4) & 5.7 (1.2) & 3600.4 (0.5) & 90.3 (0.4) & 7.3 (2.2) & 3600.1 (0.1) & 89.7 (0.8) & 12.7 (1.9) & 3600.4 (0.5) \\
          &       & 0.01  & 90.8 (0.6) & 5.5 (3.7) & 2996.3 (1350.3) & 91.2 (0.1) & 2 (2.8) & 2721.1 (1120.2) & 91.2 (0.1) & 2.2 (2.2) & 2764.4 (1144.5) & 91.3 (0.1) & 5.4 (4) & 3514.2 (192.2) & 91.2 (0.1) & 10.7 (3.5) & 3600.1 (0.1) \\
          &       & 0.1   & 50.3 (0.6) & 0 (0) & 67.7 (23.4) & 50.4 (0.5) & 0 (0) & 51.4 (21.9) & 50.1 (0.6) & 0 (0) & 62.1 (7.6) & 50.3 (0.6) & 0 (0) & 51.3 (13.4) & 50.1 (0.6) & 0 (0) & 66.9 (25.7) \\
          &       & 1     & 50.3 (0.6) & 0 (0) & 6.5 (1.8) & 50.4 (0.5) & 0 (0) & 6.5 (1.4) & 50.1 (0.6) & 0 (0) & 6.7 (1.3) & 50.3 (0.6) & 0 (0) & 6.7 (0.9) & 50.1 (0.6) & 0 (0) & 5.9 (1.5) \\
          &       & 10    & 50.3 (0.6) & 0.1 (0) & 6.8 (1.3) & 50.4 (0.5) & 0 (0) & 6.3 (1.2) & 50.1 (0.6) & 0 (0) & 5.1 (1) & 50.4 (0.5) & 0.1 (0) & 5.2 (0.7) & 50.1 (0.6) & 0 (0) & 5.2 (0.7) \\
\cmidrule{2-18}          & \multirow{5}[2]{*}{10} & 0     & 89.6 (0.9) & 7 (2.1) & 3600.6 (1.1) & 90 (0.4) & 6.3 (1.1) & 3600.4 (0.2) & 90.3 (0.6) & 5.3 (0.9) & 3601.5 (2.6) & 89.9 (0.6) & 6.8 (2.4) & 3600.3 (0.1) & 88.6 (1.3) & 11.1 (2.8) & 3600.3 (0.3) \\
          &       & 0.01  & 91 (0.3) & 6.1 (1) & 3600.1 (0.1) & 91.2 (0.1) & 5.9 (1.3) & 3601.7 (3.1) & 91.2 (0.1) & 5.8 (0.7) & 3600.5 (0.3) & 91 (0.7) & 6.3 (1.2) & 3600.9 (1.7) & 91.2 (0.1) & 7.9 (0.8) & 3600.6 (0.5) \\
          &       & 0.1   & 50.3 (0.6) & 0.4 (0.9) & 1424.9 (1445.3) & 50.4 (0.5) & 0.9 (1.2) & 2476.9 (1386.2) & 50.1 (0.6) & 0.8 (1.1) & 2456.3 (1446) & 50.3 (0.6) & 0 (0) & 1372 (1426.7) & 50.1 (0.6) & 0.5 (1) & 2114.2 (1599.4) \\
          &       & 1     & 50.3 (0.6) & 0 (0) & 30.3 (8.9) & 50.4 (0.5) & 0 (0) & 33.9 (10.4) & 50.1 (0.6) & 0.1 (0) & 35.4 (2.6) & 50.3 (0.6) & 0.1 (0) & 32.7 (9.1) & 50.1 (0.6) & 0 (0) & 32 (8.3) \\
          &       & 10    & 50.3 (0.6) & 0 (0) & 19.9 (4.3) & 50.4 (0.5) & 0 (0) & 20.9 (2.9) & 50.1 (0.6) & 0 (0) & 18.7 (3.3) & 50.1 (0.6) & 0 (0) & 18 (2.9) & 50.2 (0.6) & 0 (0) & 19 (2) \\
    \bottomrule
    \end{tabular}%
    }
    \caption{PNN Hyperparamter Tuning Results for 2-covariate data (Athey Model 1) with $\ell_0$ Regularization.}
  \label{l0athey1results}%
\end{table*}%

\begin{table*}[ht]
  \centering
  \resizebox{15cm}{!}{%
    \begin{tabular}{ccclllllllllllllll}
    \toprule
    \multicolumn{1}{c}{\multirow{3}[6]{1cm}{\centering Train Data Length}} & \multicolumn{1}{c}{\multirow{3}[6]{1cm}{\centering Width}} & \multicolumn{1}{c}{\multirow{3}[6]{1.5cm}{\centering Hyper parameter}} & \multicolumn{15}{c}{Probability of Correct Assignments} \\
\cmidrule{4-18}          &       &       & \multicolumn{3}{c}{0.1} & \multicolumn{3}{c}{0.25} & \multicolumn{3}{c}{0.5} & \multicolumn{3}{c}{0.75} & \multicolumn{3}{c}{0.9} \\
\cmidrule{4-18}          &       &       & \multicolumn{1}{c}{OOSP (\%)} & \multicolumn{1}{c}{Gap (\%)} & \multicolumn{1}{c}{Runtime (s)} & \multicolumn{1}{c}{OOSP (\%)} & \multicolumn{1}{c}{Gap (\%)} & \multicolumn{1}{c}{Runtime (s)} & \multicolumn{1}{c}{OOSP (\%)} & \multicolumn{1}{c}{Gap (\%)} & \multicolumn{1}{c}{Runtime (s)} & \multicolumn{1}{c}{OOSP (\%)} & \multicolumn{1}{c}{Gap (\%)} & \multicolumn{1}{c}{Runtime (s)} & \multicolumn{1}{c}{OOSP (\%)} & \multicolumn{1}{c}{Gap (\%)} & \multicolumn{1}{c}{Runtime (s)} \\
    \midrule
    \multirow{10}[4]{*}{100} & \multirow{5}[2]{*}{3} & 0     & 77.9 (4.3) & 0 (0) & 6.1 (5.9) & 76.2 (4.7) & 0 (0) & 14.2 (14.8) & 78.1 (2.6) & 0 (0) & 6.3 (3.2) & 78.3 (3.6) & 0 (0) & 4.1 (1.3) & 79.7 (2) & 0 (0) & 3.7 (3.2) \\
          &       & 0.01  & 85.1 (0.5) & 0.4 (0.9) & 888.4 (1517.1) & 85.1 (0.5) & 1.1 (2.4) & 1092.5 (1438.9) & 85.1 (0.5) & 0.1 (0.3) & 725.3 (1607.5) & 85.1 (0.5) & 0 (0) & 246.1 (313.3) & 85.1 (0.5) & 0.4 (0.8) & 853.8 (1546) \\
          &       & 0.1   & 85.1 (0.5) & 0 (0) & 2.6 (0.4) & 85.1 (0.5) & 0 (0) & 3.2 (0.9) & 85.1 (0.5) & 0.1 (0) & 2.4 (0.3) & 85.1 (0.5) & 0 (0) & 2.3 (0.4) & 85.1 (0.5) & 0 (0) & 2.6 (0.8) \\
          &       & 1     & 85.1 (0.5) & 0.1 (0) & 1.8 (0.7) & 85.1 (0.5) & 0 (0) & 2.3 (0.8) & 85.1 (0.5) & 0.1 (0) & 1.3 (1.1) & 85.1 (0.5) & 0.1 (0) & 1.5 (0.9) & 85.1 (0.5) & 0.1 (0) & 1.6 (0.7) \\
          &       & 10    & 85.1 (0.5) & 0 (0) & 1.3 (0.9) & 85.1 (0.5) & 0 (0) & 0.8 (0.2) & 85.1 (0.5) & 0 (0) & 0.9 (0.4) & 85.1 (0.5) & 0 (0) & 0.9 (0.5) & 85.1 (0.5) & 0 (0) & 0.8 (0.1) \\
\cmidrule{2-18}          & \multirow{5}[2]{*}{10} & 0     & 75.8 (3.4) & 0 (0) & 6.5 (3.3) & 76.6 (2.4) & 0 (0) & 6.5 (3) & 76.5 (2.3) & 0 (0) & 7 (3.1) & 76.3 (1.5) & 0 (0) & 8.9 (8.1) & 79.3 (1.5) & 0 (0) & 4.3 (1.9) \\
          &       & 0.01  & 85.1 (0.5) & 1.4 (1) & 3600.1 (0.1) & 85.1 (0.5) & 1.9 (1.3) & 3600.1 (0.1) & 85.1 (0.5) & 0.7 (0.9) & 3085.4 (1150.7) & 85.1 (0.5) & 0.9 (0.9) & 2483 (1580) & 85.1 (0.5) & 1.2 (0.9) & 3600.3 (0.6) \\
          &       & 0.1   & 85.1 (0.5) & 0 (0) & 9.5 (0.7) & 85.1 (0.5) & 0 (0) & 11.6 (3.9) & 85.1 (0.5) & 0 (0) & 8.7 (0.8) & 85.1 (0.5) & 0.1 (0) & 8.3 (3) & 85.1 (0.5) & 0.1 (0) & 12.1 (4.6) \\
          &       & 1     & 85.1 (0.5) & 0 (0) & 3.8 (1) & 85.1 (0.5) & 0 (0) & 5.5 (2) & 85.1 (0.5) & 0 (0) & 5.2 (1.6) & 85.1 (0.5) & 0 (0) & 4.6 (1.3) & 85.1 (0.5) & 0 (0) & 6.5 (2.3) \\
          &       & 10    & 85.1 (0.5) & 0 (0) & 4 (1.2) & 85.1 (0.5) & 0 (0) & 4.7 (2.3) & 85.1 (0.5) & 0 (0) & 4.7 (1.3) & 85.1 (0.5) & 0 (0) & 5.9 (1.4) & 85.1 (0.5) & 0 (0) & 5.3 (1.5) \\
    \midrule
    \multirow{10}[4]{*}{500} & \multirow{5}[2]{*}{3} & 0     & 81.9 (1.5) & 2.5 (1.4) & 3600.2 (0.4) & 82.3 (0.7) & 1.9 (0.4) & 3600.2 (0.1) & 82.4 (0.9) & 1.4 (0.3) & 3600.1 (0.1) & 82.3 (0.8) & 2.7 (0.3) & 3600.1 (0.1) & 81.7 (0.6) & 2.9 (1) & 3600.3 (0.4) \\
          &       & 0.01  & 85 (0.4) & 3.5 (1.5) & 3600.1 (0) & 85 (0.4) & 2 (0.9) & 3600.2 (0.2) & 85 (0.4) & 0.2 (0.2) & 2197.9 (1650.2) & 85 (0.4) & 1.9 (0.4) & 3600.1 (0) & 85 (0.4) & 4.3 (1) & 3600.2 (0.2) \\
          &       & 0.1   & 85 (0.4) & 0 (0) & 32.7 (28) & 85 (0.4) & 0 (0) & 39.6 (11) & 85 (0.4) & 0.1 (0) & 34.4 (16.9) & 85 (0.4) & 0 (0) & 37.3 (7) & 85 (0.4) & 0 (0) & 39.7 (6.6) \\
          &       & 1     & 85 (0.4) & 0.1 (0) & 31.1 (16.8) & 85 (0.4) & 0.1 (0) & 24.8 (13.9) & 85 (0.4) & 0.1 (0) & 17 (10.9) & 85 (0.4) & 0.1 (0) & 16.5 (6.1) & 85 (0.4) & 0.1 (0) & 27 (7.5) \\
          &       & 10    & 85 (0.4) & 0 (0) & 13.2 (2.2) & 85 (0.4) & 0 (0) & 9.2 (4.5) & 85 (0.4) & 0 (0) & 17.9 (17.5) & 85 (0.4) & 0 (0) & 10.5 (3.8) & 85 (0.4) & 0 (0) & 18.7 (19.1) \\
\cmidrule{2-18}          & \multirow{5}[2]{*}{10} & 0     & 79.4 (1.7) & 0.5 (0.5) & 3350.1 (568.7) & 78.6 (0.8) & 0.7 (0.1) & 3606.8 (12.4) & 79.1 (0.8) & 0.6 (0.2) & 3601 (1.2) & 79.3 (1.4) & 1.2 (0.4) & 3600.2 (0.1) & 79.3 (1.5) & 0.7 (0.5) & 3388.8 (474) \\
          &       & 0.01  & 85 (0.4) & 3.3 (0.6) & 3600.7 (0.4) & 85 (0.4) & 2.2 (0.6) & 3605.5 (11) & 85 (0.4) & 1.1 (0.1) & 3604.8 (9.7) & 85 (0.4) & 2 (0.2) & 3601.6 (1.6) & 85 (0.4) & 3.3 (0.6) & 3605.5 (11.5) \\
          &       & 0.1   & 85 (0.4) & 0.1 (0) & 185.8 (205.3) & 85 (0.4) & 0.1 (0) & 82.3 (11.9) & 85 (0.4) & 0.1 (0) & 74.6 (9.7) & 85 (0.4) & 0.1 (0) & 91.6 (9.3) & 85 (0.4) & 0.1 (0) & 87.1 (12.6) \\
          &       & 1     & 85 (0.4) & 0.1 (0) & 48.8 (16.7) & 85 (0.4) & 0 (0) & 48.5 (8.8) & 85 (0.4) & 0 (0) & 41 (13.6) & 85 (0.4) & 0 (0) & 50 (13.4) & 85 (0.4) & 0.1 (0) & 49.3 (14.4) \\
          &       & 10    & 85 (0.4) & 0 (0) & 44.6 (9.7) & 85 (0.4) & 0 (0) & 40.5 (6.6) & 85 (0.4) & 0 (0) & 38.9 (6.2) & 85 (0.4) & 0 (0) & 42.8 (6.8) & 85 (0.4) & 0 (0) & 43.5 (10.6) \\
    \bottomrule
    \end{tabular}%
    }
    \caption{PNN Hyperparamter Tuning Results for 10-covariate data (Athey Model 2) with $\ell_0$ Regularization.}
  \label{l0athey2results}%
\end{table*}%

\begin{table*}[ht]
  \centering
  \resizebox{15cm}{!}{%
    \begin{tabular}{ccclllllllllllllll}
    \toprule
    \multicolumn{1}{c}{\multirow{3}[6]{1cm}{\centering Train Data Length}} & \multicolumn{1}{c}{\multirow{3}[6]{1cm}{\centering Width}} & \multicolumn{1}{c}{\multirow{3}[6]{1.5cm}{\centering Hyper parameter}} & \multicolumn{15}{c}{Probability of Correct Assignments} \\
\cmidrule{4-18}          &       &       & \multicolumn{3}{c}{0.1} & \multicolumn{3}{c}{0.25} & \multicolumn{3}{c}{0.5} & \multicolumn{3}{c}{0.75} & \multicolumn{3}{c}{0.9} \\
\cmidrule{4-18}          &       &       & \multicolumn{1}{c}{OOSP (\%)} & \multicolumn{1}{c}{Gap (\%)} & \multicolumn{1}{c}{Runtime (s)} & \multicolumn{1}{c}{OOSP (\%)} & \multicolumn{1}{c}{Gap (\%)} & \multicolumn{1}{c}{Runtime (s)} & \multicolumn{1}{c}{OOSP (\%)} & \multicolumn{1}{c}{Gap (\%)} & \multicolumn{1}{c}{Runtime (s)} & \multicolumn{1}{c}{OOSP (\%)} & \multicolumn{1}{c}{Gap (\%)} & \multicolumn{1}{c}{Runtime (s)} & \multicolumn{1}{c}{OOSP (\%)} & \multicolumn{1}{c}{Gap (\%)} & \multicolumn{1}{c}{Runtime (s)} \\
    \midrule
    \multirow{10}[4]{*}{100} & \multirow{5}[2]{*}{3} & 0     & 56.8 (2.7) & 0 (0) & 0.1 (0) & 58.4 (5.3) & 0 (0) & 0.1 (0) & 60.2 (4.3) & 0 (0) & 0.1 (0) & 56 (6) & 0 (0) & 0.1 (0) & 61.5 (10.6) & 0 (0) & 0.5 (1) \\
          &       & 0.01  & 91.1 (6.3) & 1.3 (1.1) & 2277.4 (1821.7) & 95.8 (0.2) & 0 (0) & 129.5 (162.8) & 95.8 (0.2) & 0.2 (0.3) & 726.6 (1606.6) & 85.3 (10.9) & 1.7 (1.6) & 2200.1 (1917.9) & 85.9 (11.1) & 1.7 (1.6) & 2356.5 (1725.1) \\
          &       & 0.1   & 95.8 (0.2) & 0 (0) & 5 (4.2) & 95.8 (0.2) & 0 (0) & 1.9 (0.1) & 95.8 (0.2) & 0 (0) & 1.9 (0.6) & 95.8 (0.2) & 0.8 (1.6) & 726.3 (1606.6) & 95.8 (0.2) & 0.2 (0.3) & 723.2 (1608.3) \\
          &       & 1     & 95.8 (0.2) & 0.1 (0) & 1.4 (0.9) & 95.8 (0.2) & 0 (0) & 0.6 (0.1) & 95.8 (0.2) & 0 (0) & 0.5 (0.1) & 95.8 (0.2) & 0 (0) & 3.1 (2.5) & 95.8 (0.2) & 0.1 (0) & 2.6 (2) \\
          &       & 10    & 95.8 (0.2) & 0 (0) & 0.5 (0) & 95.8 (0.2) & 0 (0) & 0.6 (0.1) & 95.8 (0.2) & 0 (0) & 0.5 (0.1) & 95.8 (0.2) & 0 (0) & 0.7 (0.3) & 95.8 (0.2) & 0 (0) & 0.7 (0.3) \\
\cmidrule{2-18}          & \multirow{5}[2]{*}{10} & 0     & 54.1 (4.2) & 0 (0) & 0.3 (0) & 55.4 (4) & 0 (0) & 0.3 (0) & 55 (3.9) & 0 (0) & 0.3 (0) & 48.3 (6.7) & 0 (0) & 0.3 (0) & 56.8 (13) & 0 (0) & 0.7 (1) \\
          &       & 0.01  & 92.1 (4.9) & 1.4 (0.9) & 3600.2 (0.1) & 95.8 (0.2) & 0.4 (0.3) & 3002.4 (1336.6) & 95.8 (0.2) & 0.4 (0.4) & 2911.3 (1540.3) & 84.7 (11.3) & 1.6 (0.9) & 3600.2 (0.1) & 86.1 (11.1) & 1.7 (0.9) & 3600.1 (0) \\
          &       & 0.1   & 95.8 (0.2) & 0.1 (0) & 11.8 (6.1) & 95.8 (0.2) & 0 (0) & 5.9 (1.8) & 95.8 (0.2) & 0.1 (0) & 6.1 (2.4) & 95.8 (0.2) & 0.4 (0.8) & 1084.6 (1513.9) & 95.8 (0.2) & 0.2 (0.4) & 841.3 (1550.7) \\
          &       & 1     & 95.8 (0.2) & 0 (0) & 3.5 (2.1) & 95.8 (0.2) & 0 (0) & 4.2 (2.3) & 95.8 (0.2) & 0 (0) & 5.5 (4.3) & 95.8 (0.2) & 0.1 (0) & 6.5 (3.9) & 95.8 (0.2) & 0 (0) & 6.6 (2.6) \\
          &       & 10    & 95.8 (0.2) & 0 (0) & 3.3 (1.5) & 95.8 (0.2) & 0 (0) & 3.6 (1.2) & 95.8 (0.2) & 0 (0) & 3.6 (1.9) & 95.8 (0.2) & 0 (0) & 5.7 (3.2) & 95.8 (0.2) & 0 (0) & 4.6 (2.3) \\
    \midrule
    \multirow{10}[4]{*}{500} & \multirow{5}[2]{*}{3} & 0     & 86.4 (3) & 0.2 (0.1) & 1956.4 (1577.5) & 87.3 (1.6) & 0.1 (0.1) & 1801.7 (1687.7) & 89.3 (2.2) & 0.1 (0) & 55.7 (40.1) & 86.8 (1.6) & 0.1 (0) & 1285.5 (1552.7) & 87.8 (2.3) & 0.2 (0.2) & 1934.5 (1556.5) \\
          &       & 0.01  & 95.7 (0.1) & 1.4 (0.4) & 3600.4 (0.5) & 95.7 (0.1) & 0.1 (0) & 311.1 (280.7) & 95.7 (0.1) & 0.1 (0) & 23.8 (10.9) & 95.7 (0.1) & 0 (0) & 653.8 (775.3) & 95.7 (0.1) & 1.2 (0.5) & 3600.3 (0.4) \\
          &       & 0.1   & 95.7 (0.1) & 0.1 (0) & 40.4 (4.8) & 95.7 (0.1) & 0.1 (0) & 34.6 (6.2) & 95.7 (0.1) & 0.1 (0) & 17.1 (6.7) & 95.7 (0.1) & 0.1 (0) & 27.3 (13.1) & 95.7 (0.1) & 0 (0) & 52.1 (11.4) \\
          &       & 1     & 95.7 (0.1) & 0.1 (0) & 20.3 (20.4) & 95.7 (0.1) & 0 (0) & 24.9 (14.7) & 95.7 (0.1) & 0 (0) & 13.6 (16.1) & 95.7 (0.1) & 0 (0) & 8.3 (3.8) & 95.7 (0.1) & 0.1 (0) & 11.6 (13.4) \\
          &       & 10    & 95.7 (0.1) & 0 (0) & 7.3 (4.1) & 95.7 (0.1) & 0 (0) & 9.5 (3.8) & 95.7 (0.1) & 0 (0) & 14.2 (8.3) & 95.7 (0.1) & 0 (0) & 17 (15.8) & 95.7 (0.1) & 0 (0) & 7 (3.9) \\
\cmidrule{2-18}          & \multirow{5}[2]{*}{10} & 0     & 79.7 (2) & 0 (0) & 90.8 (104.4) & 79 (2.7) & 0 (0) & 86.2 (73.4) & 84.1 (1) & 0 (0) & 64.2 (68) & 80.6 (4.1) & 0 (0) & 74.9 (58.1) & 81 (3.9) & 0 (0) & 475.5 (820.1) \\
          &       & 0.01  & 95.7 (0.1) & 1.4 (0.2) & 3600.5 (0.4) & 95.7 (0.1) & 0.5 (0.1) & 3600.6 (0.5) & 95.7 (0.1) & 0.1 (0) & 1529.1 (1891.5) & 95.7 (0.1) & 0.5 (0.1) & 3603.9 (7.2) & 95.7 (0.1) & 1.3 (0.3) & 3600.5 (0.5) \\
          &       & 0.1   & 95.7 (0.1) & 0 (0) & 109.7 (24.9) & 95.7 (0.1) & 0.1 (0) & 91.4 (46.9) & 95.7 (0.1) & 0 (0) & 49.1 (24.7) & 95.7 (0.1) & 0.1 (0) & 63.4 (23.6) & 95.7 (0.1) & 0 (0) & 118.9 (34.4) \\
          &       & 1     & 95.7 (0.1) & 0 (0) & 62.9 (35.6) & 95.7 (0.1) & 0 (0) & 57.2 (20.8) & 95.7 (0.1) & 0 (0) & 72 (46.2) & 95.7 (0.1) & 0 (0) & 68.8 (30.6) & 95.7 (0.1) & 0 (0) & 72.8 (30.6) \\
          &       & 10    & 95.7 (0.1) & 0 (0) & 47.8 (15.8) & 95.7 (0.1) & 0 (0) & 48.1 (23.1) & 95.7 (0.1) & 0 (0) & 44.7 (24.1) & 95.7 (0.1) & 0 (0) & 40.6 (15.7) & 95.7 (0.1) & 0 (0) & 65 (32) \\
    \bottomrule
    \end{tabular}%
    }
    \caption{PNN Hyperparamter Tuning Results for 20-covariate data (Athey Model 3) with $\ell_0$ Regularization.}
  \label{l0athey3results}%
\end{table*}%

\begin{table*}[ht]
  \centering
  \resizebox{15cm}{!}{%
    \begin{tabular}{ccclllllllllllllll}
    \toprule
    \multicolumn{1}{c}{\multirow{3}[6]{1cm}{\centering Train Data Length}} & \multicolumn{1}{c}{\multirow{3}[6]{1cm}{\centering Width}} & \multicolumn{1}{c}{\multirow{3}[6]{1.5cm}{\centering Hyper parameter}} & \multicolumn{15}{c}{Probability of Correct Assignments} \\
\cmidrule{4-18}          &       &       & \multicolumn{3}{c}{0.1} & \multicolumn{3}{c}{0.25} & \multicolumn{3}{c}{0.5} & \multicolumn{3}{c}{0.75} & \multicolumn{3}{c}{0.9} \\
\cmidrule{4-18}          &       &       & \multicolumn{1}{c}{OOSP (\%)} & \multicolumn{1}{c}{Gap (\%)} & \multicolumn{1}{c}{Runtime (s)} & \multicolumn{1}{c}{OOSP (\%)} & \multicolumn{1}{c}{Gap (\%)} & \multicolumn{1}{c}{Runtime (s)} & \multicolumn{1}{c}{OOSP (\%)} & \multicolumn{1}{c}{Gap (\%)} & \multicolumn{1}{c}{Runtime (s)} & \multicolumn{1}{c}{OOSP (\%)} & \multicolumn{1}{c}{Gap (\%)} & \multicolumn{1}{c}{Runtime (s)} & \multicolumn{1}{c}{OOSP (\%)} & \multicolumn{1}{c}{Gap (\%)} & \multicolumn{1}{c}{Runtime (s)} \\
    \midrule
    \multirow{10}[4]{*}{100} & \multirow{5}[2]{*}{3} & 0     & 87.9 (2.5) & 0.7 (0.9) & 1972.2 (1533.3) & 86.8 (1.6) & 2.4 (1.7) & 2881.4 (1609.1) & 86.2 (2.1) & 1.5 (1.4) & 2478.2 (1563.8) & 86.4 (4.3) & 2.4 (1.9) & 3446 (345.1) & 87.2 (1.8) & 1.6 (3.5) & 1855.4 (1167.9) \\
          &       & 0.01  & 87.8 (2.1) & 5 (1.1) & 3600.3 (0.2) & 87.4 (2.1) & 6.2 (3) & 3600.3 (0.1) & 87.5 (2.9) & 5.2 (1.7) & 3600.4 (0.3) & 87.4 (2.2) & 6 (2.3) & 3600.6 (0.2) & 87.6 (2.2) & 5.5 (4.1) & 3600.3 (0.2) \\
          &       & 0.1   & 88.1 (1.7) & 9.8 (3.9) & 3600.2 (0.1) & 88.8 (1.9) & 8.2 (5.5) & 3215.8 (859.7) & 88.5 (2) & 4.7 (6.6) & 2306.2 (1467.1) & 88.8 (2.1) & 8.6 (5.4) & 3161.6 (981.3) & 88.4 (2) & 8 (8.9) & 3046.5 (780.9) \\
          &       & 1     & 87.1 (4.1) & 9.1 (14.7) & 3030.6 (749.7) & 87.1 (4.1) & 50 (30.1) & 2901.2 (1563.1) & 87.1 (4.1) & 12.5 (27.8) & 2976.9 (560.9) & 86.1 (4.2) & 27 (26) & 3225.2 (837.2) & 87.1 (4.1) & 23.1 (36.3) & 2976.1 (758.7) \\
          &       & 10    & 49.6 (0.4) & 0 (0) & 11.3 (5.5) & 49.5 (0.3) & 0 (0) & 9.5 (4.9) & 49.6 (0.4) & 0 (0) & 9.4 (5.8) & 49.5 (0.3) & 0 (0) & 5.2 (0.7) & 50.4 (0.4) & 0 (0) & 11.4 (6.5) \\
\cmidrule{2-18}          & \multirow{5}[2]{*}{10} & 0     & 85.9 (0.9) & 0 (0) & 104.3 (101.1) & 86 (2.7) & 0 (0.1) & 910.7 (1534.6) & 85.6 (1.9) & 0 (0) & 135.3 (121) & 83.9 (2.7) & 0 (0) & 319.2 (348.5) & 87.1 (2.4) & 0 (0) & 567.3 (1106.7) \\
          &       & 0.01  & 87.6 (2.5) & 5.3 (0.8) & 3600.1 (0) & 88.4 (2) & 6.6 (2.7) & 3600.1 (0.1) & 88.3 (2.5) & 5.4 (1.7) & 3600.1 (0.1) & 87.4 (2.2) & 6.6 (2.3) & 3600.1 (0) & 86.7 (3.9) & 4.9 (3) & 3600.1 (0) \\
          &       & 0.1   & 88.1 (1.7) & 15 (3.2) & 3600.1 (0) & 88.8 (2) & 14.1 (5.6) & 3600.2 (0.2) & 88.5 (2) & 11.6 (4.3) & 3600.1 (0.1) & 88.8 (2.1) & 14.8 (5.1) & 3600.1 (0) & 88.8 (1.1) & 14.7 (7.3) & 3600.1 (0) \\
          &       & 1     & 87.1 (4.1) & 118.4 (23.6) & 3600.1 (0.1) & 87.1 (4.1) & 136.2 (34) & 3600.1 (0) & 84.8 (6.1) & 116.2 (37.5) & 3600.1 (0.1) & 87.3 (4.3) & 123.1 (35.7) & 3600.1 (0) & 87.1 (4.1) & 113.6 (34.4) & 3600.1 (0) \\
          &       & 10    & 49.6 (0.4) & 0 (0) & 18 (18.4) & 49.5 (0.3) & 0 (0) & 10.7 (3.9) & 49.6 (0.4) & 0 (0) & 9.7 (3.5) & 49.5 (0.3) & 0 (0) & 11.4 (8.9) & 50.4 (0.4) & 0 (0) & 10.7 (2.5) \\
    \midrule
    \multirow{10}[4]{*}{500} & \multirow{5}[2]{*}{3} & 0     & 90.4 (0.4) & 9.5 (1.5) & 3600.4 (0.3) & 90.9 (0.4) & 7.1 (1.1) & 3600.7 (0.5) & 90.5 (0.6) & 5.6 (1) & 3600.3 (0.3) & 90.2 (0.8) & 7.1 (2.1) & 3600.1 (0) & 90.3 (0.9) & 12.9 (2.3) & 3600.2 (0) \\
          &       & 0.01  & 90.4 (0.3) & 13.2 (1.6) & 3600.2 (0.3) & 90.7 (0.5) & 10.1 (1.4) & 3600.2 (0.1) & 90.3 (0.9) & 8.6 (1.2) & 3600.6 (0.5) & 90.8 (0.4) & 11.2 (2.6) & 3600.2 (0.1) & 90.6 (0.6) & 18.2 (3.3) & 3600.2 (0.1) \\
          &       & 0.1   & 90 (1.3) & 30.1 (2.8) & 3600.3 (0.4) & 90 (1.3) & 26.4 (2.3) & 3600.2 (0.1) & 90 (1.3) & 24.6 (2.2) & 3600.1 (0.1) & 90.2 (0.8) & 28.1 (3.7) & 3600.6 (1.1) & 90.4 (0.8) & 37.4 (6.7) & 3600.1 (0.1) \\
          &       & 1     & 50.3 (0.6) & 1865.3 (1013.4) & 3600.2 (0.1) & 50.4 (0.5) & 2978.8 (2815.2) & 3600.2 (0.1) & 50.1 (0.6) & 2162.1 (1450.9) & 3600.2 (0.1) & 50.3 (0.6) & 1367.5 (606.6) & 3600.2 (0.1) & 50.1 (0.6) & 1869.8 (934.7) & 3600.2 (0.1) \\
          &       & 10    & 50.3 (0.6) & 0 (0) & 400.2 (421.9) & 50.4 (0.5) & 0 (0) & 192.7 (150.4) & 50.1 (0.6) & 0 (0) & 196.5 (160.3) & 50.3 (0.6) & 0 (0) & 165.6 (109.3) & 50.1 (0.6) & 0 (0) & 240 (106.5) \\
\cmidrule{2-18}          & \multirow{5}[2]{*}{10} & 0     & 89.4 (0.8) & 6.9 (1.6) & 3600.5 (0.3) & 89.9 (0.6) & 6.4 (0.9) & 3600.4 (0.2) & 90.1 (0.5) & 5.1 (0.9) & 3600.4 (0.3) & 90.1 (0.7) & 6.6 (1.9) & 3600.4 (0.2) & 89.5 (0.6) & 9.8 (2.5) & 3600.9 (0.8) \\
          &       & 0.01  & 90.4 (0.3) & 13.3 (1.6) & 3600.4 (0.1) & 90.7 (0.4) & 10.2 (1.3) & 3600.4 (0.1) & 90.8 (0.5) & 8.9 (1.4) & 3600.3 (0.1) & 90.8 (0.4) & 11.3 (2.6) & 3600.6 (0.8) & 90 (1.5) & 20.3 (5.6) & 3601.5 (2.3) \\
          &       & 0.1   & 90.1 (1.2) & 32.2 (3.4) & 3601.1 (1.5) & 90 (1.3) & 27.5 (2.2) & 3600.5 (0.3) & 89.9 (1.3) & 26.8 (1.9) & 3603.1 (3.6) & 90.1 (0.9) & 29.5 (3.9) & 3600.4 (0.1) & 89.4 (2.2) & 41.3 (8.8) & 3601 (1) \\
          &       & 1     & 50.3 (0.6) & 2119.1 (1162) & 3615.4 (20.6) & 50.4 (0.5) & 3306.9 (3063.4) & 3606 (12.3) & 50.1 (0.6) & 2411.4 (1580.5) & 3608.1 (17.4) & 50.3 (0.6) & 1536.7 (659.4) & 3600.4 (0.2) & 50.1 (0.6) & 2086 (990.5) & 3600.6 (0.4) \\
          &       & 10    & 50.3 (0.6) & 0 (0) & 178.4 (76.4) & 50.4 (0.5) & 0 (0) & 146.4 (28.3) & 50.1 (0.6) & 0 (0) & 158.3 (15.6) & 50.3 (0.6) & 0 (0) & 222.1 (131.4) & 50.1 (0.6) & 0 (0) & 181.9 (81.4) \\
    \bottomrule
    \end{tabular}%
    }
    \caption{PNN Hyperparamter Tuning Results for 2-covariate data (Athey Model 1) with $\ell_1$ Regularization.}
  \label{l1athey1results}%
\end{table*}%

\begin{table*}[ht]
  \centering
  \resizebox{15cm}{!}{%
    \begin{tabular}{ccclllllllllllllll}
    \toprule
    \multicolumn{1}{c}{\multirow{3}[6]{1cm}{\centering Train Data Length}} & \multicolumn{1}{c}{\multirow{3}[6]{1cm}{\centering Width}} & \multicolumn{1}{c}{\multirow{3}[6]{1.5cm}{\centering Hyper parameter}} & \multicolumn{15}{c}{Probability of Correct Assignments} \\
\cmidrule{4-18}          &       &       & \multicolumn{3}{c}{0.1} & \multicolumn{3}{c}{0.25} & \multicolumn{3}{c}{0.5} & \multicolumn{3}{c}{0.75} & \multicolumn{3}{c}{0.9} \\
\cmidrule{4-18}          &       &       & \multicolumn{1}{c}{OOSP (\%)} & \multicolumn{1}{c}{Gap (\%)} & \multicolumn{1}{c}{Runtime (s)} & \multicolumn{1}{c}{OOSP (\%)} & \multicolumn{1}{c}{Gap (\%)} & \multicolumn{1}{c}{Runtime (s)} & \multicolumn{1}{c}{OOSP (\%)} & \multicolumn{1}{c}{Gap (\%)} & \multicolumn{1}{c}{Runtime (s)} & \multicolumn{1}{c}{OOSP (\%)} & \multicolumn{1}{c}{Gap (\%)} & \multicolumn{1}{c}{Runtime (s)} & \multicolumn{1}{c}{OOSP (\%)} & \multicolumn{1}{c}{Gap (\%)} & \multicolumn{1}{c}{Runtime (s)} \\
    \midrule
    \multirow{10}[4]{*}{100} & \multirow{5}[2]{*}{3} & 0     & 79.2 (4.4) & 0 (0) & 25.7 (47.4) & 76.9 (2.7) & 0 (0) & 12 (7.7) & 76.5 (3.7) & 0 (0) & 6.7 (2.2) & 76.5 (3.7) & 0 (0) & 5.9 (2.2) & 78.3 (2) & 0 (0) & 4.4 (4.1) \\
          &       & 0.01  & 80.2 (2.9) & 0.3 (0.1) & 3600.3 (0.2) & 77.7 (3.9) & 0.4 (0.2) & 3600.5 (0.5) & 79.8 (1.7) & 0.3 (0.1) & 3600.4 (0.3) & 80.5 (1.5) & 0.3 (0.1) & 3600.4 (0.3) & 80 (1.2) & 0.2 (0.1) & 3600.3 (0) \\
          &       & 0.1   & 80.5 (3.3) & 2.1 (0.6) & 3600.3 (0.1) & 80.8 (3.6) & 2.5 (0.9) & 3600.2 (0.1) & 82.9 (1.9) & 2.1 (0.7) & 3600.2 (0.1) & 82.3 (1.2) & 2 (0.8) & 3600.3 (0.2) & 82.2 (1.2) & 1.8 (0.7) & 3600.3 (0.1) \\
          &       & 1     & 85.1 (0.5) & 0.6 (1.4) & 964.3 (1484.4) & 85.1 (0.5) & 1.4 (3.2) & 1589.4 (1379.4) & 85.1 (0.5) & 0.5 (1.1) & 744.2 (1596.6) & 85.1 (0.5) & 0 (0) & 790.1 (1573.4) & 85.1 (0.5) & 1 (1.6) & 1517.8 (1902.1) \\
          &       & 10    & 85.1 (0.5) & 0 (0) & 1.4 (1) & 85.1 (0.5) & 0 (0) & 2 (1.2) & 85.1 (0.5) & 0 (0) & 0.9 (0.8) & 85.1 (0.5) & 0 (0) & 1.9 (1.8) & 85.1 (0.5) & 0 (0) & 1.8 (0.8) \\
\cmidrule{2-18}          & \multirow{5}[2]{*}{10} & 0     & 74.7 (2.7) & 0 (0) & 8.2 (4.9) & 74.2 (2.6) & 0 (0) & 11 (9.2) & 77.8 (2.5) & 0 (0) & 9.2 (5.8) & 76.8 (1.8) & 0 (0) & 9.1 (9.1) & 78.9 (2.2) & 0 (0) & 4.5 (2.5) \\
          &       & 0.01  & 80.5 (2.2) & 0.3 (0.1) & 3600.2 (0.2) & 78.4 (4) & 0.4 (0.1) & 3600.1 (0.1) & 78.2 (3) & 0.4 (0.1) & 3600.1 (0.1) & 79.4 (2.6) & 0.3 (0.1) & 3600.2 (0.1) & 79.7 (1.5) & 0.3 (0.1) & 3600.1 (0) \\
          &       & 0.1   & 80.5 (3.5) & 2.3 (0.6) & 3600.1 (0) & 81.4 (2.4) & 2.8 (1) & 3600.1 (0.1) & 82.6 (2.2) & 2.3 (0.7) & 3600.1 (0) & 82.3 (1.2) & 2.2 (0.8) & 3600.2 (0.1) & 81.9 (0.9) & 2.1 (0.7) & 3600.1 (0.1) \\
          &       & 1     & 85.1 (0.5) & 2.8 (2.5) & 3001.8 (1337.7) & 85.1 (0.5) & 4.1 (3.6) & 3013 (1312.8) & 85.1 (0.5) & 0.6 (1.3) & 2705.7 (1276.9) & 85.1 (0.5) & 1.5 (1.8) & 2214.7 (1898.2) & 85.1 (0.5) & 2.6 (3.5) & 3084.2 (1153.5) \\
          &       & 10    & 85.1 (0.5) & 0 (0) & 31.7 (29.8) & 85.1 (0.5) & 0 (0) & 19.8 (16.3) & 85.1 (0.5) & 0 (0) & 8.6 (3.2) & 85.1 (0.5) & 0 (0) & 13 (8.8) & 85.1 (0.5) & 0 (0) & 14.1 (6.4) \\
    \midrule
    \multirow{10}[4]{*}{500} & \multirow{5}[2]{*}{3} & 0     & 81.3 (1.5) & 2.4 (1.5) & 3600.1 (0) & 82.8 (0.6) & 2.3 (0.4) & 3600.2 (0.1) & 83 (0.7) & 1.5 (0.3) & 3600.3 (0.2) & 82.4 (0.3) & 2.6 (0.4) & 3600.7 (0.5) & 80.5 (2.9) & 2.7 (0.6) & 3600.2 (0.1) \\
          &       & 0.01  & 82.9 (0.6) & 3.7 (1.6) & 3600.2 (0.1) & 83.6 (0.3) & 3.3 (0.3) & 3600.1 (0) & 84.1 (0.5) & 2.2 (0.3) & 3600.2 (0.1) & 83.9 (0.5) & 3.6 (0.4) & 3600.2 (0.1) & 83.1 (0.7) & 4.4 (0.5) & 3600.1 (0.1) \\
          &       & 0.1   & 84.8 (0.6) & 7.2 (2) & 3600.2 (0.1) & 85 (0.4) & 5.2 (1) & 3600.4 (0.2) & 85 (0.4) & 2.9 (0.2) & 3600.1 (0.1) & 85 (0.4) & 5.3 (0.8) & 3600.7 (0.9) & 84.5 (0.8) & 7.9 (1) & 3600.3 (0.3) \\
          &       & 1     & 85 (0.4) & 5.8 (1.7) & 3600.1 (0) & 85 (0.4) & 2.7 (1.4) & 3600.2 (0) & 85 (0.4) & 0.2 (0.3) & 2153.5 (1012) & 85 (0.4) & 2.3 (0.7) & 3600.4 (0.5) & 85 (0.4) & 6 (1.3) & 3600.2 (0.1) \\
          &       & 10    & 85 (0.4) & 0 (0) & 33.7 (17.4) & 85 (0.4) & 0 (0) & 32 (9.8) & 85 (0.4) & 0 (0) & 25.8 (8.3) & 85 (0.4) & 0 (0) & 36.6 (24.7) & 85 (0.4) & 0 (0) & 54.5 (27.1) \\
\cmidrule{2-18}          & \multirow{5}[2]{*}{10} & 0     & 78.7 (2.9) & 0.5 (0.4) & 3271 (737.5) & 78.1 (1.1) & 0.7 (0.2) & 3600.6 (0.6) & 79 (1.9) & 0.6 (0.2) & 3601.1 (1.9) & 78.7 (1) & 1.2 (0.2) & 3601 (0.9) & 78.8 (3.4) & 0.5 (0.3) & 3098.2 (1122.8) \\
          &       & 0.01  & 82 (1.4) & 2.7 (0.7) & 3600.9 (0.5) & 83.1 (0.7) & 2.9 (0.2) & 3602.2 (4) & 83.8 (0.7) & 2.3 (0.2) & 3600.4 (0.3) & 83 (0.9) & 3.5 (0.3) & 3600.7 (0.6) & 81.6 (0.9) & 3.5 (0.9) & 3601.1 (0.8) \\
          &       & 0.1   & 84.8 (0.5) & 7.7 (2.1) & 3601.5 (2.4) & 85 (0.4) & 5.6 (1.3) & 3600.6 (0.4) & 85 (0.4) & 3 (0.3) & 3600.5 (0.4) & 84.9 (0.5) & 5.3 (0.6) & 3602.8 (5.4) & 84.9 (0.5) & 8.6 (1.4) & 3600.8 (0.6) \\
          &       & 1     & 85 (0.4) & 7.6 (1.3) & 3600.2 (0.2) & 85 (0.4) & 4.4 (1.3) & 3600.4 (0.2) & 85 (0.4) & 1.9 (0.4) & 3600.9 (0.6) & 85 (0.4) & 4.2 (0.9) & 3601.6 (2.1) & 85 (0.4) & 7.6 (1.3) & 3604.6 (7.9) \\
          &       & 10    & 85 (0.4) & 0 (0) & 269.1 (98.1) & 85 (0.4) & 0 (0) & 201.3 (61) & 85 (0.4) & 0 (0) & 120.6 (60.1) & 85 (0.4) & 0 (0) & 174.2 (66.4) & 85 (0.4) & 0 (0) & 257.9 (41.3) \\
    \bottomrule
    \end{tabular}%
    }
    \caption{PNN Hyperparameter Tuning Results for 10-covariate data (Athey Model 2) with $\ell_1$ Regularization.}
  \label{l1athey2results}%
\end{table*}%

\begin{table*}[ht]
  \centering
  \resizebox{15cm}{!}{%
    \begin{tabular}{ccclllllllllllllll}
    \toprule
    \multicolumn{1}{c}{\multirow{3}[6]{1cm}{\centering Train Data Length}} & \multicolumn{1}{c}{\multirow{3}[6]{1cm}{\centering Width}} & \multicolumn{1}{c}{\multirow{3}[6]{1.5cm}{\centering Hyper parameter}} & \multicolumn{15}{c}{Probability of Correct Assignments} \\
\cmidrule{4-18}          &       &       & \multicolumn{3}{c}{0.1} & \multicolumn{3}{c}{0.25} & \multicolumn{3}{c}{0.5} & \multicolumn{3}{c}{0.75} & \multicolumn{3}{c}{0.9} \\
\cmidrule{4-18}          &       &       & \multicolumn{1}{c}{OOSP (\%)} & \multicolumn{1}{c}{Gap (\%)} & \multicolumn{1}{c}{Runtime (s)} & \multicolumn{1}{c}{OOSP (\%)} & \multicolumn{1}{c}{Gap (\%)} & \multicolumn{1}{c}{Runtime (s)} & \multicolumn{1}{c}{OOSP (\%)} & \multicolumn{1}{c}{Gap (\%)} & \multicolumn{1}{c}{Runtime (s)} & \multicolumn{1}{c}{OOSP (\%)} & \multicolumn{1}{c}{Gap (\%)} & \multicolumn{1}{c}{Runtime (s)} & \multicolumn{1}{c}{OOSP (\%)} & \multicolumn{1}{c}{Gap (\%)} & \multicolumn{1}{c}{Runtime (s)} \\
    \midrule
    \multirow{10}[4]{*}{100} & \multirow{5}[2]{*}{3} & 0     & 56.8 (2.7) & 0 (0) & 0.1 (0) & 58.4 (5.3) & 0 (0) & 0.1 (0) & 60.2 (4.3) & 0 (0) & 0.1 (0) & 56 (6) & 0 (0) & 0.1 (0) & 61.5 (10.6) & 0 (0) & 0.5 (0.9) \\
          &       & 0.01  & 84.9 (7.6) & 0.1 (0) & 0.1 (0.1) & 89.2 (5.4) & 0.1 (0) & 0.1 (0.2) & 92.6 (1.9) & 0.1 (0) & 4.7 (6.4) & 83.6 (9.5) & 0.1 (0) & 8.5 (18.9) & 83.1 (8.9) & 0.1 (0) & 734.6 (1602.1) \\
          &       & 0.1   & 87.8 (6.2) & 0.3 (0.1) & 3600.1 (0) & 93.6 (0.7) & 0.2 (0.2) & 2862.8 (1022) & 93.3 (1.1) & 0.2 (0.2) & 2287.7 (1802.4) & 84.8 (8.1) & 0.4 (0.2) & 3470.3 (290.3) & 85.1 (6.9) & 0.6 (0.3) & 3600.1 (0.1) \\
          &       & 1     & 93.6 (1.9) & 1.8 (1.5) & 2891.1 (1585.4) & 95.8 (0.2) & 0 (0) & 506.7 (762.7) & 95.8 (0.2) & 0.4 (0.8) & 798.9 (1568.6) & 87.6 (9.3) & 1.6 (1.4) & 2355 (1719.4) & 88.3 (8.9) & 2.1 (1.7) & 2886.5 (1595.7) \\
          &       & 10    & 95.8 (0.2) & 0 (0) & 3.6 (2.8) & 95.8 (0.2) & 0 (0) & 0.7 (0.3) & 95.8 (0.2) & 0 (0) & 1 (1) & 93 (6.1) & 4.3 (9.6) & 736.2 (1601.1) & 95.8 (0.2) & 3.8 (8.5) & 724.1 (1607.7) \\
\cmidrule{2-18}          & \multirow{5}[2]{*}{10} & 0     & 54.1 (4.2) & 0 (0) & 0.3 (0) & 55.4 (4) & 0 (0) & 0.3 (0) & 55 (3.9) & 0 (0) & 0.3 (0) & 48.3 (6.7) & 0 (0) & 0.3 (0) & 56.8 (13) & 0 (0) & 0.9 (1.4) \\
          &       & 0.01  & 84.9 (7.6) & 0.1 (0) & 0.6 (0.7) & 89.2 (5.4) & 0.1 (0) & 0.6 (0.7) & 92.8 (1.4) & 0.1 (0) & 20.5 (33.9) & 83.6 (9.5) & 0.1 (0) & 57.2 (127.1) & 83.8 (8.7) & 0.1 (0) & 836.6 (1564.8) \\
          &       & 0.1   & 87.8 (6.2) & 0.5 (0.1) & 3600.1 (0.1) & 93.6 (0.7) & 0.4 (0.2) & 3600.1 (0) & 93.3 (1.1) & 0.4 (0.2) & 3600.1 (0) & 85.5 (8.7) & 0.5 (0.2) & 3600.2 (0.1) & 84.9 (6.7) & 0.7 (0.3) & 3600.2 (0.1) \\
          &       & 1     & 93.6 (1.9) & 2.7 (1.8) & 3429.7 (381) & 95.8 (0.2) & 0.8 (0.6) & 3241 (803) & 95.8 (0.2) & 0.7 (1.2) & 2198.3 (1634.8) & 87.6 (9.3) & 2.3 (1.6) & 3600.2 (0.1) & 88.3 (8.9) & 3.3 (1.9) & 3600.1 (0) \\
          &       & 10    & 95.8 (0.2) & 0 (0) & 140.3 (169.9) & 95.8 (0.2) & 0 (0) & 15 (7.6) & 95.8 (0.2) & 0 (0) & 13.3 (7) & 95.8 (0.2) & 8.4 (18.7) & 1266.6 (1463.8) & 95.8 (0.2) & 7 (15.7) & 884.1 (1530.8) \\
    \midrule
    \multirow{10}[4]{*}{500} & \multirow{5}[2]{*}{3} & 0     & 86.4 (3) & 0.2 (0.1) & 1980.4 (1568.9) & 87.2 (1.4) & 0.1 (0.1) & 1819.6 (1676.9) & 89.3 (2.2) & 0.1 (0) & 59.3 (42.2) & 86.8 (1.6) & 0.1 (0) & 1333.6 (1566) & 87.8 (2.3) & 0.2 (0.2) & 1976.9 (1524.1) \\
          &       & 0.01  & 90.3 (0.8) & 0.7 (0.2) & 3600.4 (0.5) & 91.9 (0.5) & 0.6 (0.1) & 3600.4 (0.4) & 93.8 (0.9) & 0.2 (0) & 3600.1 (0) & 92.6 (0.7) & 0.4 (0.1) & 3600.1 (0.1) & 89.6 (1) & 0.7 (0.2) & 3600.1 (0.1) \\
          &       & 0.1   & 94.4 (0.6) & 2.6 (0.5) & 3600.1 (0.1) & 95.7 (0.1) & 1.4 (0.2) & 3600.2 (0) & 95.7 (0.1) & 0.5 (0.2) & 3600.5 (0.6) & 95.7 (0.1) & 1.3 (0.3) & 3600.4 (0.6) & 94.1 (0.3) & 2.4 (0.3) & 3600.2 (0.1) \\
          &       & 1     & 95.7 (0.1) & 2.4 (0.5) & 3600.2 (0.1) & 95.7 (0.1) & 0.1 (0.1) & 1995.3 (1388.4) & 95.7 (0.1) & 0.1 (0) & 47.7 (19.7) & 95.7 (0.1) & 0 (0) & 1416.9 (1324.2) & 95.7 (0.1) & 2.3 (0.7) & 3600.2 (0.1) \\
          &       & 10    & 95.7 (0.1) & 0 (0) & 66.1 (29.8) & 95.7 (0.1) & 0 (0) & 78.4 (18.3) & 95.7 (0.1) & 0 (0) & 37.1 (22.3) & 95.7 (0.1) & 0 (0) & 65.5 (33.3) & 95.7 (0.1) & 0 (0) & 98.4 (53.5) \\
\cmidrule{2-18}          & \multirow{5}[2]{*}{10} & 0     & 79.7 (2) & 0 (0) & 105.3 (120.5) & 79 (2.7) & 0 (0) & 99.8 (85.6) & 84.1 (1) & 0 (0) & 75.7 (79.9) & 80.6 (4.1) & 0 (0) & 84.4 (64.5) & 81 (3.9) & 0 (0) & 555.6 (958.4) \\
          &       & 0.01  & 92.8 (1.2) & 0.4 (0) & 3601.5 (1.8) & 92.5 (1.1) & 0.5 (0.1) & 3602.8 (4.1) & 94.1 (0.6) & 0.3 (0) & 3603 (5.1) & 92.7 (1.1) & 0.4 (0.1) & 3600.6 (0.3) & 92 (0.7) & 0.4 (0) & 3600.8 (0.9) \\
          &       & 0.1   & 94.5 (0.6) & 2.7 (0.5) & 3601.5 (2) & 95.7 (0.1) & 1.4 (0.2) & 3602.1 (1.9) & 95.7 (0.1) & 0.6 (0.1) & 3601.4 (2.1) & 95.7 (0.1) & 1.3 (0.3) & 3601.5 (2.7) & 94.4 (0.6) & 2.6 (0.4) & 3608.6 (10.1) \\
          &       & 1     & 95.7 (0.1) & 3.4 (0.4) & 3600.4 (0.1) & 95.7 (0.1) & 1.2 (0.2) & 3600.6 (0.1) & 95.7 (0.1) & 0.1 (0.1) & 2094.5 (904.1) & 95.7 (0.1) & 1 (0.3) & 3600.5 (0.1) & 95.7 (0.1) & 3.4 (0.7) & 3601.3 (0.9) \\
          &       & 10    & 95.7 (0.1) & 0 (0) & 471.4 (190.6) & 95.7 (0.1) & 0 (0) & 193.5 (72.4) & 95.7 (0.1) & 0 (0) & 166.7 (55) & 95.7 (0.1) & 0 (0) & 232.3 (113.5) & 95.7 (0.1) & 0 (0) & 540.7 (419.4) \\
    \bottomrule
    \end{tabular}%
    }
    \caption{PNN Hyperparamter Tuning Results for 20-covariate data (Athey Model 3) with $\ell_1$ Regularization.}
  \label{l1athey3results}%
\end{table*}%

\subsubsection{Personalized Postpartum Hypertension Treatment}
Please see Table \ref{hypertensionresults}.
\begin{table*}[ht]
  \centering
  \resizebox{15cm}{!}{%
    \begin{tabular}{ccrrrrll}
    \toprule
    \multicolumn{1}{p{4.835em}}{\centering Experiment} & \multicolumn{1}{p{4.585em}}{\centering Width} & \multicolumn{1}{p{4.5em}}{\centering Hyper Parameter} & \multicolumn{1}{p{3.5em}}{\centering Gap (\%)} & \multicolumn{1}{p{5em}}{\centering Runtime (s)} & \multicolumn{1}{p{3.915em}}{\centering Average Outcome} & \multicolumn{1}{p{6.835em}}{\centering 95\% Confidence Interval} & \multicolumn{1}{p{6.835em}}{\centering 90\% Confidence Interval} \\
    \midrule
    \multirow{10}[4]{*}{1} & \multirow{5}[2]{*}{3} & 0     & 0.01  & 3600.31 & 148.41 & [147.08, 149.74] & [147.08, 149.74] \\
          &       & 0.01  & 0.01  & 3600.31 & 148.5 & [147.14, 149.87] & [147.14, 149.87] \\
          &       & 0.1   & 0.03  & 3600.42 & 146.43 & [145.2, 147.66] & [145.2, 147.66] \\
          &       & 1     & 0.01  & 3041.54 & 146.43 & [145.2, 147.66] & [145.2, 147.66] \\
          &       & 10    & 0     & 5.49  & 146.43 & [145.2, 147.66] & [145.2, 147.66] \\
\cmidrule{2-8}          & \multirow{5}[2]{*}{10} & 0     & 0     & 2333.51 & 145.76 & [143.71, 147.81] & [143.71, 147.81] \\
          &       & 0.01  & 0.01  & 3600.36 & 149.02 & [147.79, 150.26] & [147.79, 150.26] \\
          &       & 0.1   & 0.04  & 3600.38 & 146.43 & [145.2, 147.66] & [145.2, 147.66] \\
          &       & 1     & 0.03  & 3600.37 & 146.43 & [145.2, 147.66] & [145.2, 147.66] \\
          &       & 10    & 0     & 34.95 & 146.43 & [145.2, 147.66] & [145.2, 147.66] \\
    \midrule
    \multirow{10}[4]{*}{2} & \multirow{5}[2]{*}{3} & 0     & 0.03  & 3600.14 & 148.1 & [146.32, 149.88] & [146.32, 149.88] \\
          &       & 0.01  & 0.03  & 3601.03 & 149.46 & [147.71, 151.22] & [147.71, 151.22] \\
          &       & 0.1   & 0.05  & 3602.06 & 149.77 & [147.67, 151.87] & [147.67, 151.87] \\
          &       & 1     & 0.07  & 3600.48 & 157.07 & [146.98, 167.17] & [146.98, 167.17] \\
          &       & 10    & 0     & 102.39 & 148.11 & [146.26, 149.97] & [146.26, 149.97] \\
\cmidrule{2-8}          & \multirow{5}[2]{*}{10} & 0     & 0.01  & 3600.97 & 147.3 & [144.48, 150.11] & [144.48, 150.11] \\
          &       & 0.01  & 0.02  & 3600.4 & 158.12 & [147.85, 168.39] & [147.85, 168.39] \\
          &       & 0.1   & 0.07  & 3600.3 & 154.68 & [144.66, 164.7] & [144.66, 164.7] \\
          &       & 1     & 0.09  & 3600.37 & 156.82 & [146.72, 166.92] & [146.72, 166.92] \\
          &       & 10    & 0     & 644.74 & 150.68 & [148.66, 152.69] & [148.66, 152.69] \\
    \midrule
    \multirow{10}[4]{*}{3} & \multirow{5}[2]{*}{3} & 0     & 0.05  & 3600.11 & 148.83 & [147.17, 150.49] & [147.17, 150.49] \\
          &       & 0.01  & 0.06  & 3600.17 & 146.89 & [145.93, 147.85] & [145.93, 147.85] \\
          &       & 0.1   & 0.07  & 3600.61 & 149.1 & [146.97, 151.24] & [146.97, 151.24] \\
          &       & 1     & 0.09  & 3600.16 & 146.05 & [144.07, 148.03] & [144.07, 148.03] \\
          &       & 10    & 0     & 1408.52 & 146.92 & [145.06, 148.79] & [145.06, 148.79] \\
\cmidrule{2-8}          & \multirow{5}[2]{*}{10} & 0     & 0.06  & 3600.31 & 147.4 & [144.15, 150.64] & [144.15, 150.64] \\
          &       & 0.01  & 0.05  & 3600.5 & 157.74 & [153.55, 161.92] & [153.55, 161.92] \\
          &       & 0.1   & 0.09  & 3600.34 & 145.38 & [143.65, 147.12] & [143.65, 147.12] \\
          &       & 1     & 0.07  & 3601.26 & 146.79 & [145.12, 148.46] & [145.12, 148.46] \\
          &       & 10    & 0     & 3233.17 & 146.14 & [144.56, 147.71] & [144.56, 147.71] \\
    \bottomrule
    \end{tabular}%
    }
    \caption{PNN Hyperparamter Tuning Results for Hypertension Experiments with $\ell_0$ Regularization.}
  \label{hypertensionresults}%
\end{table*}%


\subsection{Model Interpretation Results}\label{interpret_appendix}
In this section, we provide the complete results of the model interpretation analyses. 

\subsubsection{Personalized Postpartum Hypertension Treatments}
For the postpartum hypertension experiments, we ranked the features by their SHAP values for each model. To account for the 10 cross-validation folds, these rankings were then averaged. Since the prescriptive trees branch on at most 3 features, those that were not chosen to branch on were given the lowest ranking. Recall that the prescriptive trees optimize on a binarized feature space. To fairly compare the PNN and CF models with the tree models, we took the minimum rank for each binarized feature as the reported rank. Finally, to account for varied number of features due the binarization, the rankings were normalized between 0 and 1.

The feature names in Tables \ref{hypertension shap ranking exp1}-\ref{hypertension shap ranking exp3} correspond to the demographic and medical characteristics Table \ref{features-descriptions}.

\begin{table*}[ht]
\centering
\resizebox{15cm}{!}{%
\begin{tabular}{|>{\raggedright\arraybackslash}m{30mm}|m{100mm}|}
\hline
\multicolumn{1}{|>{\centering\arraybackslash}m{30mm}|}{\textbf{Feature Name}} 
    & \multicolumn{1}{>{\centering\arraybackslash}m{100mm}|}{\textbf{Description}}\\
\hline
mode.del & Mode of delivery: Cesarean/Vaginal \\
\hline
chronicHTN & Chronic Hypertension: Yes/No \\
\hline
gestHTN & Gestational Hypertension: Yes/No \\
\hline
bmi.prenatal & Pre-natal maternal BMI (Continuous) \\
\hline
sipe  & Superimposed Pre-eclampsia: Yes/No \\
\hline
race  & White, Black, Other \\
\hline
ethnicity & Non-hispanic, Hispanic \\
\hline
feeding.br & Breast feeding: Yes/No \\
\hline
mom.age & Maternal age (Discrete) \\
\hline
ethnicity & Non-hispanic, Hispanic \\
\hline
insurance & Private, Medicaid, Other, Unknown \\
\hline
tobacco & Tobacco use: Yes/No \\
\hline
DM.gest & Gestational Diabetes: Yes/No \\
\hline
gest.age & Gestational Age (Discrete) \\
\hline
pree  & Pre-eclampsia: Yes/No \\
\hline
pree.w.sf & Pre-eclampsia with Severe Features: Yes/No \\
\hline
eclampsia & Eclampsia: Yes/No \\
\hline
hellp & Hemolysis, Elevated Liver Enzymes, Low Platelet Count: Yes/No \\
\hline
\end{tabular}
}
\caption{Feature names and their descriptions for the postpartum hypertension experiments.}
  \label{features-descriptions}%
\end{table*}

\begin{table*}[ht]
  \centering
  \resizebox{15cm}{!}{%
    \begin{tabular}{ccccccccccccccccccccc}
    \toprule
    Model & \multicolumn{1}{l}{Architecture} & mode.del & chronicHTN & gestHTN & bmi.prenatal & sipe  & race  & ethnicity & feeding.br & mom.age & DM.pregest & insurance & tobacco & DM.gest & gest.age & pree  & pree.w.sf & eclampsia & hellp & dcDayPP \\
    \midrule
    CF    & 100   & 0.600 & 0.509 & 0.691 & 0.915 & 0.412 & 0.376 & 0.309 & 0.273 & 0.915 & 0.200 & 0.055 & 0.158 & 0.279 & 0.776 & 0.636 & 0.545 & 0.048 & 0.000 & 0.697 \\
    \midrule
    \multirow{2}[2]{*}{J-PT} & 1     & 0.000 & 0.000 & 0.000 & 0.000 & 0.000 & 0.000 & 0.000 & 0.000 & 0.000 & 0.000 & 0.000 & 0.000 & 0.000 & 0.000 & 0.000 & 0.000 & 0.000 & 0.000 & 0.981 \\
          & 2     & 0.000 & 0.200 & 0.000 & 0.300 & 0.000 & 0.000 & 0.000 & 0.200 & 0.100 & 0.200 & 0.494 & 0.000 & 0.000 & 0.400 & 0.000 & 0.000 & 0.100 & 0.094 & 0.100 \\
    \midrule
    \multirow{2}[2]{*}{B-PT} & 1     & N/A   & N/A   & N/A   & N/A   & N/A   & N/A   & N/A   & N/A   & N/A   & N/A   & N/A   & N/A   & N/A   & N/A   & N/A   & N/A   & N/A   & N/A   & N/A \\
          & 2     & 0.778 & 0.222 & 0.000 & 0.222 & 0.000 & 0.000 & 0.000 & 0.333 & 0.222 & 0.111 & 0.222 & 0.000 & 0.000 & 0.222 & 0.000 & 0.000 & 0.222 & 0.000 & 0.222 \\
    \midrule
    \multirow{2}[2]{*}{K-PT} & 1     & N/A   & N/A   & N/A   & N/A   & N/A   & N/A   & N/A   & N/A   & N/A   & N/A   & N/A   & N/A   & N/A   & N/A   & N/A   & N/A   & N/A   & N/A   & N/A \\
          & 2     & 0.000 & 0.000 & 0.000 & 0.556 & 0.000 & 0.111 & 0.000 & 0.111 & 0.111 & 0.000 & 0.111 & 0.000 & 0.000 & 0.333 & 0.000 & 0.000 & 0.000 & 0.000 & 0.111 \\
    \midrule
    \multirow{2}[2]{*}{PNN} & 3     & N/A   & N/A   & N/A   & N/A   & N/A   & N/A   & N/A   & N/A   & N/A   & N/A   & N/A   & N/A   & N/A   & N/A   & N/A   & N/A   & N/A   & N/A   & N/A \\
          & 10    & 0.614 & 0.659 & 0.777 & 0.677 & 0.373 & 0.277 & 0.186 & 0.168 & 0.682 & 0.164 & 0.282 & 0.150 & 0.382 & 0.627 & 0.850 & 0.595 & 0.000 & 0.250 & 0.877 \\
    \bottomrule
    \end{tabular}%
    }
  \caption{Normalized Rankings of Features for Postpartum Hypertension Experiment 1.}
  \label{hypertension shap ranking exp1}%
\end{table*}%

\begin{table*}[ht]
  \centering
    \resizebox{15cm}{!}{%
    \begin{tabular}{ccccccccccccccccccccc}
    \toprule
    Model & \multicolumn{1}{l}{Architecture} & mode.del & chronicHTN & gestHTN & bmi.prenatal & sipe  & race  & ethnicity & feeding.br & mom.age & DM.pregest & insurance & tobacco & DM.gest & gest.age & pree  & pree.w.sf & eclampsia & hellp & bpmed \\
    \midrule
    CF    & 100   & 0.474 & 0.684 & 0.605 & 0.874 & 0.605 & 0.426 & 0.168 & 0.342 & 0.932 & 0.153 & 0.247 & 0.142 & 0.405 & 0.868 & 0.616 & 0.689 & 0.147 & 0.000 & 0.621 \\
    \midrule
    \multirow{2}[2]{*}{J-PT} & 1     & 0.000 & 0.000 & 0.000 & 0.000 & 0.100 & 0.000 & 0.000 & 0.300 & 0.000 & 0.000 & 0.000 & 0.000 & 0.000 & 0.000 & 0.000 & 0.000 & 0.600 & 0.000 & 0.000 \\
          & 2     & 0.096 & 0.000 & 0.000 & 0.265 & 0.100 & 0.100 & 0.590 & 0.194 & 0.100 & 0.100 & 0.400 & 0.100 & 0.000 & 0.200 & 0.000 & 0.000 & 0.000 & 0.000 & 0.000 \\
    \midrule
    \multirow{2}[2]{*}{B-PT} & 1     & 0.007 & 0.380 & 0.007 & 0.380 & 0.000 & 0.007 & 0.007 & 0.007 & 0.007 & 0.007 & 0.007 & 0.007 & 0.007 & 0.007 & 0.007 & 0.007 & 0.007 & 0.007 & 0.251 \\
          & 2     & 0.006 & 0.006 & 0.006 & 0.168 & 0.495 & 0.006 & 0.404 & 0.105 & 0.801 & 0.006 & 0.006 & 0.000 & 0.006 & 0.006 & 0.006 & 0.006 & 0.199 & 0.105 & 0.090 \\
    \midrule
    \multirow{2}[2]{*}{K-PT} & 1     & 0.000 & 0.000 & 0.000 & 0.000 & 0.000 & 0.000 & 0.000 & 0.000 & 0.000 & 0.000 & 0.000 & 0.000 & 0.000 & 0.167 & 0.000 & 0.000 & 0.000 & 0.000 & 0.817 \\
          & 2     & 0.780 & 0.000 & 0.000 & 0.900 & 0.000 & 0.000 & 0.000 & 0.049 & 0.000 & 0.000 & 0.090 & 0.000 & 0.000 & 0.000 & 0.000 & 0.000 & 0.000 & 0.100 & 0.969 \\
    \midrule
    \multirow{2}[2]{*}{PNN} & 3     & 0.821 & 0.663 & 0.737 & 0.532 & 0.558 & 0.437 & 0.337 & 0.168 & 0.658 & 0.211 & 0.279 & 0.168 & 0.368 & 0.668 & 0.837 & 0.658 & 0.000 & 0.147 & 0.753 \\
          & 10    & 0.926 & 0.593 & 0.825 & 0.667 & 0.450 & 0.402 & 0.175 & 0.175 & 0.661 & 0.233 & 0.233 & 0.206 & 0.434 & 0.656 & 0.815 & 0.587 & 0.000 & 0.116 & 0.794 \\
    \bottomrule
    \end{tabular}%
    }
  \caption{Normalized Rankings of Features for Postpartum Hypertension Experiment 2.}
  \label{hypertension shap ranking exp2}%
\end{table*}%

\begin{table*}[ht]
  \centering
  \resizebox{15cm}{!}{%
    \begin{tabular}{cccccccccccccccccccc}
    \toprule
    Model & \multicolumn{1}{l}{Architecture} & mode.del & chronicHTN & gestHTN & bmi.prenatal & sipe  & race  & ethnicity & feeding.br & mom.age & DM.pregest & insurance & tobacco & DM.gest & gest.age & pree  & pree.w.sf & eclampsia & hellp \\
    \midrule
    CF    & 100   & 0.500 & 0.772 & 0.617 & 0.889 & 0.544 & 0.494 & 0.178 & 0.333 & 0.906 & 0.167 & 0.256 & 0.078 & 0.439 & 0.867 & 0.622 & 0.656 & 0.183 & 0.000 \\
    \midrule
    \multirow{2}[2]{*}{J-PT} & 1     & 0.000 & 0.000 & 0.000 & 0.800 & 0.000 & 0.000 & 0.000 & 0.000 & 0.000 & 0.000 & 0.100 & 0.000 & 0.000 & 0.000 & 0.000 & 0.000 & 0.000 & 0.000 \\
          & 2     & 0.386 & 0.492 & 0.100 & 0.200 & 0.000 & 0.100 & 0.000 & 0.000 & 0.000 & 0.000 & 0.100 & 0.000 & 0.000 & 0.200 & 0.394 & 0.000 & 0.200 & 0.000 \\
    \midrule
    \multirow{2}[2]{*}{B-PT} & 1     & 0.784 & 0.000 & 0.000 & 0.000 & 0.000 & 0.000 & 0.100 & 0.000 & 0.000 & 0.000 & 0.100 & 0.000 & 0.000 & 0.000 & 0.000 & 0.000 & 0.000 & 0.000 \\
          & 2     & 0.362 & 0.000 & 0.000 & 0.100 & 0.000 & 0.000 & 0.000 & 0.300 & 0.290 & 0.200 & 0.500 & 0.300 & 0.000 & 0.000 & 0.000 & 0.000 & 0.100 & 0.000 \\
    \midrule
    \multirow{2}[2]{*}{K-PT} & 1     & 0.980 & 0.000 & 0.000 & 0.000 & 0.000 & 0.000 & 0.000 & 0.000 & 0.000 & 0.000 & 0.000 & 0.000 & 0.000 & 0.000 & 0.000 & 0.000 & 0.000 & 0.000 \\
          & 2     & 0.568 & 0.000 & 0.000 & 0.300 & 0.100 & 0.000 & 0.296 & 0.494 & 0.100 & 0.000 & 0.000 & 0.000 & 0.000 & 0.100 & 0.000 & 0.196 & 0.000 & 0.000 \\
    \midrule
    \multirow{2}[2]{*}{PNN} & 3     & 0.463 & 0.510 & 0.503 & 0.698 & 0.228 & 0.376 & 0.342 & 0.148 & 0.510 & 0.262 & 0.322 & 0.020 & 0.362 & 0.544 & 0.497 & 0.651 & 0.087 & 0.000 \\
          & 10    & N/A   & N/A   & N/A   & N/A   & N/A   & N/A   & N/A   & N/A   & N/A   & N/A   & N/A   & N/A   & N/A   & N/A   & N/A   & N/A   & N/A   & N/A \\
    \bottomrule
    \end{tabular}%
    }
  \caption{Normalized Rankings of Features for Postpartum Hypertension Experiment 3.}
  \label{hypertension shap ranking exp3}%
\end{table*}%

\end{appendices}


\end{document}